\documentclass[preprint]{article}

\usepackage{neurips_2025}
\usepackage{amssymb}
\usepackage{amsmath}
\usepackage{amsthm}
\usepackage{lineno}
\usepackage{xargs}
\usepackage{xcolor}
\usepackage{hyperref}
\usepackage{cleveref}
\usepackage{autonum} 
\usepackage{booktabs}
\usepackage{bm}
\usepackage{graphicx}
\usepackage{subcaption}
\usepackage{datatool}
\usepackage{array}
\usepackage{wrapfig}
\usepackage[toc,page,header]{appendix}
\usepackage{minitoc}
\setcitestyle{numbers,square}
\newcolumntype{M}[1]{>{\centering\arraybackslash}m{#1}}
\newtheorem{theorem}{Theorem}[section]

\newtheorem{remark}{Remark}[theorem]

\newcommand{\mike}[1]{}
\newcommand{\mikec}[1]{}
\def\githublink{https://github.com/gabrielvc/dgm_anisotropic_grf}
\def\maxsw{Max-SW}
\def\c2st{C2ST}
\def\edm{Heun-EDM}
\def\ddpm{DDPM}
\def\ddim{DDIM}
\def\dps{DPS}
\def\mgps{MGPS}
\def\vecchia{VMCMC}
\def\priorvae{PriorVAE}

\def\paramcond{\theta}
\def\paramprior{\alpha}
\def\paramvi{\phi}
\def\param{\theta}
\def\paramset{\varTheta}
\def\enddiff{T}

\def\gauss{\mathcal{N}}
\DeclareMathOperator*{\argmin}{arg\,min}

\newcommand{\pset}[1]{\mathcal{P}_{#1}}

\newcommand{\kl}[2]{\operatorname{D}_{\operatorname{KL}}\left(#1||#2\right)}

\newcommand{\dens}[2]{#1\left(#2\right)}
\newcommand{\condd}[3]{\dens{#1}{#3 | #2}}

\newcommand{\fwmarg}[1]{p_{#1}}
\newcommand{\mseloss}[2]{\operatorname{MSE}(#1; #2)}
\newcommand{\measop}[1]{f(#1)}
\def\measnoise{\varepsilon_\lmeas}
\newcommand{\vpalt}[1]{\textcolor{orange}{#1}}
\def\meas{Y}
\def\lmeas{y}
\def\measdim{d_y}
\def\ppost{p}
\newcommandx{\ppostd}[2][2=\lmeas]{\condd{\ppost}{#2}{#1}}
\newcommand{\fwmargd}[2]{\dens{p_{#1}}{#2}}

\newcommandx{\pfwmargd}[3][3=\lmeas]{\condd{\ppost_{#1}}{#3}{#2}}
\newcommand{\fwtrans}[2]{p_{#2 | #1}}
\newcommandx{\fwtransd}[4]{\condd{\fwtrans{#1}{#2}}{#3}{#4}}
\newcommandx{\bwmarg}[2][2=]{q^{#2}_{#1}}
\newcommandx{\bwmargd}[3][2=]{\dens{\bwmarg{#1}[#2]}{#3}}
\newcommandx{\bwtrans}[3][3=]{q^{#3}_{#2 | #1}}
\newcommandx{\bwtransd}[5][3=\param]{\condd{\bwtrans{#1}{#2}[#3]}{#4}{#5}}
\newcommandx{\lik}[3][2=\lmeas,3=]{\ell_{#3}(#2|#1)}

\def\rset{\mathbb{R}}
\def\R{\rset}

\def\statedim{{d_x}}

\def\dataset{\mathcal{D}}
\def\pdata{p_{\dataset}}
\def\ndata{n}

\def\bwmean{\mu}
\def\bwstd{\eta}

\newcommandx{\wpdist}[3][1=1]{\operatorname{W}_{#1}(#2, #3)}

\newcommandx\pgen[1][1=\param]{p_{#1, \operatorname{G}}}
\newcommandx{\pgend}[2][1=\param]{\dens{\pgen[#1]}{#2}}
\newcommandx{\empmeas}[1][1=\chunk{\state}{1}{\ndata}]{\mu_{\operatorname{emp}}(#1)}
\newcommand{\chunk}[3]{{#1}_{#2:#3}}
\newcommand{\pdatad}[1]{\dens{\pdata}{#1}}
\newcommandx{\pcond}[1][1=\paramcond]{p_{#1}}
\newcommandx{\pcondd}[3][1=\paramcond]{\condd{\pcond[#1]}{#2}{#3}}
\newcommandx{\pprior}[1][1=\paramprior]{\pi_{#1}}
\newcommandx{\ppriord}[2][1=\paramprior]{\dens{\pprior[#1]}{#2}}
\newcommandx{\pvi}[1][1=\paramvi]{q_{#1}}
\newcommandx{\pvid}[3][1=\paramvi]{\condd{\pvi[#1]}{#2}{#3}}
\newcommandx\pvae[1][1=\param]{p_{#1, \operatorname{full}}}
\newcommandx{\pvaed}[2][1=\param]{\dens{\pvae[#1]}{#2}}
\newcommandx{\dloss}[1]{L(#1)}
\newcommandx{\mloss}[3][3=\lstate]{m(#1, #2; #3)}
\newcommandx{\dmloss}[3][3=\lstate]{g(#1, #2; #3)}
\newcommandx{\empmloss}[3][3=\chunk{\state}{1}{\ndata}]{M_{n}(#1, #2; #3)}
\newcommandx{\tmloss}[2]{M^{\star}(#1, #2)}
\newcommandx{\lpnorm}[3][1=2,2=\pdata]{\left\|#3\right\|_{L^{#1}(#2)}}
\newcommandx{\erisk}[2]{R(#1, #2)}
\newcommand{\lpsp}[2]{\operatorname{L}^{#1}(#2)}
\newcommandx{\dnet}[2][1=\param]{\operatorname{D}_{#1}(#2)}

\def\state{X}

\newcommand{\fwtransvp}[2]{\textcolor{orange}{p_{#2 | #1}}}
\newcommandx{\fwtransvpd}[4]{\textcolor{orange}{\condd{\fwtransvp{#1}{#2}}{#3}{#4}}}

\newcommand{\PE}[2]{\mathbb{E}_{#1}\left[#2\right]}
\newcommand{\PV}[2]{\mathbb{V}_{#1}\left[#2\right]}
\def\rmd{\mathrm{d}}
\def\eqdef{:=}
\def\eqsp{\;}

\newcommand{\nofrac}[2]{#1 \bigg/ #2}
\newcommand{\indi}[1]{\mathrm{1}_{#1}}

\def\lstate{x}

\def\vestd{\sigma}
\def\idm{\mathrm{I}}

\newcommand{\E}{\mathbb{E}} % expectation
\newcommand{\Var}{\mathrm{Var}} % variance
\newcommand{\tr}[1]{\mathrm{Trace}\left(#1\right)} % Trace
\newcommand{\cD}{\mathcal{D}} % Unit square 

\newcommand{\mc}[1]{\mathcal{#1}} % Mathcal
\newcommand{\adir}{v} % Vector field of anisotropy vectors
\newcommand{\ascl}{\rho} % Field of anisotropy ranges
\newcommand{\rmet}{g} % Riemannian metric
\newcommand{\lbo}{-\Delta_\rmet} % Laplace Beltrami operator (LBO)
\newcommand{\grad}{\nabla} % Riemannian gradien
\newcommand{\LD}{\lpsp{2}{\cD,\rmet}} % L2 space on the Riem. man.
\newcommand{\psd}{\gamma} % Variance scaling function for GRF definition
\newcommand{\evlb}{\lambda} % Eigenval of LBO
\newcommand{\eflb}{e} % Eigenfunctions of LBO
\newcommand{\ptD}{s} % Point in D

\newcommand{\rnorm}{\noise} % Symbol for (vector of) standard Gaussian variables
\newcommand{\femb}{\psi} % Finite element basis functions
\newcommand{\evglb}{\Lambda} % Eigenvalues of discretized operator/stiffness matrix
\newcommand{\efglb}{E} % Eigenvectors of discretized operator/stiffness matrix

\newcommand{\cZ}{\mathcal{\state}} % "True" GRF 

\newcommand{\gZ}{\widehat{\mathcal{\state}}} % FEM discretized GRF
\newcommand{\gZfem}{\state} % Node values of FEM discretized GRF
\newcommand{\gZcov}{\mat{\Sigma}_{\state}} % Covariance matrix of  FEM discretized GRF

\newcommand{\Mmat}{\mat{C}} % Mass lumped matrix
\newcommand{\Smat}{\mat{S}} % Scaled stiffness matrix
\newcommand{\Rmat}{\mat{R}} % Stiffness matrix

\newcommand{\euclsp}[2]{\langle #1,\, #2\rangle} % Inner product of vectors
\newcommand{\funcsp}[2]{\langle #1,\, #2\rangle_{\LD}} % Inner product on L^2 space of Riem. man.
\newcommand{\onb}{\efglb}

\newcommand{\noise}{W}

\newcommand{\mat}[1]{\bm{#1}} % Matrices
\renewcommand{\vec}[1]{#1} % Vectors

\newcommand{\ceil}[1]{\lceil #1\rceil} % Vectors

\newcommand{\cDext}{\cD_{\mathrm{ext}}}

\newcommand{\pfixed}{p_{\mathrm{fixed}}}
\newcommand{\sfixed}{\gZfem_0}
\newcommand{\nfixed}{\gZfem_\vestd}
\newcommand{\nnfixed}{x_\vestd}
\newcommand{\cmin}{c_{\min}}
\newcommand{\cmax}{c_{\max}}
\newcommand{\sorder}[1]{\asymp \big( #1 \big)}
\newcommand{\intvect}[2]{\left\{#1, \cdots, #2\right\}}
\newcommand{\gabriel}[1]{}
\def\mgdm{MGDM}

\newcommandx{\q}[1]{``#1''}

\DTLloaddb{max_sws}{data/max_sws.csv}
\DTLloaddb{max_sws_all}{data/max_sws_complete.csv}
\DTLloaddb{classif_table}{data/classifier_metrics.csv}%
\DTLloaddb{crps_temp}{data/crps_post_temp.csv}
\title{Predictive posterior sampling from non-stationnary Gaussian process priors via Diffusion models with application to climate data.}

\author{
    Gabriel V. Cardoso\thanks{Both authors contributed equally.}\\
    Geostatistics team, Centre for geosciences and geoengineering\\
    Mines Paris, PSL University\\
    Fontaineableau, France \\
    \texttt{gabriel.victorino\_cardoso@minesparis.psl.eu}\\
    \And
    Mike Pereira$^*$\\
    Geostatistics team, Centre for geosciences and geoengineering\\
    Mines Paris, PSL University\\
    Fontaineableau, France \\
    \texttt{mike.pereira@minesparis.psl.eu} \\
}
\begin{document}
\begin{filecontents*}{data/max_sws.csv}
,sampler,n_steps,rho,mean,std,len,clt_gap,value
12,DDIM,100,3.000,0.204,0.010,20,0.004,0.204 + 0.004
3,DDPM,100,3.000,0.171,0.008,20,0.003,0.171 + 0.003
8,DDPM,1000,3.000,0.095,0.008,20,0.004,0.095 + 0.004
18,Heun,128,3.000,0.181,0.009,20,0.004,0.181 + 0.004
0,Train,0,,0.069,0.006,20,0.002,0.069 + 0.002
\end{filecontents*}

\begin{filecontents*}{data/max_sws_complete.csv}
,sampler,n_steps,rho,mean,std,len,clt_gap,value
10,DDIM,100,1.000,0.224,0.008,20,0.004,0.224 + 0.004
11,DDIM,100,2.000,0.202,0.010,20,0.004,0.202 + 0.004
12,DDIM,100,3.000,0.204,0.010,20,0.004,0.204 + 0.004
13,DDIM,100,4.000,0.202,0.009,20,0.004,0.202 + 0.004
14,DDIM,100,5.000,0.207,0.012,20,0.005,0.207 + 0.005
15,DDIM,100,6.000,0.207,0.010,20,0.004,0.207 + 0.004
16,DDIM,100,7.000,0.205,0.009,20,0.004,0.205 + 0.004
1,DDPM,100,1.000,0.201,0.005,20,0.002,0.201 + 0.002
2,DDPM,100,2.000,0.175,0.008,20,0.004,0.175 + 0.004
3,DDPM,100,3.000,0.171,0.008,20,0.003,0.171 + 0.003
4,DDPM,100,4.000,0.173,0.008,20,0.004,0.173 + 0.004
5,DDPM,100,5.000,0.179,0.008,20,0.003,0.179 + 0.003
6,DDPM,100,6.000,0.182,0.009,20,0.004,0.182 + 0.004
7,DDPM,100,7.000,0.183,0.007,20,0.003,0.183 + 0.003
17,DDPM,250,3.000,0.147,0.007,20,0.003,0.147 + 0.003
8,DDPM,1000,3.000,0.095,0.008,20,0.004,0.095 + 0.004
9,DDPM,1000,1.000,0.137,0.006,20,0.002,0.137 + 0.002
18,Heun,128,3.000,0.181,0.009,20,0.004,0.181 + 0.004
0,train,0,,0.069,0.006,20,0.002,0.069 + 0.002
\end{filecontents*}

\begin{filecontents*}{data/classifier_metrics.csv}
,sampler_name,n_steps,network,abs_05-mean,acc-mean,auc-mean,c2st_pvalue-mean,c2st_stat-mean,epoch-mean,f1-mean,lr-Adam-mean,mse_05-mean,train_loss-mean,val_loss-mean,abs_05-std,acc-std,auc-std,c2st_pvalue-std,c2st_stat-std,epoch-std,f1-std,lr-Adam-std,mse_05-std,train_loss-std,val_loss-std
0,DDIM,100,resnet18,0.464,0.651,0.711,0.000,0.651,999.000,0.634,0.000,0.224,0.000,3.278,0.001,0.009,0.013,0.000,0.009,0.000,0.008,0.000,0.001,0.000,0.298
2,DDPM,1000,resnet18,0.446,0.530,0.541,0.000,0.529,999.000,0.507,0.000,0.211,0.000,3.651,0.002,0.001,0.002,,,0.000,0.005,0.000,0.001,0.000,0.112
3,DDPM,1000,resnet50,0.463,0.525,0.535,0.000,0.525,999.000,0.507,0.000,0.223,0.000,5.521,0.000,0.002,0.001,0.000,0.002,0.000,0.006,0.000,0.000,0.000,0.045
1,DDPM,1000,resnet101,0.461,0.520,0.527,0.000,0.520,999.000,0.501,0.000,0.222,0.000,5.312,0.003,0.003,0.003,0.000,0.003,0.000,0.007,0.000,0.002,0.000,0.399
\end{filecontents*}

\begin{filecontents*}{data/crps_post_temp.csv}
,img_id,vecchia,priorvae,mgdm
0,0,0.101 (0.004),0.107 (0.008),0.038 (0.002)
0,1,0.202 (0.01),0.214 (0.012),0.085 (0.008)
0,2,0.352 (0.021),0.36 (0.027),0.22 (0.02)
\end{filecontents*}

\raggedbottom
\maketitle
\begin{abstract}
    Bayesian models based on Gaussian processes (GPs) offer a flexible framework to predict spatially distributed variables with uncertainty. But the use of non-stationary priors, often necessary for capturing complex spatial patterns, makes sampling from the predictive posterior distribution (PPD) computationally intractable. In this paper, we propose a two-step approach based on diffusion generative models (DGMs) to mimic PPDs associated with non-stationary GP priors: we replace the GP prior by a DGM surrogate, and leverage recent advances on training-free guidance algorithms for DGMs to sample from the desired posterior distribution. We apply our approach to a rich non-stationary GP prior from which exact posterior sampling is untractable and validate that the issuing distributions are close to their GP counterpart using several statistical metrics. We also demonstrate how one can fine-tune the trained DGMs to target specific parts of the GP prior. Finally we apply the proposed approach to solve inverse problems arising in environmental sciences, thus yielding state-of-the-art predictions.
\end{abstract}
\section{Introduction}
\label{sec:intro}
In many applied domains, from geosciences \cite{camps2016survey} to climate and environmental sciences \cite{petelin2013evolving,kupilik2024bias} or even cosmology \cite{shafieloo2012gaussian}, it is often the case that the quantities of interest (QOI) are defined across a spatial domain but only measured at a finite set of locations.
Notable examples of QOIs are temperature or humidity, but also concentrations of different chemical substances on different media.

Inferring QOIs across the whole spatial domain from sparse observations, while quantifying the related uncertainties, then becomes a crucial task.
If we denote the spatial values of the QOI by $\state$, our goal is to infer $\state$ from a set of partial observations $\lmeas$. It is often known how the observations are obtained from a given $\state$, theirs relationship being described by a measurement equation of the form
\begin{equation}
    \label{eq:measurement_equation}
    \meas = \measop{\state} + \measnoise \eqsp,
\end{equation} 
where $\measnoise$ is the noise random variable (independent of $\state$), $\measop{\cdot}$ is a measurable known function. The link is established by assuming that  $\lmeas \sim \meas$.
\footnote{In the problem described in the first paragraph, $\measop{\cdot}$ is simply the projection into the space corresponding to the coordinates of the observed locations, but more complex measurement equations are possible.}

In Bayesian inverse problems, one associates to \eqref{eq:measurement_equation} a prior distribution $\bwmarg{0}$ encoding beliefs on the possible values of $\state$.
One is interested in the \emph{a posteriori} distribution of $\state$ given $\lmeas$, which by Bayes theorem is given by $\ppostd{\lstate} \eqdef \lik{\lstate}\bwmargd{0}{\lstate} / L(\lmeas)$ where $L(\lmeas) = \int \lik{\tilde{\lstate}}\bwmargd{0}{\tilde{\lstate}} \rmd \tilde{\lstate}$ and
$\lik{\lstate}$ is the likelihood associated with \eqref{eq:measurement_equation}.
This is particularly useful in the cases of so-called ill-posed inverse problems, which arise when several maxima of $\lstate \rightarrow \lik{\lstate}$ exist.

The choice of prior distribution is key in Bayesian statistics in general, but particularly in ill-posed inverse problems. %, as it will play a role in favoring certain solutions amongst the pool of available solutions.
Particularly in spatial statistics, Gaussian random fields (GRFs) have played a great role as priors %are often used as prior distributions, \gabriel{Here we should probably show some evidence that we can construct physical meaningful priors using GRFs} 
due to the fact that for a relative large subset of inverse problems one can obtain the posterior distribution in closed form \cite{diggle1998model}.
Since, a great effort has been put into creating GRF priors that express different knowledge about the underlying modeled phenomena\cite{gelfand2017bayesian}.

GRFs are specified by a mean and covariance functions. %which greatly impact the posterior distribution.
For several applications, non-stationary models for GRFs, and in particular GRFs exhibiting local anisotropies, are considered the ideal choice, due to the flexibility to represent spatially varying correlation patterns observed in the data. 
%But it is rare that one can be so certain of the underlying physical properties of the solution as to determine a unique covariance function. 
In Bayesian statistics, they are transformed into priors by the usage of a parametric form for the covariance kernels, coupled with a prior distribution over the parameter space. 
Unfortunately, except for a restricted class of distributions, this approach yields intractable posterior distributions.
While methods such as Markov Chain Monte Carlo (MCMC) or INLA (Integrated Nested Laplace approximation) \cite{rue2017bayesian} could in theory be used to sample from these posterior distributions, their practical implementation is limited to specific cases of non-stationarities \cite{rue2009approximate}. % due to the severely ill-posed nature of the inverse problem when working with anisotropic GRFs \cite{rue2009approximate}. 

\begin{wrapfigure}[19]{r}{0.5\linewidth}
    \vspace{-8pt}
    \centering
    \begin{tabular}{
        M{0.245\linewidth}@{\hspace{0.07\tabcolsep}}
        M{0.245\linewidth}@{\hspace{0.07\tabcolsep}}
        M{0.245\linewidth}@{\hspace{0.07\tabcolsep}}
        M{0.245\linewidth}@{\hspace{0.07\tabcolsep}}
        }
        %(a) Data & (b) Observation & (c) {\mgdm} sample & (d) Standard deviation \\
        \includegraphics[width=\hsize]{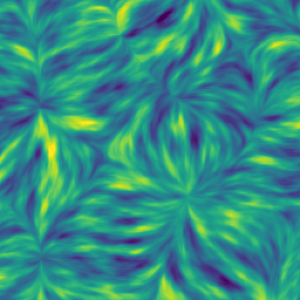} & \includegraphics[width=\hsize]{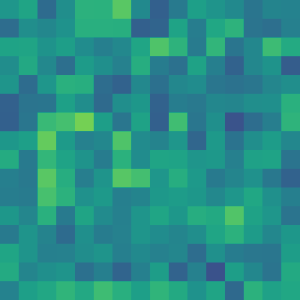} & \includegraphics[width=\hsize]{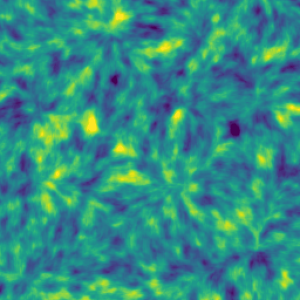} & \includegraphics[width=\hsize]{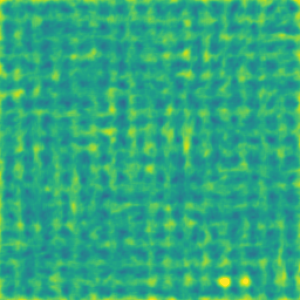} \\\addlinespace[-0.7ex]
        \includegraphics[width=\hsize]{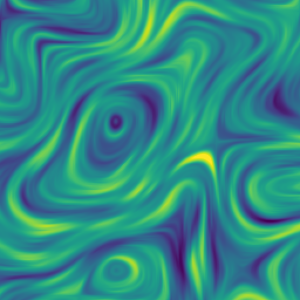} & \includegraphics[width=\hsize]{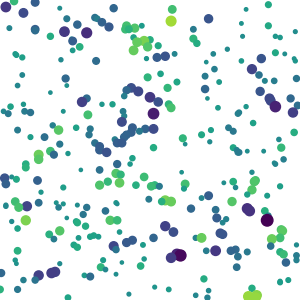} & \includegraphics[width=\hsize]{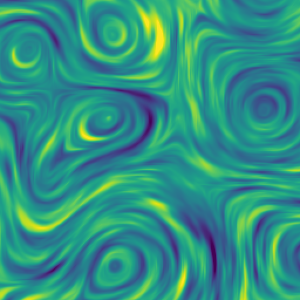} & \includegraphics[width=\hsize]{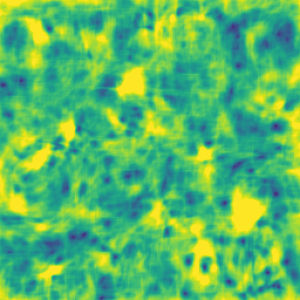} \\\addlinespace[-0.7ex]
        (a) & (b) & (c) & (d)
        %Cloud  & \includegraphics[width=\hsize]{images/clouds/clouds_true_observation.png} & \includegraphics[width=\hsize]{images/clouds/clouds_with_mask.png} & \includegraphics[width=\hsize]{images/clouds/clouds_sample_0.png} &  \includegraphics[width=\hsize]{images/clouds/clouds_std.png}
    \end{tabular}
    \caption{Illustration of the proposed method for super resolution (top row) and inpainting (bottom row) inverse problems. Column (a) shows the complete variable, (b) its partial observation, and (c) shows a sample from {\mgdm} with $\sigma_y = 0.05$ (with colors in the range $[-3,3]$). Column (d) shows the standard deviation over $32$ posterior samples (with colors in the range $[0,1]$).}
    \label{fig:inv_problem_illustration}
\end{wrapfigure}
Concurrently, the use of generative models as \emph{informative} priors has emerged as promising and complementary alternative for solving ill-posed inverse problems (see \cite{elad2023image,ongie2020deep,scarlett2022theoretical,calvetti2018inverse} and references therein).
In those approaches, a generative model is trained using a dataset of $\state$s and the distribution defined by the pre-trained generative model is used as a prior. While this implies a considerable amount of 
work to create the generative model, several efficient algorithms have been proposed to sample from the resulting posterior distribution (see \citet{daras2024survey} and references therein).

In particular, denoising generative models (DGM \cite{song2021scorebased}), also known as score-based generative models, have emerged as one of the most used generative models. They achieve state of the art generative performance
on different modalities such as image \cite{dhariwal2021diffusion}, audio \cite{pascual2023fullband} and video \cite{chen2024videocrafter2}, while avoiding the difficulties of adversarial losses. 
The generative procedure, relying on a Markovian denoising process, is also particularly suited for conditioning and thus makes them one of the most used priors for solving ill-posed inverse problems.
Indeed, they have been successfully applied to medical imaging \cite{chung2023solving}, cardiology \cite{bedin2024leveraging}, audio separation\cite{moufad2024variationala} amongst others applications.

Unfortunately, it is often hard (if not impossible) to obtain direct observation of $\state$ for several QOIs, therefore excluding the possibility of directly training a DGM on real data. 
In this work, we propose to leverage both the richness and physical knowledge expressed by the GRFs and the ability of \emph{post training} conditioning of DGMs. %by proposing an alternative Bayesian approach to infer the posterior distribution when using anisotropic GRF priors.
Namely, we start by generating GRFs realizations from a complex model with a given prior distribution over the underlying parameters.
Then, we learn a DGM and directly condition the resulting DGM distribution on the observations.
Our contributions can be summarized as:
\begin{itemize}
    \item Propose a theoretically sound framework to sample posterior predictive distributions associated with local anisotropic GRF priors using DGMs,
    \item Establish a link between GRF sampling with SPDEs and DGMs, allowing for theoretically backed validation metrics,
    \item Conduct thorough experiments to establish the approximation properties of DGMs for anisotropic GRF priors,
    \item Benchmark several posterior samplers for DGM and their ability to correctly reproduce uncertainties from GRFs in simulated and real world data.
\end{itemize}
\section{Background}
\label{sec:background}
\subsection{Locally anisotropic Gaussian random fields}\label{sec:grf_spde}
Let  $\cD=[0,1]^2$. A GRF $\cZ$ on $\cD$ is locally anisotropic if there exists a radial covariance function $C_0$, a unit-norm vector field  $\adir : \cD \rightarrow \rset^2$ and two scalar fields $\ascl_1, \ascl_2 : \cD \rightarrow (0, \infty)$ such that for any $\ptD \in \cD$,
\begin{equation}
	\text{Cov}(\cZ(\ptD), \cZ(\ptD + h)) \sim C_0(\Vert\bm Q_{\ptD} h\Vert) \quad \text{as } h\in\R^2 \rightarrow 0,
\end{equation}
where $\bm Q_{\ptD}$ is the positive definite matrix with eigenvalues $\ascl_1^{-1}(\ptD), \ascl_2^{-1}(\ptD)$ and associated eigenvectors $\adir(\ptD)$ and its orthogonal. This means in particular that the field $\cZ$ exhibits local directions of correlations defined by $\adir$, and local correlation lengths along the direction $\adir$ (resp. orthogonal to $\adir$) given by $a\ascl_1$ (resp. $a\ascl_2$) where  $a$ is the correlation length associated with $C_0$.

%To model such GRFs, 
We follow the approach described in \cite{pereira2022geostatistics} which consists in defining GRFs as random functions on the Riemannian manifold $(\cD, \rmet)$  obtained by equipping $\cD$ with Neumann boundary conditions and the Riemannian metric $g$ defined at any point $\ptD\in\cD$, by $\rmet_{\ptD}(u_1,u_2)
=\euclsp{ \bm Q_{\ptD} u_1}{\bm Q_{\ptD} u_2}$, where $u_1,u_2\in\rset^2$.
%\begin{equation}
%	\rmet_{\ptD}(u_1,u_2)
%	=\euclsp{ \bm Q_{\ptD} u_1}{\bm Q_{\ptD} u_2}, \quad u_1,u_2\in\rset^2.
%\end{equation}
A spectral theorem ensures that the  Laplace--Beltrami operator $\lbo$ has a discrete spectrum $0\le \evlb_1\le \cdots\le \evlb_k \le \cdots \rightarrow +\infty$  associated with eigenfunctions  $\lbrace \eflb_k\rbrace_{k\in\mathbb{N}}$ forming an orthonormal basis of the set $\LD$ of square-integrable functions of $(\cD, \rmet)$ \cite{lablee_spectral_2015}.
Centered GRFs are then obtained through expansions of the form
\begin{equation}\label{eq:def_Z}
	\cZ = \sum\nolimits_{k\in\mathbb{N}} \psd(\evlb_k) \rnorm_k \eflb_k
\end{equation}
where $\lbrace \rnorm_k\rbrace_{k\in\mathbb{N}}$ is a sequence of independent standard Gaussian variables, and
\begin{equation}\label{def:psd_matern}
	\psd(\lambda) = \tau \big((\sqrt{8\nu}/a)^2+\lambda\big)^{-(\nu+1)/2}, \quad \lambda\in\rset,
\end{equation}
for some $\tau,a,\nu>0$.  Note that $\cZ$ is the spectral decomposition of the solution (in $\LD$) of the stochastic partial differential equation (SPDE) given by $\big((\sqrt{8\nu}/a)^2 \lbo\big)^{(\nu+1)/2}\cZ =\tau \mc{W}$, 
%\begin{equation}\
%	\big((\sqrt{8\nu}/a)^2 \lbo\big)^{(\nu+1)/2}\cZ =\tau \mc{W}
%\end{equation}
where $\mc{W}$ denotes a Gaussian white noise, and as such corresponds to a Whittle-Matérn the ``SPDE approach'' to GRFs introduced by \cite{lindgren2011explicit}. On Euclidean domains, the stationary solutions of such SPDEs are GRFs with a Matérn covariance function with correlation length $a$ and regularity parameter $\nu$, meaning that such a GRF would be $\ceil{\nu}-1$ times differentiable in the mean-square sense.

In practice, the field $\cZ$ is discretized into a grid of $\statedim = 256\times 256$ (regularly-spaced) nodes using the finite element method. Following the Galerkin--Chebyshev approach of \cite{lang2023galerkin} (cf. \Cref{app:grf_spde} for details), the resulting discretized random field $\gZfem \in\rset^{\statedim}$ is a centered Gaussian vector with covariance matrix 
\begin{equation}\label{eq:cov_mat}
	\gZcov = \Mmat^{-1/2}\psd^2(\Smat)\Mmat^{-1/2}
\end{equation}
where  $\Mmat$ and $\Smat$ are  respectively a diagonal and a sparse matrix arising  from the finite element method, and are built using the metric $g$ (cf. \Cref{app:grf_spde}).
% \begin{figure}
% 	\centering
% 	\begin{tabular}{
% 		M{0.16\linewidth}@{\hspace{0.1\tabcolsep}}
% 		M{0.16\linewidth}@{\hspace{0.1\tabcolsep}}
% 		M{0.16\linewidth}@{\hspace{0.1\tabcolsep}}
% 		M{0.16\linewidth}@{\hspace{0.1\tabcolsep}}
% 		M{0.16\linewidth}@{\hspace{0.1\tabcolsep}}
% 		}
% 		\includegraphics[width=\hsize]{images/anisotropic_samples/0_true.png} & \includegraphics[width=\hsize]{images/anisotropic_samples/9_true.png} & \includegraphics[width=\hsize]{images/anisotropic_samples/2_true.png} & \includegraphics[width=\hsize]{images/anisotropic_samples/3_true.png} & \includegraphics[width=\hsize]{images/anisotropic_samples/4_true.png} 
% 	\end{tabular}
%     \caption{Illustration of samples generated with the method described in \Cref{sec:grf_spde}.}
%     \label{fig:grf_illustration}
% \end{figure}
\subsection{Generative modelling via Gaussian denoising (DGM)}
\label{sec:background:dgm}
\subsubsection{Gaussian denoising}\label{sec:gauss_denois}
Gaussian denoising refers to the task of reconstructing a sample $X_0$ from some distribution $\pdata$ using its noisy observation defined as $\state_\vestd \eqdef \state_0 + \vestd \noise$,
% \begin{equation}
%     \label{eq:noised_dist}
%     
% \end{equation}
where $\noise\sim \gauss(0, \idm_{\statedim})$ is independent of $\state_0$.
The goal is to find, within some fixed set $\mathcal{F}$, a function $f^\star : \rset^{\statedim} \rightarrow \rset^{\statedim}$ which minimizes the mean squared error between $X_0$ and its reconstruction $f^\star(\state_{\vestd})$, i.e. 
\begin{equation}
    \label{eq:denoising}
    f^{\star} \in \argmin_{f \in \mathcal{F}} \mseloss{f}{\vestd} \eqdef \PE{}{\|f(\state_\vestd) - \state_0\|^2} \eqsp.
\end{equation}
Note in particular that when $\PE{}{\|\state_0\|^2} < \infty$ and $\mathcal{F} = \lpsp{2}{\pdata}$, $f^\star$ is no other than the conditional expectation of $\state_0$ given $\state_{\vestd}$ : $f^{\star}(\state_{\vestd})=\PE{}{\state_0 | \state_\vestd}$.
However, except for a small class of distributions $\pdata$, the conditional expectation is not available in closed-form. In such cases, one often consider a smaller 
family $\mathcal{F}$, typically the set of linear operators or a parametric family $\mathcal{F} = \{\dnet\cdot | \param \in \paramset\}$ of neural networks (parametrized by  $ \param \in \paramset$).
\subsubsection{Linear denoising with fixed basis}\label{sec:lin_den_fixed}
We assume in this subsection that $\sfixed$ is sampled from the distribution $\pfixed$ of GRFs described in \Cref{sec:grf_spde}, for a fixed choice of range and anisotropy parameters. The particular form of the covariance matrix~\eqref{eq:cov_mat} allows to decompose $\sfixed\sim \pfixed$  as
\begin{equation}\label{eq:decomp_X_basis}
	\sfixed = \Mmat^{-1/2}\sum\nolimits_{k=1}^{\statedim} \rnorm_k(\sfixed) \onb_k
\end{equation}
where  $\lbrace \onb_k\rbrace_{1\le k\le \statedim}$  is an orthonormal basis of  $\rset^\statedim$ composed of eigenvectors of the matrix $\Smat$, and $\lbrace\rnorm_k(\sfixed)\rbrace_{1\le k \le \statedim}$ are independent centered Gaussian  variables satisfying  $\Var[\rnorm_k(\sfixed)]=\psd(\evglb_k)^2$.
In this case, Gaussian denoising of $\nfixed = \sfixed + \sigma \noise$ has an explicit solution owing to the fact that $\sfixed$ and $\noise$ are independent Gaussian vectors. Indeed, since $(\state_{0},\state_{\vestd})$ is also a Gaussian vector, we can write that $(\sfixed \vert \nfixed = \nnfixed) \sim \gauss(\vestd^{-2}\mat{Q}_\sigma^{-1} \nnfixed, \mat{Q}_\sigma^{-1})$
%\begin{equation}
%	(\sfixed \vert \nfixed = \nnfixed) \sim \gauss(\vestd^{-2}\mat{Q}_\sigma^{-1} \nnfixed, \mat{Q}_\sigma^{-1})
%\end{equation}
with $\PE{}{\state_0 | \state_\vestd = \nnfixed} =\vestd^{-2}\mat{Q}_\sigma^{-1} \nnfixed$ and 
\begin{equation}
	\mat{Q}_\vestd=\Var[\sfixed \vert \nfixed = \nnfixed]
	= (\Mmat^{1/2}\psd^{-2}(\bm S)\Mmat^{1/2} + \vestd^{-2}\idm_{\statedim})^{-1},
\end{equation}
Hence, the optimal Gaussian denoiser of $\gZfem_\vestd$ is linear and given by
$f^\star(\gZfem_\vestd) =\vestd^{-2}\mat{Q}_\sigma^{-1} \state_\vestd$.

The resulting MSE between $\state_0$ and its denoised counterpart $f^\star(\gZfem_\vestd)$ can be computed as follows. Let $\cmax = \max_{1\le i\le \statedim}[\Mmat^{-1}]_{ii}$ and $\cmin = (1/2)\min_{1\le i\le \statedim}[\Mmat^{-1}]_{ii}$. We have
%\begin{equation}
%	\begin{aligned}
%		\mseloss{f^\star}{\vestd} &= \PE{}{\|f(\gZfem_\vestd) - \gZfem\|^2} = \tr(\mat{Q}_\vestd^{-1})\\
%		&= \tr(\Mmat^{-1}(\psd^{-2}(\Smat)+\vestd^{-2}\Mmat^{-1})^{-1}) 
%		\le \cmax~\tr((\psd^{-2}(\Smat)+\vestd^{-2}\Mmat^{-1})^{-1})\\
%		&\le \frac{\cmax}{4}(\tr((\psd^{-2}(\Smat)+\cmin\vestd^{-2}\idm_{\statedim})^{-1})+\tr((\vestd^{-2}(\Mmat^{-1}-\cmin\idm_{\statedim}))^{-1})) 
%	\end{aligned}
%\end{equation}
%where we used the convexity of the map $\mat{M} \mapsto \tr(\mat{M}^{-1})$ over the set of positive definite real symmetric matrices to derive the last inequality \citep{muir1974inequalities}.
%Recalling then that by definition of $\psd$ in \eqref{def:psd_matern} and by application of Weyl's asymptotic law, 
%$\psd(\evglb_k)^2 \sorder{\evglb_k^{-(\nu+1)}} \sorder{k^{-(\nu+1)}}$ we can apply the same arguments as \citet{kadkhodaie2023generalization} to conclude that
%\begin{equation}
%	\begin{aligned}
%		\sum_{k=1}^{\statedim} \frac{1}{\psd^{-2}(\evglb_k)+\cmin\vestd^{-2}} \sorder{ \left(\frac{\vestd^2}{\cmin}\right)^{\frac{\nu}{\nu+1}}}
%	\end{aligned}
%\end{equation}
%which in turn yields the following order of magnitude for the optimal denoising error: 
%\begin{equation}
%	\label{eq:background:optimal_denoising}
%	\begin{aligned}
%		\mseloss{f^\star}{\vestd} &\lesssim { \vestd^{\frac{2\nu}{\nu+1}}+\vestd^{2}}
%	\end{aligned}
%\end{equation}
\begin{equation}
		\mseloss{f^\star}{\vestd} = \PE{}{\|f(\gZfem_\vestd) - \state_0\|^2} = \tr(\mat{Q}_\vestd^{-1})=\sum\nolimits_{k=1}^{\statedim}\left[\mu_k + \sigma^{-2}\right]^{-1}
%		&= \tr(\Mmat^{-1}(\psd^{-2}(\Smat)+\vestd^{-2}\Mmat^{-1})^{-1}) 
%		\ge \cmin~\tr((\psd^{-2}(\Smat)+\vestd^{-2}\Mmat^{-1})^{-1})\\
%		&\le \frac{\cmax}{4}(\tr((\psd^{-2}(\Smat)+\cmin\vestd^{-2}\idm_{\statedim})^{-1})+\tr((\vestd^{-2}(\Mmat^{-1}-\cmin\idm_{\statedim}))^{-1})) 
\end{equation}
where $\lbrace \mu_k\rbrace_{1\le k \le \statedim}$ denote the eigenvalues 
of the matrix $\Mmat^{1/2}\psd^{-2}(\bm S)\Mmat^{1/2}$, which are the same as the eigenvalues of the generalized eigenvalue problem associated with the matrices $\psd^{-2}(\bm S)$ and $\Mmat^{-1}$. Writing $\Mmat^{-1} = \cmin\idm_{\statedim}+(\Mmat^{-1}-\cmin\idm_{\statedim})$, and following \cite{crawford_stable_1976}, we get $\mu_k \le \cmin^{-1}\psd^{-2}(\evglb_k) + \vert \cmin^{-1}\psd^{-2}(\evglb_k) - \mu_k\vert \le \cmin^{-1}(1+\Vert \Mmat^{-1} \Vert)\psd^{-2}(\evglb_k)=\cmin^{-1}(1+\cmax)\psd^{-2}(\evglb_k) $, which in turn gives
\begin{equation}
	\begin{aligned}
		\mseloss{f^\star}{\vestd} &\ge \sum\nolimits_{k=1}^{\statedim}\left[\cmin^{-1}(1+\cmax)\psd^{-2}(\evglb_k) + \sigma^{-2}\right]^{-1}\\
	\end{aligned}
\end{equation}
%\mike{To derive a bound for the error above, we use the same arguments as \citet{kadkhodaie2023generalization}, where they derive the same error bound while considering the problem of finding an optimal linear denoiser in a fixed basis, assuming that the coordinates of the original sample in this basis are known.} 
Recalling that by definition of $\psd$ in \eqref{def:psd_matern} and by application of Weyl's asymptotic law, 
 $\psd(\evglb_k)^2 \sorder{\evglb_k^{-(\nu+1)}} \sorder{k^{-(\nu+1)}}$ we can apply the same arguments as \cite{kadkhodaie2023generalization} to conclude that
\begin{equation}
	\begin{aligned}
		\sum\nolimits_{k=1}^{\statedim}\left[\cmin^{-1}(1+\cmax)\psd^{-2}(\evglb_k) + \sigma^{-2}\right]^{-1}
		 \sorder{ \left(\vestd^2\right)^{\frac{\nu}{\nu+1}}}
	\end{aligned}
\end{equation}
which in turn yields the following order of magnitude for the optimal denoising error: 
\begin{equation}
	\label{eq:background:optimal_denoising}
	\begin{aligned}
		\mseloss{f^\star}{\vestd} & \gtrsim  \vestd^{\frac{2\nu}{\nu+1}} %+\vestd^{2}
	\end{aligned}
\end{equation}
Note however in practice, working with the denoiser $f^\star$ introduced above would require to have access to both the parameters defining $\gamma$ (i.e. the range $a$ and regularity $\nu$), but also the the full anistropy fields which are required to build the matrices $\Smat$ and $\Mmat$. 
%Hence, one of the motivations of this work: leveraging VAEs to find a common representation space suitable for a large class of parameter values, and which a denoiser can be learned.

\begin{remark}
	An error bound similar to~\eqref{eq:background:optimal_denoising} has been derived by \cite{kadkhodaie2023generalization}, where
	they show that the denoising error for fixed sample $x_0 \in \rset^\statedim$ can be lower-bounded (for linear denoisers) by $\vestd^{\frac{2\nu}{\nu + 1}}$ (with $\nu >0$) 
	when there exists a basis $\chunk{e}{1}{\statedim}$ such that $\lstate_0^t e_k \sim k^{-(\nu + 1){/2}}$. Determining such basis from noisy versions of $x_0$ is a challenging problem. The authors show
	that the standard DGMs are able to match this lower bound for $C^{\nu}$ images, and retrieve the same decay on the CelebHQ dataset.
\end{remark}
\subsubsection{Generative models from denoising}
\label{subsec:sgm}

The idea of DGMs is to sample from a distribution $\pdata$ by progressively denoising perturbed versions of $\pdata$. Indeed, following \cite{ho2020denoising}, let $(X_0, \dots, X_T)$ be the Markov chain with joint law
\begin{equation}
	\label{eq:noising_joint}
	\fwmargd{0:T}{\chunk{\lstate}{0}{T}} = \pdatad{\lstate_0} \prod\nolimits_{t=0}^{T-1} \fwtransd{t}{t+1}{\lstate_t}{\lstate_{t+1}}\eqsp, \;\; \fwtransd{t}{t+1}{\lstate_t}{\lstate_{t+1}} = \gauss(\lstate_{t}; (\vestd_{t+1}^2 - \vestd_t^2)\idm) \eqsp,
\end{equation}
where $\fwtransd{s}{t}{\cdot}{\cdot}$ is the law of $\state_{t}$ given $\state_{s}$, and by $\fwmarg{t}$  is the marginal law of $\state_{t}$ (with $\fwmarg{0}=\pdata$). Sampling from $\pdata$ can be done by sampling from the backward decomposition of $\fwmarg{0:\enddiff}$, namely:¨
\begin{equation}
	\label{eq:backward_denoising}
	\fwmargd{0:\enddiff}{\chunk{\lstate}{0}{\enddiff}} = \fwmargd{\enddiff}{\lstate_\enddiff} \prod\nolimits_{t = 0}^{\enddiff - 1} \fwtransd{t+1}{t}{\lstate_{t+1}}{\lstate_t} \eqsp,
\end{equation}
While this decomposition is in general intractable, DGMs build a tractable (backward) Markov Chain variational approximation of $\fwmarg{0:\enddiff}$ in \eqref{eq:backward_denoising} from a parametrized family $\mathcal{F} \eqdef \{\bwmarg{0:T}[\param] \in \pset{1}[\rset^\statedim] | \param \in \paramset\}$ of distributions over $(\rset^{\statedim})^{T+1}$, where each $\bwmarg{0:T}[\param]\in\mc{F}$ can be decomposed as
\begin{equation}
	\bwmargd{0:T}[\param]{\chunk{\lstate}{0}{T}} \eqdef \gauss(\lstate_T; \bwmean_T, \bwstd_T^2\idm) \prod\nolimits_{t=0}^{T-1} \gauss(\lstate_t; \bwmean_{t, \param}(\lstate_{t+1}), \bwstd_t^2 \idm), 
\end{equation}
with $\bwstd_t >0$ and, for each $t$, $\mu_{t, \param} : \rset^{\statedim} \rightarrow \rset^\statedim$ is a neural network. To do so, DGMs seek to minimize the variational inference objective
\begin{equation}
	\label{eq:vi_dgm}
	\kl{\fwmarg{0:T}}{\bwmarg{0:T}[\param]} = \kl{\fwmarg{T}}{\bwmarg{T}} + \sum\nolimits_{t=0}^{T-1} \PE{}{\kl{\fwtransd{t+1}{t}{\state_{t+1}}{\cdot}}{\bwtransd{t+1}{t}{\state_{t+1}}{\cdot}}} \eqsp.
\end{equation}

Following \cite{ho2020denoising}, we consider (see \Cref{app:dgm} for a detailed derivation)  for $t \in \intvect{0}{T-1}$,
\begin{equation}\label{eq:rel_nn_mean}
	\mu_{t, \param}(\lstate_{t+1}) = \dnet{\lstate_{t+1}, \vestd_{t+1}} + ({\vestd_{t}^2}/{\vestd_{t+1}^2}){(\lstate_{t+1} - \dnet{\lstate_{t+1}, \vestd_{t+1}})} \eqsp,
\end{equation}
and we take $\bwstd_t = ({\vestd_t}/{\vestd_{t+1}})\sqrt{\vestd^2_{t+1} - \vestd^2_t}$ and $\bwstd_T = \sqrt{\vestd_T^2 + 1}$, $\bwmean_T = 0$, where
$\dnet{\cdot, \cdot} : \rset^{\statedim} \times \rset \rightarrow \rset^{\statedim}$ is taken as a neural network which is trained to minimize jointly $\{\mseloss{\dnet{\cdot, \vestd_{t}}}{\vestd_t}\}_{t=1}^{T}$. 
Note that since for any fixed $t$ the minimizer of $\mseloss{\dnet{\cdot, \vestd_{t}}}{\vestd_t}$ is a Gaussian denoiser as defined in \Cref{sec:gauss_denois}, $\dnet{\cdot, \vestd_t}$ can be seen as an neural approximation of the Gaussian denoiser of $X_t$.

Note that the minimization of \eqref{eq:vi_dgm} is related to learning the score of the marginal distributions $\fwmarg{t}$, as it can be shown  that 
$\PE{}{\state_0|\state_t = \lstate_t} = \lstate_t + \vestd_t^2 \nabla \log \fwmargd{t}{\lstate_{t}}$ \cite{vincent2011connection}.
Since $\dnet{\state_t, \vestd_t}$ is trained to approximate $\PE{}{\state_0|\state_t = \state_t}$ when $\state_t \sim \fwmarg{t}$,
$\vestd_t^{-2}(\dnet{\state_t, \vestd_t} - \state_t)$ approximates $\nabla \log \fwmargd{t}{\state_{t}}$.
% \begin{remark}
	% 	The trained neural network $\dnet{\cdot, \cdot}$ introduced above can  be used to sample $(\state_0,\dots,\state_T)$ using the variational approximation $\bwmargd{0:T}[\param]{\cdot}$ of the joint law $\fwmarg{0:\enddiff}(\cdot)$: first, we sample $\state_T \sim \gauss(0, \vestd_T^2)$, and then we iteratively apply $\dnet{\cdot, \cdot}$ in~\eqref{eq:rel_nn_mean} to sample $\state_{t} \sim \gauss( \mu_{t, \param}(\state_{t+1}), \gamma_{t}^2 \idm)$ until we reach $\state_0$.
	% \end{remark}
\begin{remark}
	While we focus our presentation of DGM on the formulation of \cite{ho2020denoising}, there are several other frameworks (such as \cite{song2021denoising} and \cite{karras2022elucidating}) that solely rely 
	in jointly minimizing $\{\mseloss{\dnet{\cdot, \vestd_{t}}}{\vestd_t}\}_{t=1}^{T}$. Other formulations are used in the numerical part, but we refer the reader to \cite{yang2023diffusion} for a general overview of the different frameworks and the links between them.
\end{remark}

\subsection{Solving Bayesian inverse problems with DGM prior}
\label{sec:background:invproblem}
When using a pre-trained DGM as prior\footnote{We omit $\param$ from the notation as the DGM is pre-trained.}, the extended posterior distribution is defined by %as $\ppostd{\chunk{\lstate}{0}{T}}\propto \lik{\lstate_0}\fwmargd{0:T}{\chunk{\lstate}{0}{T}}$ can be approximated by
% \begin{equation}
%     \label{eq:ext_posterior}
%     \ppostd{\chunk{\lstate}{0}{T}} \propto \lik{\lstate_0}\bwmargd{0:T}[]{\chunk{\lstate}{0}{T}} \eqsp,
% \end{equation}
$\ppostd{\chunk{\lstate}{0}{T}} \propto \lik{\lstate_0}\bwmargd{0:T}[]{\chunk{\lstate}{0}{T}}$.
This distribution  admits also a backward decomposition
\begin{equation}
    \label{eq:ext_posterior_bwd}
     \ppostd{\chunk{\lstate}{0}{T}} \propto \bwmargd{T}[\lmeas]{\lstate_T}  \prod\nolimits_{t=0}^{T-1} \bwtransd{t+1}{t}[\lmeas]{\lstate_{t+1}}{\lstate_t} \eqsp,
\end{equation}
where $\bwtransd{t+1}{t}[\lmeas]{\lstate_{t+1}}{\lstate_t} \propto \lik{\lstate_t}[\lmeas][t] \bwtransd{t+1}{t}[]{\lstate_{t+1}}{\lstate_t}$ with $\lik{\lstate_t}[\lmeas][t]\eqdef\int \lik{\lstate_0}[\lmeas]\bwtransd{t}{0}[]{\lstate_t}{\lstate_0} \rmd \lstate_0$.
Posterior sampling with DGM (also called training-free guidance) consists in approximately sampling from \eqref{eq:ext_posterior_bwd} without retraining the original DGM network.
This can be done by either deriving tractable approximations of $\bwtransd{t+1}{t}[\lmeas]{\lstate_{t+1}}{\lstate_t}$ \citep{chung2023diffusion,song2023pseudoinverseguided,janati2024divideandconquer,moufad2024variationala}, or by deriving asymptotically exact samplers of~\eqref{eq:ext_posterior_bwd}
 using Langevin \citep{zhang2024improving} or sequential Monte Carlo methods \citep{cardoso2024monte, wu2023practical,kelvinius2025solving}.
\section{Related Works}
\label{sec:related_works}
An alternative way to define locally anisotropic GRFs, widely used in applications, is the non-stationary covariance kernel proposed by \cite{paciorek2006spatial}. We favored the SPDE approach described in \Cref{sec:grf_spde} as it yields faster sampling algorithms and allows us to derive explicit optimal error bounds as shown in \Cref{sec:lin_den_fixed}.
When it comes to computing PPDs based on locally anisotropic GRFs, most works focus on deriving scalable methods for (frequentist) parameter estimation (see eg. \cite{li2019efficient,beckman2023scalable,huang2025nonstationary}). 
Such approaches fail to account for uncertainties on the model parameters. 
In their work, \cite{sang2012full} use a fully Bayesian approach using MCMC, but with severe restrictions on the covariance model (namely tapering and a limitation on the anisotropy variability across space).
To make these computations more amenable when dealing with non-stationary GRFs, \cite{risser2020bayesian} propose to use sparse Vecchia approximations of GRFs \cite{katzfuss2021general}.
But, as noted by these authors (and confirmed by our numerical experiments, cf. \Cref{app:num:vecchia}) this approach does not scale well for cases with more than a few thousands observations.

\cite{semenova2022priorvae} proposes a framework similar to ours, namely to use a variational autoencoder (VAE \cite{kingma2014autoencoding}) to learn the spatial distribution from a Besag-York-Mollié Gaussian process (BYM) \cite{besag1991bayesian}, which is later used for inference of the PPD.
As noted in \cite[Figure 1]{semenova2022priorvae}, the proposed approach, while scaling favorably for inference, still yields a rather different prior than the starting model.
This second issue is eased in \cite{semenova2023priorcvae} where the VAE is conditioned on hyperparameters of the stochastic process, but still in much simpler models.
Therefore, this work can be seen as a considerable extension of the framework proposed in \cite{semenova2022priorvae}, by first considering DGM instead of VAE and also more complex Gaussian process models than the BYM model. 
% \cite{li2024learning} also applies the idea of capturing a physical prior using a DGM that is used as a prior for inverse problem solving, but for spatiotemporal data. In \cite{li2024learning} the physical model comes from
% PDEs from fluid-mechanics. While promising, \cite{li2024learning} does not provide a Bayesian analysis of the PPD for the proposed model and rely only on reconstruction metrics.
\section{Numerical investigation}
\subsection{Definition of the GRF prior}
\label{sec:num:data}
We build a prior for centered locally anisotropic GRFs based on the approach described  in \Cref{sec:grf_spde}. The parameters $\adir$ is modeled as the gradient of a function $f$ (scaled to be unit-norm), which in turn is modeled  using a thin-plate spline interpolation based on 36 equidistant nodes in $\cD$. The value at each node is drawn independently from $\gauss(0, 1)$. The parameter $a$ is drawn from a $\mathcal{U}([0.05,0.3])$ distribution to ensure that the correlation length of the GRF does not exceed a third of the size of the domain. The parameters $\ascl_1, \ascl_2$ are taken to be constant across $\cD$, and drawn such that $\max \lbrace \ascl_1, \ascl_2\rbrace = 1$  and $\min \lbrace \ascl_1, \ascl_2\rbrace \sim \mathcal{U}([0.1,1])$. The parameter $\nu$ is kept constant, at a value $\nu=2$ (to get differentiable GRFs). The marginal variance of the GRF is set to $1$.

We create a dataset consisting of 300,000 simulations of  the GRFs. $\gZfem$ is built by repeating the following steps: first the parameters $\adir, \ascl_1, \ascl_2, a$  and $\nu$ are drawn as described above, and $5$ samples of the resulting GRF are drawn.
As our choice of prior distribution may seem subjective, we tested the ability of our trained generative model to adapt to other priors through fine tuning on a dataset generated by a different prior over the parameter space. This is presented in \Cref{app:num:fine_tuning}.
\subsection{Training}
\label{sec:num:training}
We have adapted the training procedure and architectures proposed in \cite{karras2024analyzing}\footnote{Code for all experiments available at \url{\githublink}}. We have done two main adaptations: adapting the size and number of the Unet layers' to obtain a deeper network but with a smaller memory footprint to suit our hardware environment (see details in \Cref{app:num:architecture} and \Cref{app:num:hardware})
and changing the sampling function used in the loss (see \cite[Section 5]{karras2022elucidating} or \cite[Section B]{karras2024analyzing}). The full details are given in \Cref{app:num:architecture}.
All the training was done using 8 Nvidia V1000 GPUs for a total of 80 epochs with batch size 2048 and learning rate scheduling as per \cite{karras2024analyzing}.
We used $250000$ data points for training with a $(5\%, 95\%)$ split between cross-validation and training.
\subsection{Evaluating the generative model}
\label{sec:num:uncond_gen}
In this section, we evaluate how well the generative model captures the target distribution.
%While those metrics are extremely pertinent in computer vision, as they are related to human perception, in our case, it is important to guarantee statistical properties of the distribution. 
Following \cite{bischoff2024practical}, we rely on statistical pseudo-metrics, the maximum sliced-Wasserstein (\maxsw) \cite{nietert2022statistical} and the classifier two-sample test (C2ST) \cite{lopez-paz2017revisiting}.
We highlight the word "pseudo-metric" because both are not true metrics, although they are related (asymptotically) to statistical metrics.

Given two sets of samples $\dataset_{1}$ and $\dataset_{2}$ from distributions $\mu_1$ and $\mu_2$, the {\maxsw} corresponds to the maximum, over a large number of (uniformly drawn) directions, of the 1-d Wasserstein distance between the projected samples of the two sets.
Note that for general distributions, two sets of independently drawn samples can have a non-zero {\maxsw}. \citet{nietert2022statistical} establishes concentration bounds for this estimator around the true Maximum Wasserstein metric.
{\c2st} is also applied to two sets of samples $\dataset_{1}$ and $\dataset_{2}$.
Each set is divided into train $\dataset_{1, \operatorname{train}}, \dataset_{2, \operatorname{train}}$ and test $\dataset_{1, \operatorname{test}}, \dataset_{2, \operatorname{test}}$. 
A classifier is trained to distinguish between $\dataset_{1, \operatorname{train}}$ and $ \dataset_{2, \operatorname{train}}$.
It is then evaluated over the test set consisting of $\dataset_{1, \operatorname{test}}$ and $\dataset_{2, \operatorname{test}}$.
The lack of performance on the test task indicates that $\mu_ 1 \approx \mu_2$.

Formally, under the hypothesis that the class of functions being used to construct the classifier is able to approximate the Bayes classifier, it is possible to derive an asymptotic two-sample test with null hypothesis being that the $\mu_1 = \mu_2$ \cite{lopez-paz2017revisiting}.
While {\c2st} has performed extremely well in several applications \cite{linhart2023lc2st,lueckmann2021benchmarking,bischoff2024practical}, specially for high dimensional datasets, when dealing with pixel space classifiers, they are known
to be extremely sensitive. %In \cite{lopez-paz2017revisiting}, the authors show that neural networks often are able to accurately distinguish (accuracy $>90\%$) between GAN generated images and real images. This happens even for models achieving considerably low perceptual metrics like FID.

For generation, we rely on three samplers, two deterministic: an ODE-based Heun sampler (\edm) introduced in \cite{karras2022elucidating} and the DDIM sampler (\ddim) from \cite{song2021denoising}, and a stochastic sampler (\ddpm) from \cite{ho2020denoising} and presented in \Cref{sec:background:dgm}.
For each sampler configuration we draw $50000$ and compare via both metrics to a held-out dataset of $50000$ draws from the data distribution.
For the \maxsw, we use a total of $2^{16}$ slices and draw $10000$ random samples from the pool of available samples from both the generated and the validation data.
%We repeat this draw several times and report mean and CLT confidence intervals.
For \c2st, both the generated data and held-out dataset into train and test sets have $25000$ on each partition. %We draw five different partitions for each combination of classifier and sampler.
The details of the training for \c2st are given in \Cref{app:num:c2st}.%
\paragraph{Denoiser performance:}
\begin{wrapfigure}[19]{r}[0pt]{0.4\linewidth}
    \vspace{-14pt}
    \begin{tabular}{
        M{1\linewidth}@{\hspace{0\tabcolsep}}}
        \includegraphics[width=\hsize]{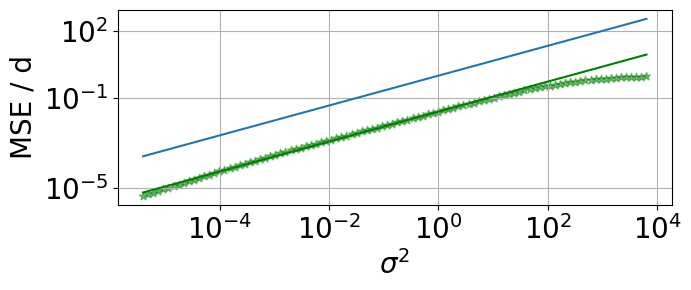}\\\addlinespace[-2pt]
        \includegraphics[width=\hsize]{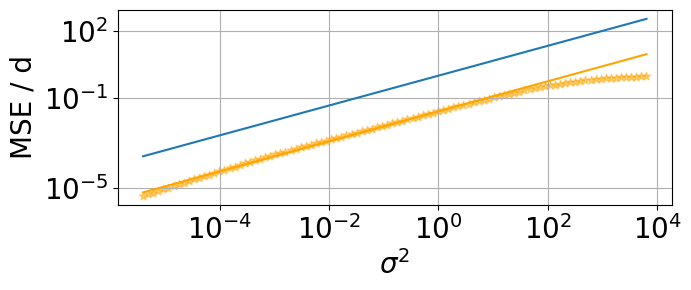}
    \end{tabular}
    \caption{MSE vs $\sigma^2$ for anisotropic data showing both the optimal slope of \eqref{eq:background:optimal_denoising} in blue and the performance of the trained model on both training (green, top) and validation (orange, bottom) datasets. The MSE was calculated by a Monte Carlo estimate with $640$ samples.}
    \label{fig:mse_vs_sigma}
\end{wrapfigure} The first experiment goal is to test the performance of the denoiser. Following \cite{kadkhodaie2023generalization} and considering \eqref{eq:background:optimal_denoising}, we investigate
the variation of $\mseloss{\dnet{\cdot, \vestd}}{\vestd}$ with $\vestd^2$ over the training and validation datasets. The results are shown in \Cref{fig:mse_vs_sigma}.
There are two main conclusions that can be drawn from the results from \Cref{fig:mse_vs_sigma}. Firstly, that the trained denoiser has the same slope as the optimal denoiser defined in \eqref{eq:background:optimal_denoising}.
Secondly, the slope is the same in both train and validation datasets, indicating that the denoiser generalizes well.
\paragraph{Choice of scheduler:}
We focus on the choice of scheduler for the generation of samples. We follow the choice of parametrization from \cite{karras2022elucidating} for a given number of steps $N$ and a shape parameter $\rho$
sets $\vestd_{t_i} = \{\vestd_{T}^{1/\rho} + \left[1 - i/(N-1)\right] (\vestd_{T}^{1/\rho} - \vestd_{0}^{1/\rho})\}^{\rho}$.
In \cite[Appendix D1]{karras2022elucidating} the authors show that $\rho=3$ minimizes the discretization error for the Heun sampler but recommend using $\rho=7$ for better image quality (based on FID).
We calculated the {\maxsw} for both a deterministic (\ddim) and a stochastic sampler (\ddpm) with $N=100$ for several choices of $\rho$. We obtain that indeed $\rho=3$ performs best in both cases (cf. \Cref{app:additional_exps:rho}).
\paragraph{Generative results:}
\begin{table}
    \begin{subtable}[h]{0.42\textwidth}
        \begin{tabular}{|c|c|c|}%
            \hline
            Sampler & N steps & \maxsw  \\
            \hline
            \DTLforeach*{max_sws}{
                \sampler=sampler,
                \nstepsw=n_steps,
                \maxswmean=mean,
                \maxswstd=std}%
                {%
                \sampler & \nstepsw & \maxswmean \quad (\maxswstd) \DTLiflastrow{}{\\}
                }
            \\\hline
        \end{tabular}%  
    \end{subtable}%
    \hspace{12pt}
    \begin{subtable}[h]{0.42\textwidth}
        \begin{tabular}{|c|c|c|c|}%
            \hline
            Sampler & N steps & Network & \c2st  \\
            \hline
            \DTLforeach*{classif_table}{
                \sampler=sampler_name,
                \nsteps=n_steps, \network=network,\accmean=acc-mean, \accstd=acc-std}%
                {%
                \sampler & \nsteps &\network & \accmean \quad (\accstd)
                \DTLiflastrow{}{\\}}%
            \\\hline
        \end{tabular}%        
    \end{subtable}%
\caption{Results of the {\maxsw} and {\c2st} metrics on the form "mean (standard deviation)". For {\maxsw} the replicates correspond to 20 different slices and samples draws. For {\c2st} they correspond to 5 different train / test splits of the datasets. The train value on {\maxsw} correspond to the {\maxsw} between train and test samples.}
\label{table:aniso_gen}
\vspace{-18pt}
\end{table}
We proceed to an evaluation of the quality of the generated samples with $\rho=3$. \Cref{table:aniso_gen} shows the results of the \c2st and {\maxsw} for several classifiers architectures and several samplers.
For the \c2st statistic from \Cref{table:aniso_gen}, the cutoff corresponding to a $5\%$ p-value would be at $\approx 0.502$, thus, one could safely reject the hypothesis that the two distributions are equal.

There are two things to keep in mind: First that those tests presuppose that a classifier is able to reach the Bayes classifier and second that the {\c2st} obtained here is extremely strong compared to the existing literature (See \cite[]{lopez-paz2017revisiting} or \cite[]{bischoff2024practical}).
For the first point, note that as the capacity of the classifier increases, the test statistic decreases. As for the second, the classifier is barely able to distinguish between datasets, thus suggesting that while not exactly the same the two distributions must be close (see \Cref{app:additional_exps:c2st} for confusion matrices and roc curve examples).
%We then look into the eigenvalues of $\nabla {\dnet{\cdot, \vestd}}$ \gabriel{On fait dans le main ou l'appendix?}.
\subsection{Posterior sampling with DGM}
\label{sec:num:cond_gen}
\paragraph{Choice of DGM posterior:}
\begin{filecontents*}{data/max_sws_posterior.csv}
,daps-400-0-mean,dps-1000-0-mean,mgdm-100-1-mean,mgdm-100-2-mean,mgps-100-0-mean,mgps-300-0-mean,img_id,size_id,mask_mode,n_points,daps-400-0-std,dps-1000-0-std,mgdm-100-1-std,mgdm-100-2-std,mgps-100-0-std,mgps-300-0-std
0,1.65,0.71,0.64,0.54,0.62,0.53,0,2,clust,364,0.08,0.04,0.04,0.03,0.03,0.03
1,2.03,1.16,1.38,0.83,0.78,0.70,0,2,unif,300,0.14,0.06,0.11,0.06,0.04,0.04
2,1.88,0.71,0.99,0.66,0.90,0.78,1,2,clust,229,0.12,0.04,0.06,0.03,0.04,0.05
3,2.74,0.75,1.10,0.51,0.35,0.32,1,2,unif,300,0.16,0.03,0.07,0.03,0.02,0.02
4,1.36,0.94,0.83,0.74,0.96,0.79,1,3,clust,355,0.05,0.05,0.04,0.05,0.09,0.06
5,2.36,0.43,0.81,0.30,0.35,0.34,1,3,unif,600,0.13,0.02,0.04,0.02,0.02,0.02
6,1.93,0.78,0.92,0.59,0.74,0.70,4,3,clust,612,0.08,0.06,0.04,0.03,0.05,0.04
7,2.54,0.49,0.85,0.35,0.47,0.52,4,3,unif,600,0.15,0.02,0.07,0.02,0.03,0.04
\end{filecontents*}
\DTLloaddb{max_sws_post}{data/max_sws_posterior.csv}
\begin{table}[]
    \centering
        \begin{tabular}{|c|c|c|c|c|c|}%
            \hline
                $\measdim$ & index & type & \dps \citep{chung2023diffusion} &  \mgdm\citep{moufad2024variationala} &  \mgps \citep{janati2024divideandconquer} \\
            \hline
            \DTLforeach*{max_sws_post}{
                \ydim=n_points,
                \index=img_id,
                \masktype=mask_mode,
                \dpsval=dps-1000-0-mean,
                \mgdmvalfine=mgdm-100-2-mean,
                \mgpsvalfine=mgps-300-0-mean,
                \dapsval=daps-400-0-mean,
                \dpsstd=dps-1000-0-std,
                \mgdmvalfinestd=mgdm-100-2-std,
                \mgpsvalfinestd=mgps-300-0-std,
                \dapsstd=daps-400-0-std}%
                {%
                \ydim & \index & \masktype &\dpsval \quad (\dpsstd) & \mgdmvalfine \quad (\mgdmvalfinestd) & \mgpsvalfine  \quad (\mgpsvalfinestd) \DTLiflastrow{}{\\}
                }
                \\\hline
        \end{tabular}%
    \caption{{\maxsw} between MCMC and different DGM posterior sampling algorithms for different inpainting inverse problems (see \Cref{app:additional_exps:posteriorsampling} for details) in the form "mean (standard deviation)". A total of $2^{16}$ slices were used. Quantities where aggregated over 20 different slices and different $10^4$ subsets draws from the pool of available generated samples ($2\times 10^4$). Implementation details and runtime for the DGM posterior sampling algorithms are given in \Cref{app:num:post_sampling_details}.}
    \label{table:posterior_sampling}
\end{table}%
%To test different DGM posterior sampling methods, we first apply it to a simpler setting for which MCMC is tractable.
{We consider three possible DGM posterior sampling methods:  \dps \citep{chung2023diffusion}, \mgdm \citep{moufad2024variationala}, amd \mgps \citep{janati2024divideandconquer}, which we evaluate on inverse problems based on a simpler GRF prior, for which MCMC posterior sampling is tractable.}
To do so, we fine-tuned  our DGM prior (cf. \Cref{app:num:fine_tuning}) to a case where the parameter space consists of only 3 scalar parameters: the range $a$, the anisotropy ratio $\min\lbrace\rho_1, \rho_2\rbrace/\max\lbrace\rho_1, \rho_2\rbrace$ and an angle $\theta$ parametrizing the unique (global) direction of correlation of the GRF.
We focus on severely ill-posed inpainting problems, as they are often encountered in the environmental sciences. We generate 8 different inpainting inverse problems by changing the initial image, the pattern of the observation points (uniform across the domain or clustered) and the number of observations.

We use a Random Walk Metropolis Hastings MCMC (MH-MCMC) algorithm to generate $10^4$ independent chains of length $2.5\times 10^3$ to generate the reference samples. 
Only the last element of each chain was kept to avoid correlation. 
We then compare different state-of-the-art DGM posterior sampling algorithms using the {\maxsw} to those MCMC samples
The results are shown in \Cref{table:posterior_sampling} and samples from configuration are displayed in \Cref{app:additional_exps}. As the {\mgdm} systematically outperforms {\dps} and {\mgps}, we only use this method for the rest of our numerical experiments.%
\paragraph{Illustration on simulated data:}We apply {\mgdm} for different inverse problems using the \q{full} DGM prior from \Cref{sec:num:uncond_gen}.
We draw a "true" sample from the test set, from which we obtain a realization of \eqref{eq:measurement_equation} for each different inverse problem.
The results are shown in \Cref{fig:inv_problem_illustration} and in \Cref{app:additional_exps}. We see that the {\mgdm} is able to accurately capture
the anisotropies of the underlying process even with a considerably small number of observation points.
\paragraph{Application to sea surface temperature anomaly data:}
\begin{figure}
    \centering
        \begin{tabular}{
            M{0.16\linewidth}@{\hspace{0.1\tabcolsep}}
            M{0.16\linewidth}@{\hspace{0.1\tabcolsep}}
            M{0.16\linewidth}@{\hspace{0.1\tabcolsep}}
            M{0.16\linewidth}@{\hspace{0.1\tabcolsep}}
            }
            Observation & {\mgdm} &  \vecchia &  \priorvae  \\
            \includegraphics[width=\hsize]{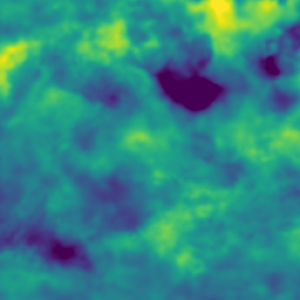} 
            & \includegraphics[width=\hsize]{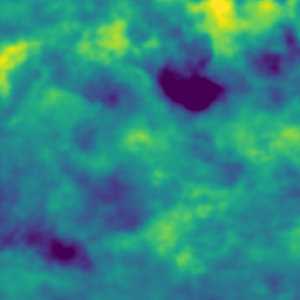} 
            & \includegraphics[width=\hsize]{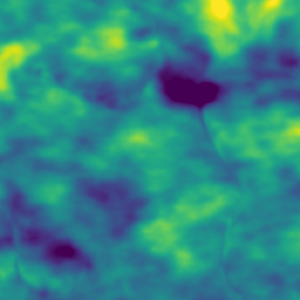} 
            & \includegraphics[width=\hsize]{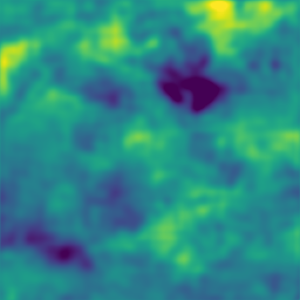}\\\addlinespace[-2.6pt]
            \includegraphics[width=\hsize]{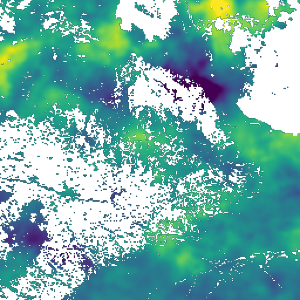} 
            & \includegraphics[width=\hsize]{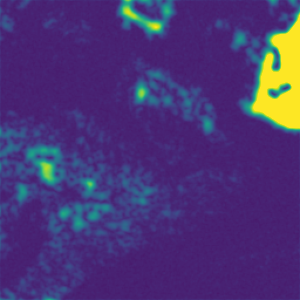} 
            & \includegraphics[width=\hsize]{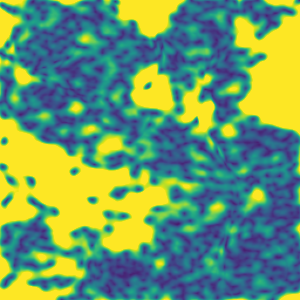} 
            & \includegraphics[width=\hsize]{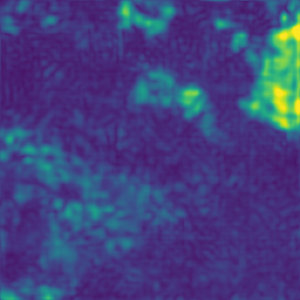}
        \end{tabular}%
    \caption{Illustration of different posterior sampling methods on the SSTA problem for case 1 from \Cref{table:rec_cloud}.
    First column show the data (top) and observation (bottom, with cloud). The other columns correspond to a sample (top) and the standard deviation obtained over $100$ replicates for each method.
    The colors are normalized between $-3$ (blue) and $3$ (yellow) and $0$ (blue) and $0.3$ (yellow) for top and bottom rows respectively.}
    \label{fig:inv_problem_cloud:1}
\end{figure}%
Inspired by \cite{beckman2023scalable}, we consider the problem of reconstructing the sea surface temperature anomalies (SSTA) from partial observations.
We focus on the case where the partial observations are due to the presence of clouds, which provide a natural inpainting mask. This problem is common in QOI that are measured through satellite imaging.
The SSTA data are extracted from the NOAA Coral Reef Watch database \cite{noaa2019coral}, and corresponds to SSTA observed on different parts of the globe (cf. \Cref{app:num:sea_temperature_details}). 
%coast of South Africa.
The cloud mask is extracted from NASA’s MODIS/Aqua Cloud Mask product \cite{MODIS_Atmosphere_Science_Team2017}. We extract three pairs of (observation, clouds).%and correspond to the sky coverage.

We set the observation noise to $\sigma_y = 0.05$ and use the prior proposed above with {\mgdm} as sampler.
We compare our results to posterior samples from a MH-MCMC algorithm based on a Vecchia approximation of our GRF prior ({\vecchia}) with $10^3$ subsampled observations, and to posterior samples from the  {\priorvae} approach trained on data from the same prior (cf. \Cref{app:num:prior_vae,app:num:vecchia}). For each sampler, 100 samples are generated.
The results are shown in \Cref{fig:inv_problem_cloud:1}.
Since a reference Bayesian posterior is unavailable, we rely on the Continuous Ranked Probability Score (CRPS) to evaluate the PPDs (cf. \Cref{app:scores,app:additional_exps:SSTA} for details and additional metrics). 
The results are displayed in \Cref{table:rec_cloud} and show that {\mgdm} outperforms significantly the other methods on the three inverse problems considered in the experiment.
\begin{table}
	\centering
		\begin{tabular}{|c|c|c|c|}%
			\hline
			 Case & \vecchia & \priorvae & \mgdm \\
			\hline
			\DTLforeach*{crps_temp}{
				\index=img_id,
				\vec=vecchia,
				\vae=priorvae,
				\mgd=mgdm}%
			{%
			 \index & \vec &\vae & \mgd  \DTLiflastrow{}{\\}
			}
			\\\hline
		\end{tabular}%
	\caption{CRPS on the SSTA problem for three cases, in the form \q{mean (standard deviation)} (lower is better). The unobserved locations are randomly separated into 32 disjoint subsets, on which the average CRPS is computed. The mean and standard deviation of these values are shown above.}
	\label{table:rec_cloud}
    \vspace{-18pt}
\end{table}%
\subsection{Conclusion}
In this work, we show that DGMs offer a viable solution to PPD sampling with non-stationary GRF priors.
We show it outperforms existing approximation methods in statistical quality (CRPS) while being much more scalable (once the DGM prior is trained).
We show the potential of a generalized use of such complex GRF priors as agnostic priors for real world problems, as they allow for straightforward and scalable spatial predictions accounting for uncertainty in the prior.
\subsection{Limitations}\label{sec:limits}
In this work, we only considered GRF priors and posteriors discretized over a regular grid of fixed size. 
A natural extension is to allow the GRF priors (and posteriors) to be defined continuously in space by following the approach of \cite{chen2024image}.
Another approach is to consider the discretized fields generated by the DGMs as a discretization of "continuous" GRFs through a finite element approach (cf. \Cref{app:grf_spde}), thus allowing the value of the field at any spatial location to be computed as a linear combination of the pixel values.
This straightforward extension would directly fit into our framework as it can be cast as a special choice of measurement equation~\eqref{eq:measurement_equation}.

We only considered centered GRF priors for our inverse problems, and considered fixed the variance of the measurement noise.
Including non-zero means and inferring the noise level could be done using an expectation maximization (EM) approach in the same way as in \cite[Section 3]{bedin2024leveraging}.  %\mikec{GC: Please complete} 
Besides, the regularity parameter $\nu$ of GRFs was also fixed. This parameter is often fixed by the practitioners, even though its correct determination, though challenging, is paramount \cite{de2022information}.
One could include the regularity parameter in the GRF prior using for instance the priors proposed by \cite{han2024default} and train a conditional DGM. %We leave those research lines for future work. 

For applications where the values of the underlying parameters are important, one could re-identify them via a Maximum likelihood estimation for each posterior sample.

Finally, considerable resources were needed to train the DGMs (cf \Cref{app:num:impact}).%, which can be an obstacle for users searching to reimplement the method from scratch.
Our work can however be seen as first step towards a DGM-based Bayesian prior for spatial data, which could be built as a common work by the spatial statistics community, and used as an off-the-shelf method by practitioners.

\subsection{Acknowledgments:}
The authors acknowledge the financial support of the chair Geolearning, funded by ANDRA, BNP Paribas, CCR and the SCOR Foundation for Science.
\newpage
\bibliographystyle{abbrvnat}
\bibliography{references}

\clearpage
\doparttoc % Tell to minitoc to generate a toc for the parts
\faketableofcontents % Run a fake tableofcontents command for the partocs

\part{} % Start the document part% Insert the document TOC
\appendix
\addcontentsline{toc}{section}{Appendix} % Add the appendix text to the document TOC
\part{Appendix} % Start the appendix part
\parttoc % Insert the appendix TOC
\clearpage
\section{Finite element discretization of random fields}\label{app:grf_spde}

We consider a discretization of $\cD$ consisting of a grid of $\statedim = 256\times 256$ nodes, upon which a triangulation of $\cD$ is defined. Let  $\lbrace \femb_k\rbrace_{1\le k \le \statedim}$ be the linear finite element basis associated with this triangulation (meaning that $\femb_k$ is the piecewise linear function taking the value $1$ at node $k$ and $0$ at all the other nodes). Following \cite{pereira2022geostatistics} we combine  a Galerkin approximation of the Laplace--Beltrami operator $\lbo$ with a mass lumping approximation to obtain the following closed-form  for the finite element approximation $\gZ$ of the field $\cZ$ defined in~\eqref{eq:def_Z}, thus giving
\begin{equation}
	\gZ = \sum_{k=1}^{\statedim} \gZfem_k \femb_k,
\end{equation}
where  $\gZfem=(\gZfem_1,\dots,\gZfem_\statedim)$ forms a centered Gaussian vector with covariance matrix
\begin{equation}
	\gZcov = \Mmat^{-1/2}\psd^2(\Smat)\Mmat^{-1/2}
\end{equation}
where $\Mmat \in\rset^{\statedim\times\statedim}$ is the diagonal mass-lumped matrix with entries 
\begin{equation}
	[\Mmat]_{ii}=~\funcsp{\femb_i}{1}=\int_{\cD} \femb_i(\ptD) \sqrt{\vert \det \bm G_{\ptD}\vert} d{\ptD}
\end{equation}
and we denote by $\bm G_{\ptD}=\bm Q_{\ptD}^T\bm Q_{\ptD}$ the metric tensor at $s\in\mathcal{D}$. The matrix $\Smat\in \rset^{\statedim\times\statedim}$ is the scaled stiffness matrix defined as $\Smat=\Mmat^{-1/2}\Rmat \Mmat^{-1/2}$ with $\Rmat\in\rset^{\statedim\times\statedim}$ being the (stiffness) matrix with entries 
\begin{equation}
	[\Rmat]_{ij}=\funcsp{\grad\femb_i}{ \grad\femb_j}=\int_{\cD} (\grad\femb_i(\ptD))^T \bm G_{\ptD}^{-1} \grad\femb_j(\ptD) \sqrt{\vert \det \bm G_{\ptD}\vert} d{\ptD}
\end{equation}
Note in particular the inner-products account the local changes of metric across the manifold.  As for the matrix function $\psd^2(\Smat)$, it is obtained by applying the function $\psd(\cdot)^2$ to the eigenvalues of $\Smat$, while keeping the corresponding eigenvectors intact. 

Note that since we consider linear finite elements, $\gZfem_k$ actually corresponds to the value of the field $\gZ$ at the $k$-th discretization node of $\cD$. Hence, over the discretization nodes of $\cD$, the field $\gZ$ is entirely determined by the vector of weights $\gZfem=(\gZfem_1,\dots,\gZfem_\statedim)$. Therefore, we from now on focus only this (Gaussian) vector. In particular,
a reparametrization trick allows to rewrite any sample $\gZfem\sim\mathcal{N}(0,\gZcov)$ as
\begin{equation}\label{eq:def_vectZ}
	\gZfem={\Mmat}^{-1/2} \psd(\Smat) \rnorm, \quad \rnorm\sim\gauss(0, \idm_\statedim).
\end{equation}
This last expression is used to sample $\gZfem$. For computational purposes, the product $\psd(\Smat) \rnorm$ in~\eqref{eq:def_vectZ} can be approximated by the product $P_\psd(\Smat) \rnorm$ where $P_\psd$ is a polynomial approximation of $\psd$ over an interval containing the eigenvalues of $\Smat$, thus avoiding the need to know the eigenvalues and eigenvectors of $\Smat$ (required in the definition of the matrix function $\psd(\Smat)$). 

Finally note that, as precaution, we applied the approach outline above on a slight expanded domain $\cDext = [-0.1, 1.1]^2\supset \cD$ to mitigate the effect of the boundary condition that need to be imposed on the simulation domain. The discretization we used consists of $320 \times 320$ nodes over $\cDext$ and is picked so that $\cD$ is indeed discretized by a mesh with $256 \times 256$: in essence, the actual samples $X$ over $\cD$ are in practice subvectors of the samples generated by the finite element approach, which only marginally changes the rest of the arguments in the paper.

\section{DGM: Additional details and derivations}
\label{app:dgm}
\subsection{Derivation of \eqref{eq:vi_dgm}}

Throughout this section we assume $\PE{}{\state_0} = 0$ and $\PV{}{\state_0} = \idm$. We first start by rewritting \eqref{eq:vi_dgm} as:
\begin{align}
    &\kl{\fwmarg{0:T}}{\bwmarg{0:T}[\param]} = \PE{}{\kl{\pdata}{\bwtransd{1}{0}[\param]{\state_1}{\cdot}}} \\
    &\qquad + \sum_{t=1}^{T-1} \PE{}{\kl{\fwtransd{t+1}{t}{\state_{t+1}}{\cdot}}{\bwtransd{t+1}{t}[\param]{\state_{t+1}}{\cdot}}} + \kl{\fwmarg{T}}{\bwmarg{T}} {+C} \eqsp.   \label{eq:vi_dgm_expanded}
\end{align}
{for some constant $C$ independent of $\param$.}
We then use the fact that for any $\lambda \in \pset{2}$ with mean $\mu_{\lambda}$ and covariance $\Sigma_{\lambda}$, 
\begin{equation}
    (\mu_\lambda, \Sigma_\lambda) \in \argmin_{\mu, \Sigma} \kl{\lambda}{\gauss(\mu, \Sigma)} \eqsp,
    \mu_\lambda \in \argmin_{\mu} \kl{\lambda}{\gauss(\mu, \Sigma)} \eqsp,
\end{equation}
for all $\Sigma$, meaning that it is enough to match the first two moments of the two distributions to minimize their KL-divergence. Hence,
we focus on the calculation of $\PE{}{\state_t | \state_{t+1}}$ and $\PV{}{\state_t | \state_{t+1}}$. 

Here is where the three aforementioned frameworks separate.
In \cite{song2021scorebased}, an expression for both terms are explicitly obtained by choosing a discretization of the backward SDE (see \cite[Eq. 6]{song2021scorebased}).
We follow \cite{ho2020denoising} and note that by Bayes law and Gaussian conjugation, the p.d.f of $\state_t | \state_{t+1}, \state_0$ is given by
\begin{align}
    \fwtransd{0, t+1}{t}{\lstate_{0}, \lstate_{t+1}}{\lstate_t} &\eqdef \gauss\left(\lstate_t; \lstate_0 + \frac{\vestd_{t}^2}{\vestd^2_{t+1}}{(\lstate_{t+1} - \lstate_0)}, \frac{\sigma_t^2 }{\sigma_{t+1}^2}(\vestd_{t+1}^2 - \vestd_{t}^2)\idm\right)\eqsp.
\end{align}
Thus, we can finally calculate
\begin{align}
    \PE{}{\state_t | \state_{t+1}} &= \PE{}{\PE{}{\state_t | \state_{t+1}, \state_{0}} | \state_{t+1}} \\
    &=  \PE{}{\state_0|\state_{t+1}}\left(1 -  \frac{\vestd_{t}^2}{\vestd_{t+1}^2}\right) +  \frac{\vestd^2_{t}}{\vestd^2_{t+1}} \state_{t+1} \eqsp, \\
    \PV{}{\state_t | \state_{t+1}} &= \PE{}{\PV{}{\state_t | \state_{t+1}, \state_0}| \state_{t+1}} + \PV{}{\PE{}{\state_t | \state_{t+1}, \state_0} | \state_{t+1}} \\
    &= \frac{\vestd_t^2}{\vestd_{t+1}^2}(\vestd_{t+1}^2 - \vestd_t^2)\idm + \left(1 -  \frac{\vestd_{t}^2}{\vestd_{t+1}^2}\right)^2 \PV{}{\state_0 | \state_{t+1}} \\
    &= \vestd_t^2\left(1 - \frac{\vestd_t^2}{\vestd_{t+1}^2}\right)\idm +  \left(1 -  \frac{\vestd_{t}^2}{\vestd_{t+1}^2}\right)^2\PV{}{\state_0 | \state_{t+1}} \eqsp.
\end{align}
Minimizing \eqref{eq:vi_dgm}  is equivalent to approximating the conditional means $\PE{}{\state_0 | \state_{t+1}}$ and $\PV{}{\state_0 | \state_{t+1}}$.
However, in the literature \cite{ho2020denoising,song2021denoising}, the term  $\PV{}{\state_0 | \state_{t+1}}$ is often neglected. 
\cite{guo2011estimation}
Indeed, by Markov inequality,
\begin{align}
    \PE{}{\indi{\tr{\PV{}{\state_0|\state_t}} \geq a^{-1}}} &\leq a \PE{}{\tr{\PV{}{\state_0 | \state_t}}} = a\PE{}{\PE{}{\|\state_0 - \PE{}{\state_0|\state_t}\|^2 |\state_t}} \\
    &\leq a \PE{}{\PE{}{\|\state_0 - \state_t\|^2 |\state_t}} = a \statedim \vestd_t^2 \eqsp.
\end{align}
In particular, with probability $1-\delta$, we have that $\tr{\PV{}{\state_0|\state_t}} \leq \statedim \vestd_t^2 / \delta$.
Therefore, this implies that with probability $1-\delta$,
\begin{equation}
    \tr{\PV{}{\state_t|\state_{t+1}} - \vestd_t^2\left(1 - \frac{\vestd_t^2}{\vestd_{t+1}^2}\right)\idm} \leq \frac{\statedim \vestd_t^2}{\delta}\left(1 - \frac{\vestd_t^2}{\vestd_{t+1}^2}\right)^2 \eqsp,
\end{equation}
showing that the error between the variance approximation and the true variance can be make arbitrarily small by an appropriate choice of scheduling.
Neglecting $\PV{}{\state_0 | \state_t}$ is particularly important in high dimensional cases, where its estimation would be costly. Therefore, following \cite{ho2020denoising}, we obtain 
\begin{equation}
    \bwtransd{t+1}{t}[\param]{\lstate_{t+1}}{\lstate_t} =
    \gauss\left(\lstate_t;  \dnet{\lstate_{t+1}, \vestd_{t+1}} + \frac{\vestd^2_{t}}{\vestd^2_{t+1}}{(\lstate_{t+1} - \dnet{\lstate_{t+1}, \vestd_{t+1}})},
    \frac{\vestd_t^2}{\vestd_{t+1}^2}(\vestd_{t+1}^2 - \vestd_{t}^2)\idm\right)\eqsp,
\end{equation}
which correspond to $\mu_{t, \param}(\lstate_{t+1}) = \dnet{\lstate_{t+1}, \vestd_{t+1}} + \frac{\vestd_{t}^2}{\vestd_{t+1}^2}{(\lstate_{t+1} - \dnet{\lstate_{t+1}, \vestd_{t+1}})}$ and 
where the Network $\dnet{\lstate_{t+1}, \vestd_{t+1}}$ is trained to jointly minimize $\{\mseloss{\dnet{\cdot, \vestd_{t}}}{\vestd_t}\}_{t=1}^{T}$.
For $t=T$, note that we obtain that $\PE{}{\state_T} = \PE{}{\PE{}{\state_T|\state_0}} = 0$ and
$\PV{}{\state_T} = \PE{}{\PV{}{\state_T|\state_0}} + \PV{}{\PE{}{\state_T | \state_0}} = (\sigma_T^2 + 1)\idm$.

While we know that jointly minimizing $\{\mseloss{\dnet{\cdot, \vestd_{t}}}{\vestd_t}\}_{t=1}^{T}$ minimizes \eqref{eq:vi_dgm} , one might estimate an upper bound of \eqref{eq:vi_dgm} via the data-processing inequality
\begin{align}
    &\PE{}{\kl{\fwtransd{t+1}{t}{\state_{t+1}}{\cdot}}{\bwtransd{t+1}{t}[\param]{\state_{t+1}}{\cdot}}} \\
    &\qquad \leq \PE{}{\kl{\fwtransd{0, t+1}{t}{\cdot, \state_{t+1}}{\cdot} \fwtransd{t+1}{0}{ \state_{t+1}}{\cdot}}{\bwtransd{t+1}{t}[\param]{\state_{t+1}}{\cdot}\fwtransd{t+1}{0}{ \state_{t+1}}{\cdot}}} \\
    &\qquad= \PE{}{\kl{\fwtransd{0, t+1}{t}{\state_0, \state_{t+1}}{\cdot}}{\bwtransd{t+1}{t}[\param]{\state_{t+1}}{\cdot}}} \\
    &\qquad= C + \frac{1}{2\bwstd_t^{2}}\PE{}{\left\|\mu_{t, \param}(\state_{t+1}) - \left( \state_0 + \frac{\vestd_{t}^2}{\vestd_{t+1}^2}(\state_{t+1} - \state_0)\right) \right\|^2} \\
    &\qquad= C + \frac{\left(1 -  \frac{\vestd_{t}^2}{\vestd^2_{t+1}}\right)^2}{2\bwstd_t^{2}}\underbrace{\PE{}{\|\dnet{\state_{t+1}, \vestd_{t+1}}- \state_0 \|^2}}_{=\mseloss{\dnet{\cdot, \vestd_{t+1}}}{\vestd_{t+1}}\eqsp.} \eqsp,
\end{align}
where $C$ is a constant independent of $\param$.

For all the other terms, we have
\begin{align}
    \PE{}{\kl{\pdata}{\bwtransd{1}{0}{\state_1}{\cdot}}}  &= C - \PE{}{\log \bwtransd{1}{0}{\state_1}{\state_0}} \\
    &= C + \frac{1}{2 \bwstd_0^2}\underbrace{\PE{}{\|\state_0 - \dnet{\state_1, \vestd_1}\|^2}}_{=\mseloss{\dnet{\cdot, \vestd_1}, \vestd_1}\eqsp.} \eqsp,
\end{align}
and 
$\PE{}{\kl{\fwmarg{T}}{\gauss(0, (\vestd_T^2 + 1)\idm)}} = \frac{1}{2(\vestd_{T}^{2} + 1)} \PE{}{\|\state_0\|^2} \eqsp.$

Therefore, leading to
\begin{align}
    \kl{\fwmarg{0:T}}{\bwmarg{0:T}[\param]} \leq \sum_{t=1}^{T} \gamma_{t}^2 \mseloss{\dnet{\cdot, \vestd_{t}}}{\vestd_t} + \frac{1}{2(\vestd_{T}^{2} + 1)} \PE{}{\|\state_0\|^2} + C \eqsp,
\end{align}
where again $C$ does not depend on $\param$ and $\gamma_{t+1}^2 =  \nofrac{\left(1 -  \frac{\vestd^2_{t}}{\vestd^2_{t+1}}\right)^2}{2\bwstd_t^{2}}$ for $t > 0$ and $\gamma_{1}^2 = (2 \bwstd_0^2)^{-1}$.

While this upper bound is a logical candidate, several other propositions of averaged losses have been used for jointly minimizing $\dnet{\lstate_{t+1}, \vestd_{t+1}}$, see for example \cite[Section 3.4]{ho2020denoising}
or \cite[Section 5]{karras2022elucidating}.

\subsection{Connection between variance preserving (VP) and variance exploding (VE) frameworks}

In this section, we show that if the \vpalt{VP} framework (see \cite{ho2020denoising}) and the VE framework presented in \Cref{sec:background:dgm} are equivalent, in the sense that the two scores are related, and knowing the score in one framework gives the score in the other.
\vpalt{VP} defines the noising process via the Markov chain
\begin{equation}
\vpalt{\state_{t} = \sqrt{1 - \beta_{t}}\state_{t-1} +\sqrt{\beta_{t}} \noise_{t}} \eqsp,
\end{equation}
where $\vpalt{\beta_{t}} \in [0, 1]$. In this case, the forward transition kernel is $\vpalt{\fwtransd{0}{t}{\lstate_0}{\lstate_t} = \gauss(\lstate_t; \sqrt{\alpha_t}\lstate_0, (1 - \alpha_t)\idm)}$ where $\vpalt{\alpha_t = \prod_{s=1}^{t}(1 - \beta_s)}$.

The key property is that if $\vpalt{(\state_t, \state_0) \sim \fwmarg{t, 0}}$ and we set $\state_{s(t)} = \vpalt{\sqrt{\alpha_t}^{-1}\state_{t}}$, then $(\state_{s(t)}, \state_0)$ is distributed according to (the VE) $\fwmarg{s, 0}$ where $s$ is such that $\vestd_{s} = \sqrt{\frac{1 - \alpha_t}{\alpha_t}}$.
In particular, for all $\lstate_t$ the conditional distributions $\vpalt{\fwtransd{t}{0}{\lstate_t}{\cdot}}$ and $\fwtransd{s}{0}{\lstate_s = \vpalt{\sqrt{\alpha_t}^{-1}} \lstate_t}{\cdot}$ are the same.

By the denoising score formula \cite{vincent2011connection},
\begin{align}
    \nabla \log \vpalt{\fwmargd{t}{\lstate_t}} &= \PE{}{\nabla \log \vpalt{\fwtransd{0}{t}{\state_0}{\state_t}} | \vpalt{\state_t = \lstate_t}} \\
    &= \PE{}{-\vpalt{\frac{\state_t - \sqrt{\alpha_t}\state_0}{1 - \alpha_t}} | \vpalt{\state_t = \lstate_t}} =
    \vpalt{\sqrt{\alpha_t}^{-1}}\PE{}{-\vpalt{\frac{\sqrt{\alpha_t}^{-1}\state_t - \state_0}{\alpha_t^{-1}(1 - \alpha_t)}} | \vpalt{\state_t = \lstate_t}} \eqsp.
\end{align}
But by the equality of the conditional distributions, this can be written as
\begin{equation}
    \nabla \log \vpalt{\fwmargd{t}{\lstate_t}} = \vpalt{\sqrt{\alpha_t}^{-1}}\PE{}{-\frac{\state_s - \state_0}{\vestd_s^2} | \state_s = \vpalt{\sqrt{\alpha_t}^{-1} \lstate_t}} = \vpalt{\sqrt{\alpha_t}} \nabla \log \fwmargd{s}{\vpalt{\sqrt{\alpha_t}^{-1} \lstate_t}}\eqsp.
\end{equation}

\section{Additional Experiments}
\label{app:additional_exps}
\subsection{Choice of $\rho$}
\label{app:additional_exps:rho}
\begin{figure}
    \centering
    \begin{tabular}{
            M{0.4\linewidth}@{\hspace{0.1\tabcolsep}}
            M{0.4\linewidth}@{\hspace{0.1\tabcolsep}}
            }
    \includegraphics[width=\hsize]{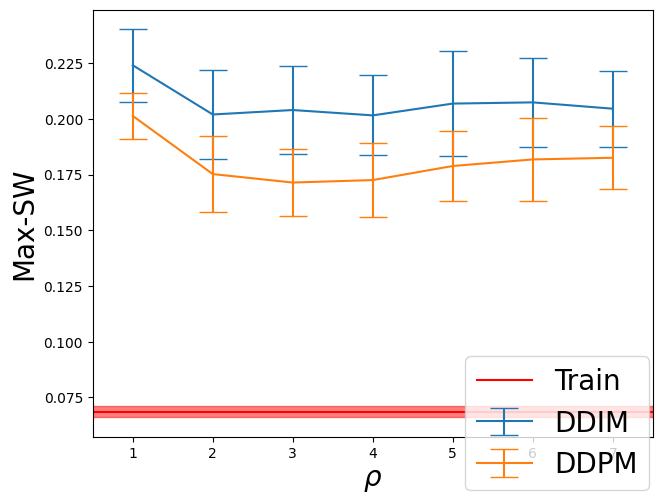} & 
    \includegraphics[width=\hsize]{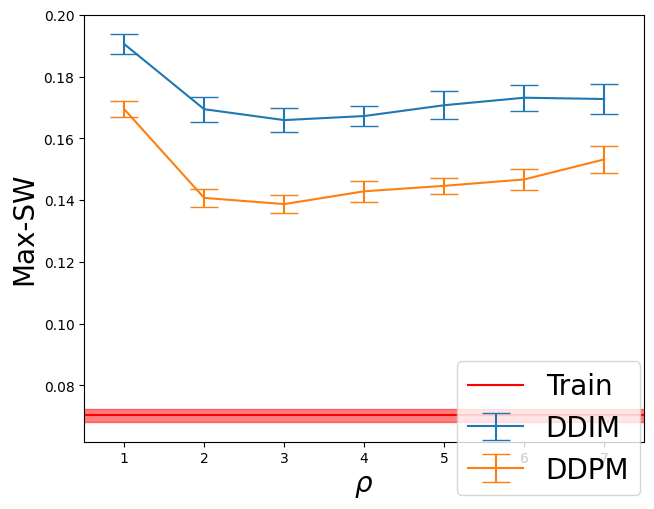}
    \end{tabular}
    \caption{Figure showing the evolution of the {\maxsw} with respect to $\rho$ for both {\ddim} and {\ddpm}.
    Left graph error bar are $2\sigma$ error bars while right error bars are $95\%$ assymptotic intervals (CLT based).}
    \label{fig:max_sw_vs_rho}
\end{figure}

In this section, we investigate the generation performance of samplers with varying the schedule parameter, namely $\rho$.
To do so, we focus on two samplers, {\ddpm} and {\ddim} and vary $\rho$ to generate for each configuration $50 000$ samples.
We did it for the model without fine-tuning (data generation described in \Cref{sec:num:data}). Then we calculated the {\maxsw} with $2^{16}$ slices and $50 000$ samples, with $20$ replicates (randomized over slices and subsamples).
The results are shown in \Cref{fig:max_sw_vs_rho}, where the error bars correspond to $2$ times the standard deviation.
We also display all the values of {\maxsw} and {\c2st} produced during the experiments in \Cref{table:all_maxsw}.
\begin{filecontents*}{data/final_aniso_metrics_generation.csv}
,sampler_name,n_steps,resnet101-acc-std,resnet18-acc-std,resnet50-acc-std,resnet101-acc-mean,resnet18-acc-mean,resnet50-acc-mean,sampler,rho,Unnamed: 0,mean,std,len,clt_gap,value,nfe
18,,0,,,,,,,Train,,0,0.07,0.005,20,0.002,0.070 + 0.002,0
0,,100,,,,,,,DDIM,1.0,1,0.191,0.007,20,0.003,0.191 + 0.003,100
7,,100,,,,,,,DDPM,1.0,9,0.169,0.006,20,0.003,0.169 + 0.003,100
1,,100,,,,,,,DDIM,2.0,2,0.169,0.009,20,0.004,0.169 + 0.004,100
8,,100,,,,,,,DDPM,2.0,10,0.141,0.007,20,0.003,0.141 + 0.003,100
2,ddim,100,,0.008669958129392736,,,0.651339852809906,,DDIM,3.0,3,0.166,0.009,20,0.004,0.166 + 0.004,100
9,,100,,,,,,,DDPM,3.0,11,0.139,0.007,20,0.003,0.139 + 0.003,100
3,,100,,,,,,,DDIM,4.0,4,0.167,0.008,20,0.003,0.167 + 0.003,100
10,,100,,,,,,,DDPM,4.0,12,0.143,0.008,20,0.003,0.143 + 0.003,100
4,,100,,,,,,,DDIM,5.0,5,0.171,0.01,20,0.005,0.171 + 0.005,100
11,,100,,,,,,,DDPM,5.0,13,0.145,0.006,20,0.003,0.145 + 0.003,100
5,,100,,,,,,,DDIM,6.0,6,0.173,0.01,20,0.004,0.173 + 0.004,100
12,,100,,,,,,,DDPM,6.0,14,0.147,0.008,20,0.003,0.147 + 0.003,100
6,,100,,,,,,,DDIM,7.0,7,0.173,0.011,20,0.005,0.173 + 0.005,100
13,,100,,,,,,,DDPM,7.0,15,0.153,0.01,20,0.004,0.153 + 0.004,100
14,ddpm,250,,0.03296370327483524,,,0.6197499990463257,,DDPM,3.0,16,0.111,0.007,20,0.003,0.111 + 0.003,250
17,edm,128,,0.0025530527022312754,,,0.5683392286300659,,Heun,3.0,8,0.146,0.01,20,0.005,0.146 + 0.005,256
15,,1000,,,,,,,DDPM,1.0,17,0.109,0.009,20,0.004,0.109 + 0.004,1000
16,ddpm,1000,0.0028002676951471663,0.0014479281025406767,0.0015836097324546314,0.5197693586349488,0.5296322584152222,0.5247514247894287,DDPM,3.0,18,0.096,0.009,20,0.004,0.096 + 0.004,1000
\end{filecontents*}
\DTLloaddb{all_metrics_aniso}{data/final_aniso_metrics_generation.csv}
\begin{table}[h]
    \begin{center}
        \begin{tabular}{|c|c|c|c|c|c|c|}%
            \hline
            Sampler & N steps & $\rho$ & {\maxsw} & {\c2st} Resnet18 & {\c2st} Resnet50 & {\c2st} Resnet101\\
            \hline
            \DTLforeach*{all_metrics_aniso}{
                \samplerallg=sampler,
                \nstepsallg=n_steps,
                \rhovalg=rho,
                \mswmeanallg=mean,
            	\mswstdallg=std,
                \rsmallallg=resnet18-acc-mean,
            	\rsmallallgstd=resnet18-acc-std,
                \rmediumallg=resnet50-acc-mean,
            	\rmediumallgstd=resnet50-acc-std,
                \rbigallg=resnet101-acc-mean,
            	\rbigallgstd=resnet101-acc-std}
                {
                \samplerallg & \nstepsallg & \DTLround{\rhovalg}{\rhovalg}{1}\rhovalg & \DTLround{\mswmeanallg}{\mswmeanallg}{3}$\mswmeanallg$ (\DTLround{\mswstdallg}{\mswstdallg}{3}$\mswstdallg$)%
                & \DTLifnullorempty{\rsmallallg}{}{\DTLround{\rsmallallg}{\rsmallallg}{3}$\rsmallallg$} \DTLifnullorempty{\rsmallallgstd}{}{(\DTLround{\rsmallallgstd}{\rsmallallgstd}{3}$\rsmallallgstd$)}%
                & \DTLifnullorempty{\rmediumallg}{}{\DTLround{\rmediumallg}{\rmediumallg}{3}$\rmediumallg$} \DTLifnullorempty{\rmediumallgstd}{}{(\DTLround{\rmediumallgstd}{\rmediumallgstd}{3}$\rmediumallgstd$)}%
                & \DTLifnullorempty{\rbigallg}{}{\DTLround{\rbigallg}{\rbigallg}{3}$\rbigallg$} \DTLifnullorempty{\rbigallgstd}{}{(\DTLround{\rbigallgstd}{\rbigallgstd}{3}$\rbigallgstd$)}\DTLiflastrow{}{\\}}
            \\\hline
        \end{tabular}% 
    \end{center}
    \caption{{\maxsw} and {\c2st} between held-out dataset and different DGM samplers in the form "mean (standard deviation)". For {\maxsw}, a total of $2^{16}$ slices were used. Quantities where aggregated over 20 different slices and different $10^4$ subsets draws from the pool of available generated samples ($5\times 10^4$).}
    \label{table:all_maxsw}
\end{table}
\subsection{Sea surface temperature anomaly dataset (SSTA):} 
\label{app:additional_exps:SSTA}
\begin{figure}
    \begin{center}
        \begin{tabular}{
            M{0.18\linewidth}@{\hspace{0.1\tabcolsep}}
            M{0.18\linewidth}@{\hspace{0.1\tabcolsep}}
            M{0.18\linewidth}@{\hspace{0.1\tabcolsep}}
            M{0.18\linewidth}@{\hspace{0.1\tabcolsep}}
            }
            Observation & {\mgdm} &  \vecchia &  \priorvae  \\
            \includegraphics[width=\hsize]{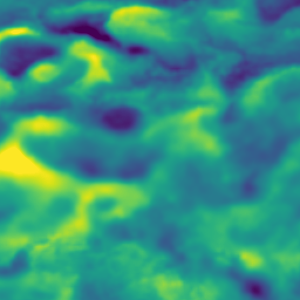} 
            & \includegraphics[width=\hsize]{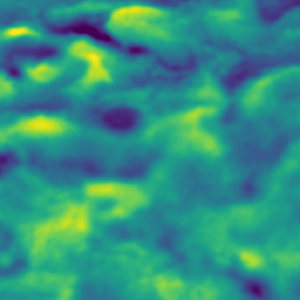} 
            & \includegraphics[width=\hsize]{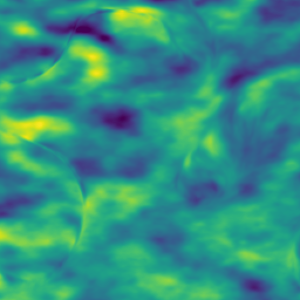} 
            & \includegraphics[width=\hsize]{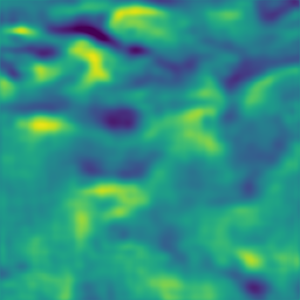}\\\addlinespace[-2.6pt]
            \includegraphics[width=\hsize]{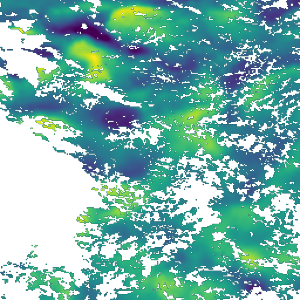} 
            & \includegraphics[width=\hsize]{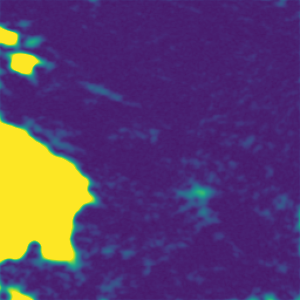} 
            & \includegraphics[width=\hsize]{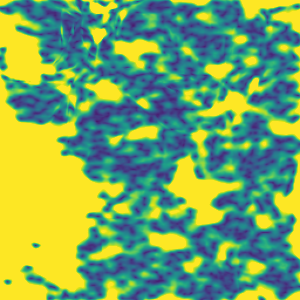} 
            & \includegraphics[width=\hsize]{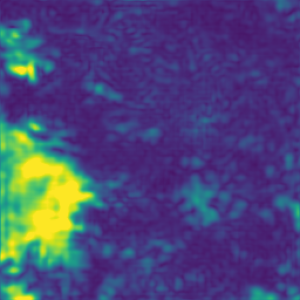}
        \end{tabular}
    \end{center}
    \caption{Illustration of from different posterior sampling methods on the sea surface temperature problem, for the case 2.
    First column show the full measurements (top) and observation (bottom, measurements with cloud). The other columns correspond to a sample (top) and the standard deviation obtained over $100$ posterior samples for each method.
    The colors are normalized between $-3$ (blue) and $3$ (yellow) except for the standard deviation, which is between $0$ (blue) and $0.3$ (yellow).}
    \label{fig:inv_problem_cloud:2}
\end{figure}

\begin{figure}
    \begin{center}
        \begin{tabular}{
            M{0.18\linewidth}@{\hspace{0.1\tabcolsep}}
            M{0.18\linewidth}@{\hspace{0.1\tabcolsep}}
            M{0.18\linewidth}@{\hspace{0.1\tabcolsep}}
            M{0.18\linewidth}@{\hspace{0.1\tabcolsep}}
            }
            Observation & {\mgdm} &  \vecchia &  \priorvae  \\
            \includegraphics[width=\hsize]{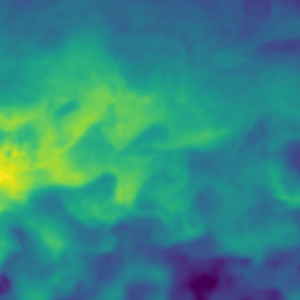} 
            & \includegraphics[width=\hsize]{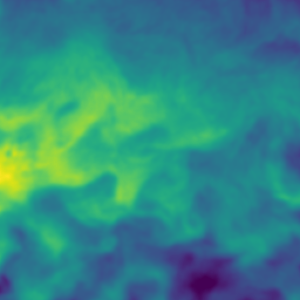} 
            & \includegraphics[width=\hsize]{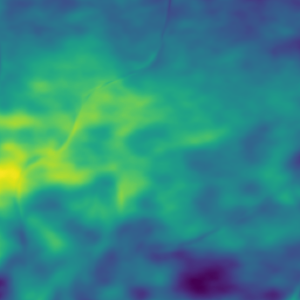}
            & \includegraphics[width=\hsize]{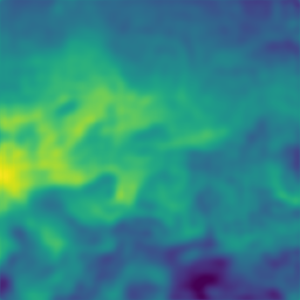}\\\addlinespace[-2.6pt]
            \includegraphics[width=\hsize]{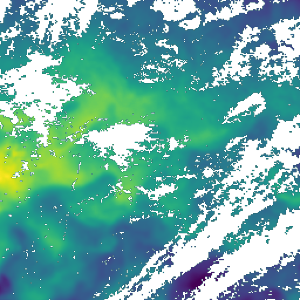} 
            & \includegraphics[width=\hsize]{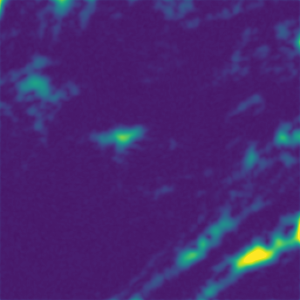} 
            & \includegraphics[width=\hsize]{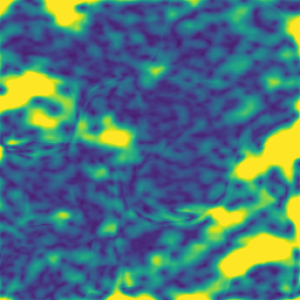} 
            & \includegraphics[width=\hsize]{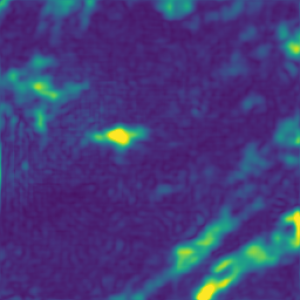}
        \end{tabular}
    \end{center}
    \caption{Illustration of from different posterior sampling methods on the sea surface temperature problem for the case 0.
    First column show the full measurements (top) and observation (bottom, measurements with cloud). The other columns correspond to a sample (top) and the standard deviation obtained over $100$ posterior samples for each method.
    The colors are normalized between $-3$ (blue) and $3$ (yellow) except for the standard deviation, which is between $0$ (blue) and $0.3$ (yellow).}
    \label{fig:inv_problem_cloud:0}
\end{figure}

In this section, we provide further visualization of the experiments in the SSTA dataset.
\Cref{fig:inv_problem_cloud:0,fig:inv_problem_cloud:2} are equivalent to \Cref{fig:inv_problem_cloud:1} but for cases 0 and 2 of \Cref{table:rec_cloud} respectively.

\begin{figure}
    \centering
    \begin{tabular}{ccccc}
        \includegraphics[width=0.19\textwidth]{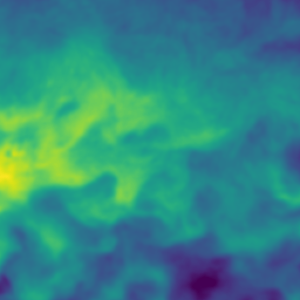}&
        \hspace{-10pt}\includegraphics[width=0.19\textwidth]{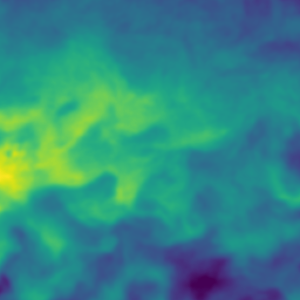}&
        \hspace{-10pt}\includegraphics[width=0.19\textwidth]{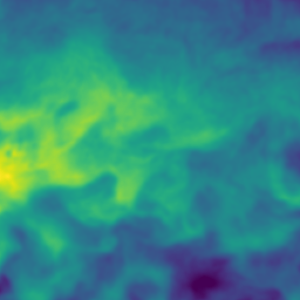}&
        \hspace{-10pt}\includegraphics[width=0.19\textwidth]{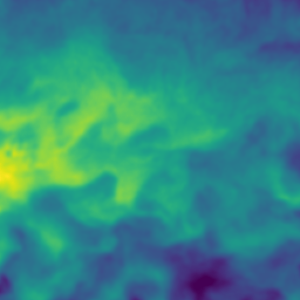}&
        \hspace{-10pt}\includegraphics[width=0.19\textwidth]{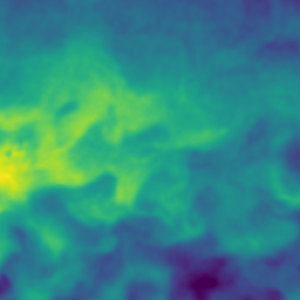}\\
        \includegraphics[width=0.19\textwidth]{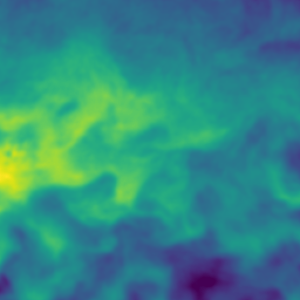}&
        \hspace{-10pt}\includegraphics[width=0.19\textwidth]{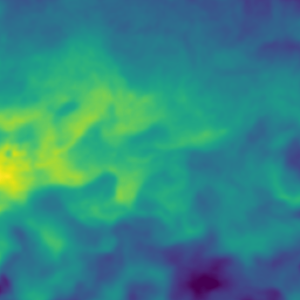}&
        \hspace{-10pt}\includegraphics[width=0.19\textwidth]{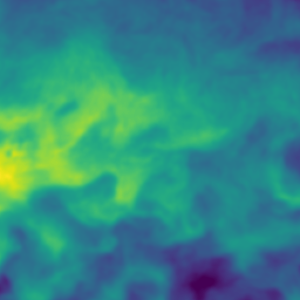}&
        \hspace{-10pt}\includegraphics[width=0.19\textwidth]{images/posterior_samples/cloud_inpainting_0/mgdm_two_times_gibbs/sample_8.png}&
        \hspace{-10pt}\includegraphics[width=0.19\textwidth]{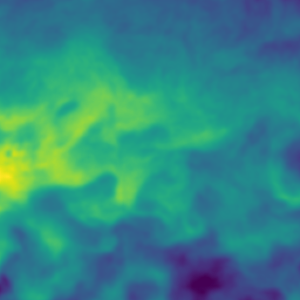}\\
        \includegraphics[width=0.19\textwidth]{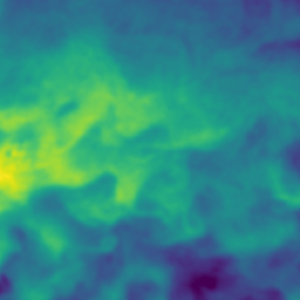}&
        \hspace{-10pt}\includegraphics[width=0.19\textwidth]{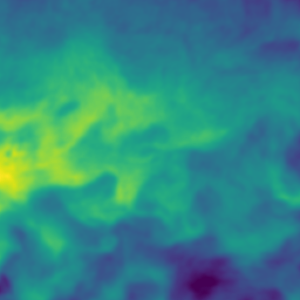}&
        \hspace{-10pt}\includegraphics[width=0.19\textwidth]{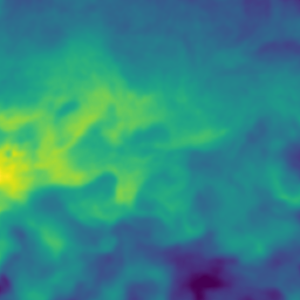}&
        \hspace{-10pt}\includegraphics[width=0.19\textwidth]{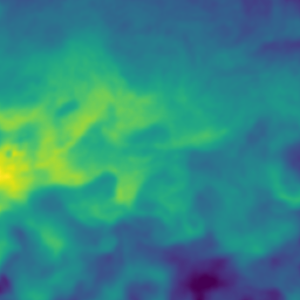}&
        \hspace{-10pt}\includegraphics[width=0.19\textwidth]{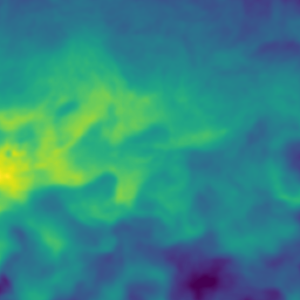}\\
        \includegraphics[width=0.19\textwidth]{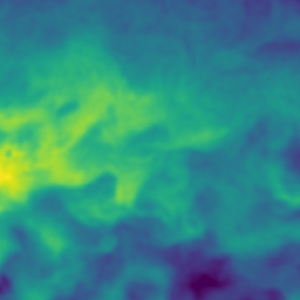}&
        \hspace{-10pt}\includegraphics[width=0.19\textwidth]{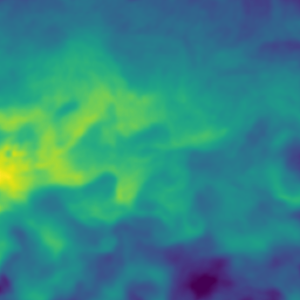}&
        \hspace{-10pt}\includegraphics[width=0.19\textwidth]{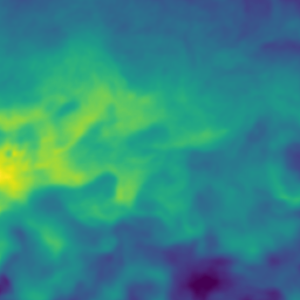}&
        \hspace{-10pt}\includegraphics[width=0.19\textwidth]{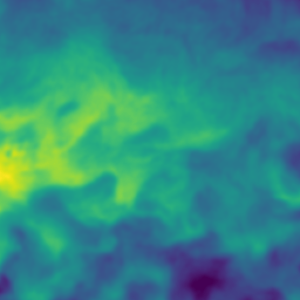}&
        \hspace{-10pt}\includegraphics[width=0.19\textwidth]{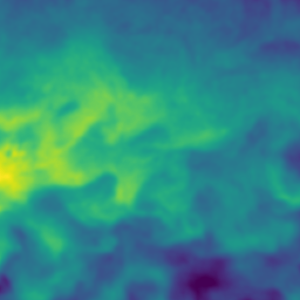}\\
        \includegraphics[width=0.19\textwidth]{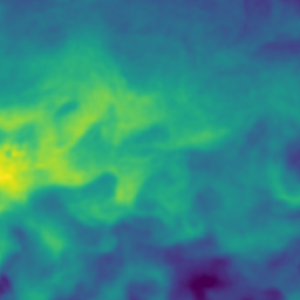}&
        \hspace{-10pt}\includegraphics[width=0.19\textwidth]{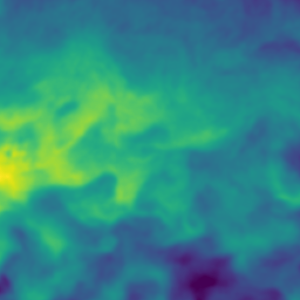}&
        \hspace{-10pt}\includegraphics[width=0.19\textwidth]{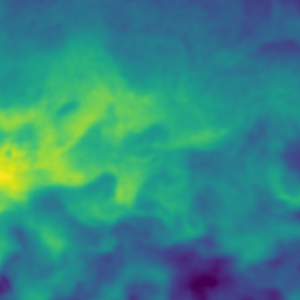}&
        \hspace{-10pt}\includegraphics[width=0.19\textwidth]{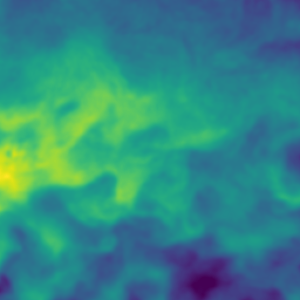}&
        \hspace{-10pt}\includegraphics[width=0.19\textwidth]{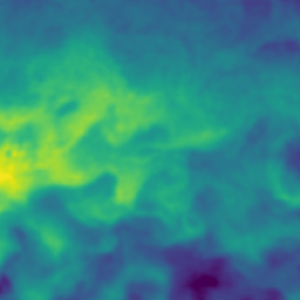}\\
        \includegraphics[width=0.19\textwidth]{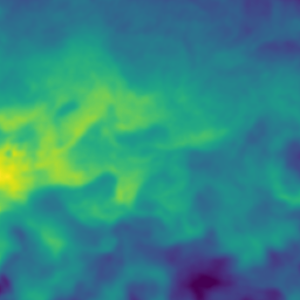}&
        \hspace{-10pt}\includegraphics[width=0.19\textwidth]{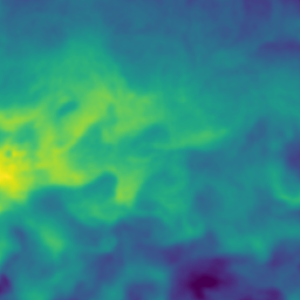}&
        \hspace{-10pt}\includegraphics[width=0.19\textwidth]{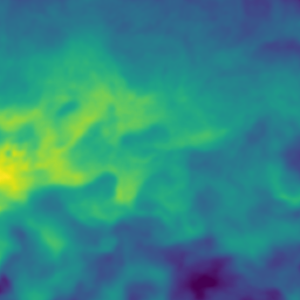}&
        \hspace{-10pt}\includegraphics[width=0.19\textwidth]{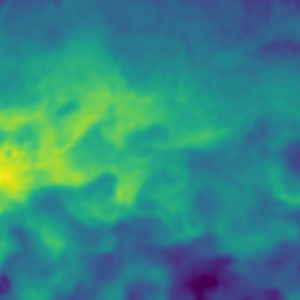}&
        \hspace{-10pt}\includegraphics[width=0.19\textwidth]{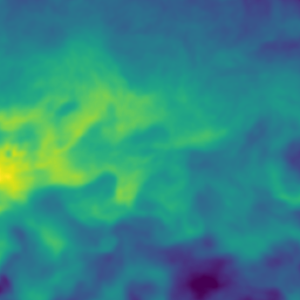}\\
    \end{tabular}
    \caption{A subset of samples of the {\mgdm} posterior for the Sea surface temperature experiment for case 0 from \Cref{table:rec_cloud}.}
    \label{fig:clouds_all_samples:0}
\end{figure}

\begin{figure}
    \centering
    \begin{tabular}{ccccc}
        \includegraphics[width=0.19\textwidth]{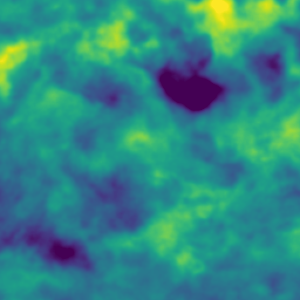}&
        \hspace{-10pt}\includegraphics[width=0.19\textwidth]{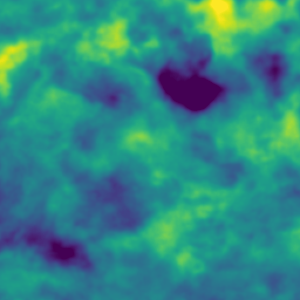}&
        \hspace{-10pt}\includegraphics[width=0.19\textwidth]{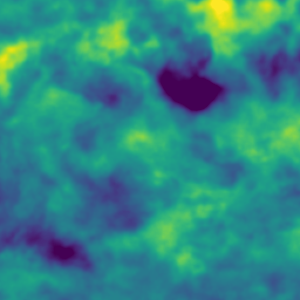}&
        \hspace{-10pt}\includegraphics[width=0.19\textwidth]{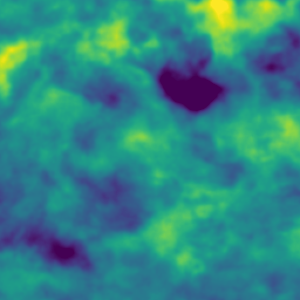}&
        \hspace{-10pt}\includegraphics[width=0.19\textwidth]{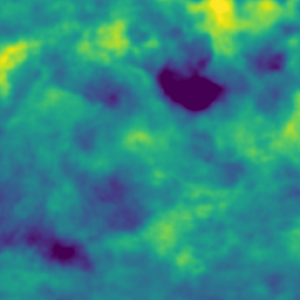}\\
        \includegraphics[width=0.19\textwidth]{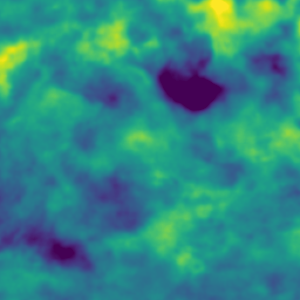}&
        \hspace{-10pt}\includegraphics[width=0.19\textwidth]{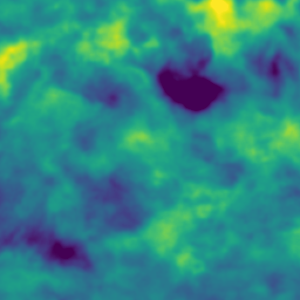}&
        \hspace{-10pt}\includegraphics[width=0.19\textwidth]{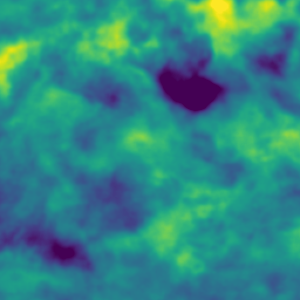}&
        \hspace{-10pt}\includegraphics[width=0.19\textwidth]{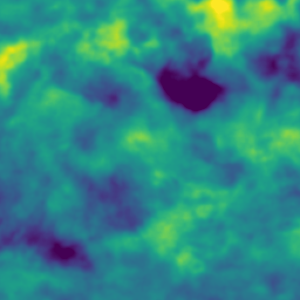}&
        \hspace{-10pt}\includegraphics[width=0.19\textwidth]{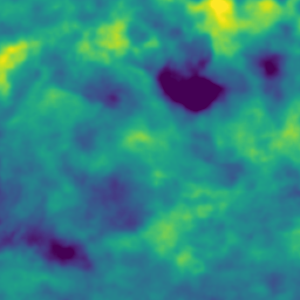}\\
        \includegraphics[width=0.19\textwidth]{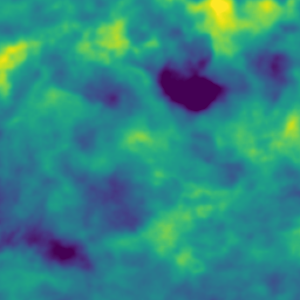}&
        \hspace{-10pt}\includegraphics[width=0.19\textwidth]{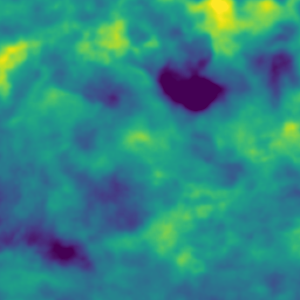}&
        \hspace{-10pt}\includegraphics[width=0.19\textwidth]{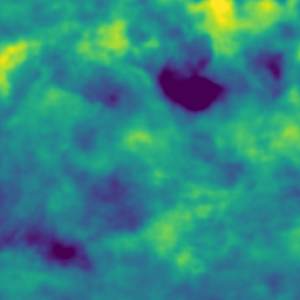}&
        \hspace{-10pt}\includegraphics[width=0.19\textwidth]{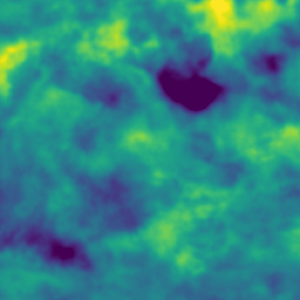}&
        \hspace{-10pt}\includegraphics[width=0.19\textwidth]{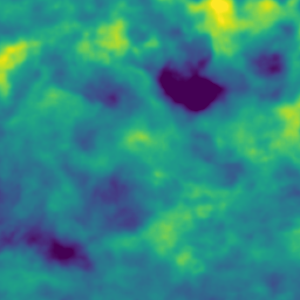}\\
        \includegraphics[width=0.19\textwidth]{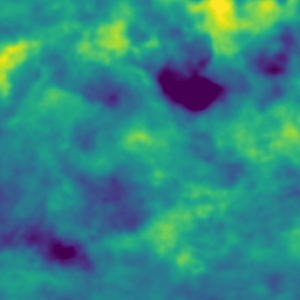}&
        \hspace{-10pt}\includegraphics[width=0.19\textwidth]{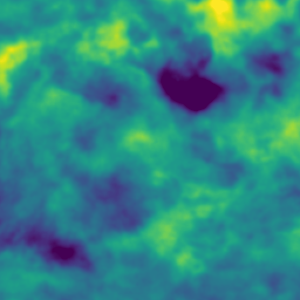}&
        \hspace{-10pt}\includegraphics[width=0.19\textwidth]{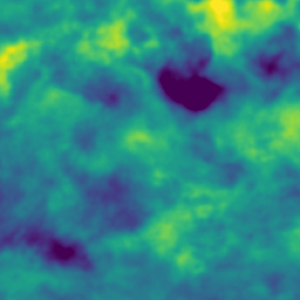}&
        \hspace{-10pt}\includegraphics[width=0.19\textwidth]{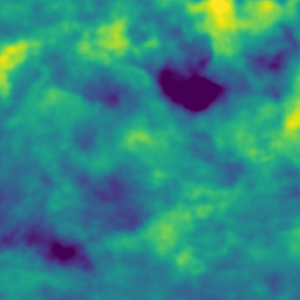}&
        \hspace{-10pt}\includegraphics[width=0.19\textwidth]{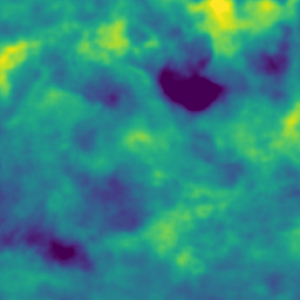}\\
        \includegraphics[width=0.19\textwidth]{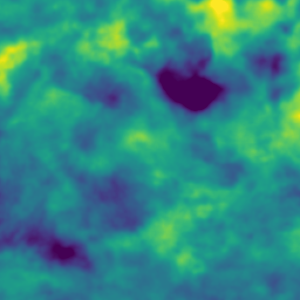}&
        \hspace{-10pt}\includegraphics[width=0.19\textwidth]{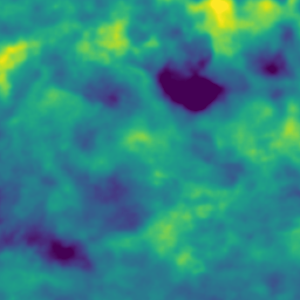}&
        \hspace{-10pt}\includegraphics[width=0.19\textwidth]{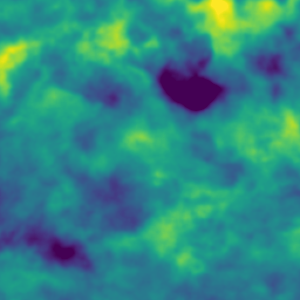}&
        \hspace{-10pt}\includegraphics[width=0.19\textwidth]{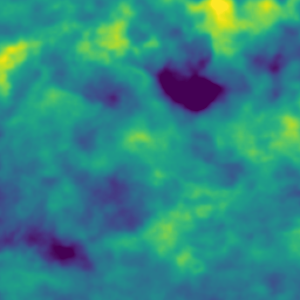}&
        \hspace{-10pt}\includegraphics[width=0.19\textwidth]{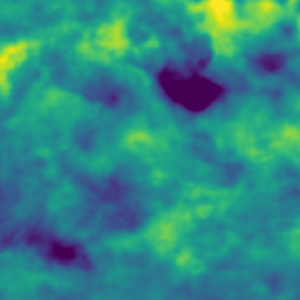}\\
        \includegraphics[width=0.19\textwidth]{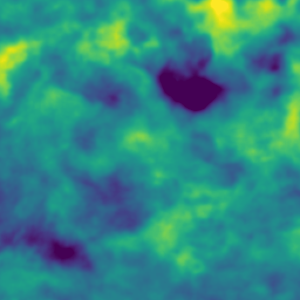}&
        \hspace{-10pt}\includegraphics[width=0.19\textwidth]{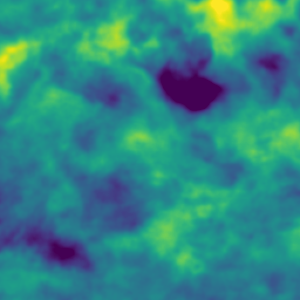}&
        \hspace{-10pt}\includegraphics[width=0.19\textwidth]{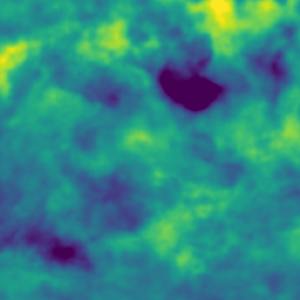}&
        \hspace{-10pt}\includegraphics[width=0.19\textwidth]{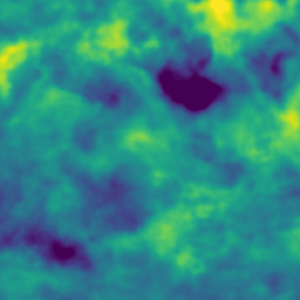}&
        \hspace{-10pt}\includegraphics[width=0.19\textwidth]{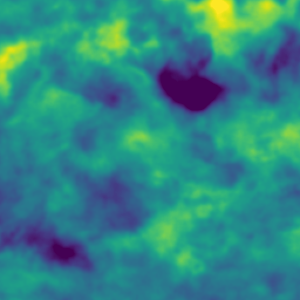}\\
    \end{tabular}
    \caption{A subset of samples of the {\mgdm} posterior for the Sea surface temperature experiment for case 1 from \Cref{table:rec_cloud}.}
    \label{fig:clouds_all_samples:1}
\end{figure}

\begin{figure}
    \centering
    \begin{tabular}{ccccc}
        \includegraphics[width=0.19\textwidth]{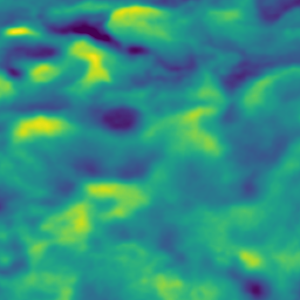}&
        \hspace{-10pt}\includegraphics[width=0.19\textwidth]{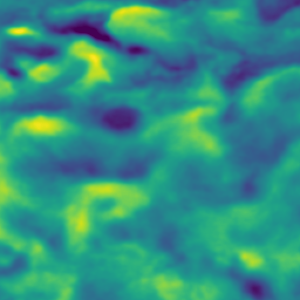}&
        \hspace{-10pt}\includegraphics[width=0.19\textwidth]{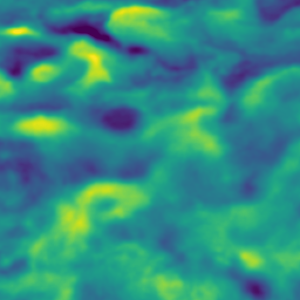}&
        \hspace{-10pt}\includegraphics[width=0.19\textwidth]{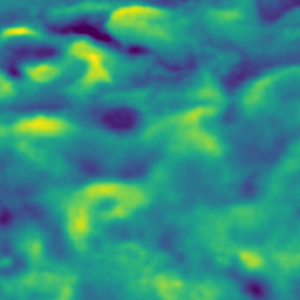}&
        \hspace{-10pt}\includegraphics[width=0.19\textwidth]{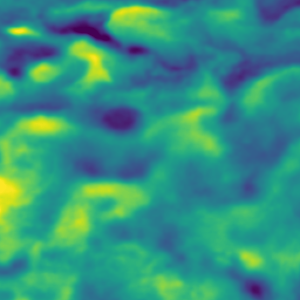}\\
        \includegraphics[width=0.19\textwidth]{images/posterior_samples/cloud_inpainting_2/mgdm_two_times_gibbs/sample_5.png}&
        \hspace{-10pt}\includegraphics[width=0.19\textwidth]{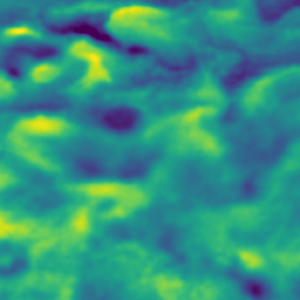}&
        \hspace{-10pt}\includegraphics[width=0.19\textwidth]{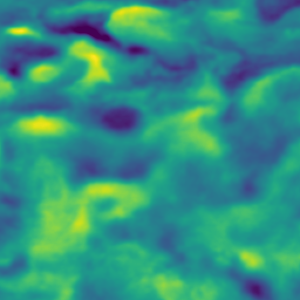}&
        \hspace{-10pt}\includegraphics[width=0.19\textwidth]{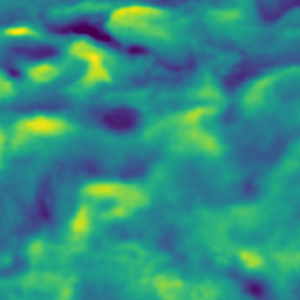}&
        \hspace{-10pt}\includegraphics[width=0.19\textwidth]{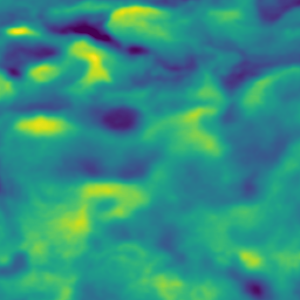}\\
        \includegraphics[width=0.19\textwidth]{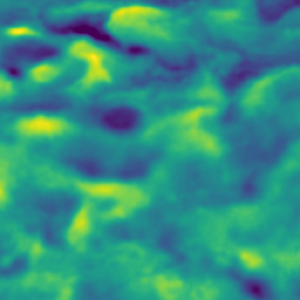}&
        \hspace{-10pt}\includegraphics[width=0.19\textwidth]{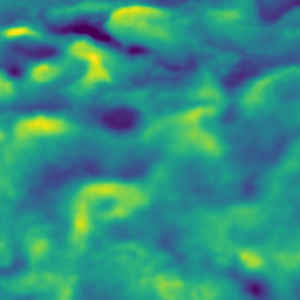}&
        \hspace{-10pt}\includegraphics[width=0.19\textwidth]{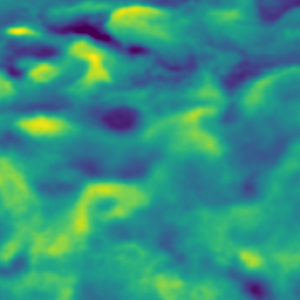}&
        \hspace{-10pt}\includegraphics[width=0.19\textwidth]{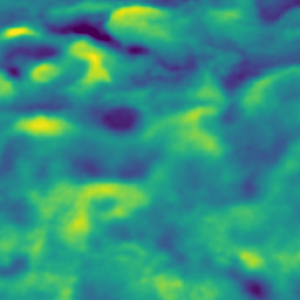}&
        \hspace{-10pt}\includegraphics[width=0.19\textwidth]{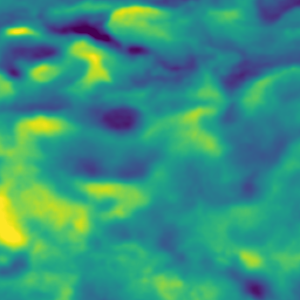}\\
        \includegraphics[width=0.19\textwidth]{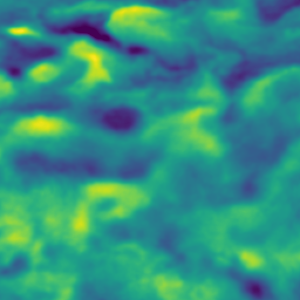}&
        \hspace{-10pt}\includegraphics[width=0.19\textwidth]{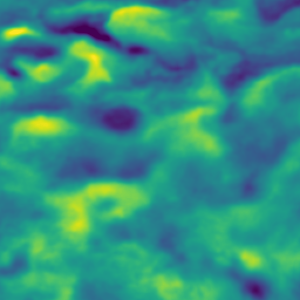}&
        \hspace{-10pt}\includegraphics[width=0.19\textwidth]{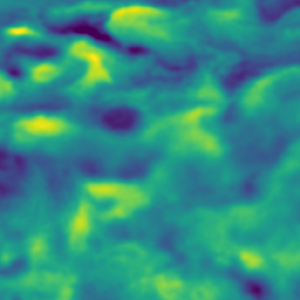}&
        \hspace{-10pt}\includegraphics[width=0.19\textwidth]{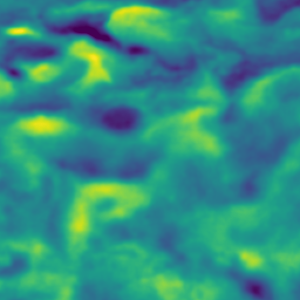}&
        \hspace{-10pt}\includegraphics[width=0.19\textwidth]{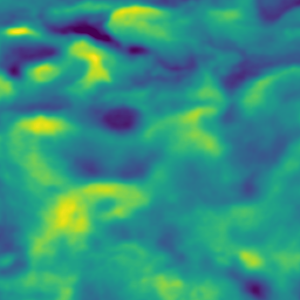}\\
        \includegraphics[width=0.19\textwidth]{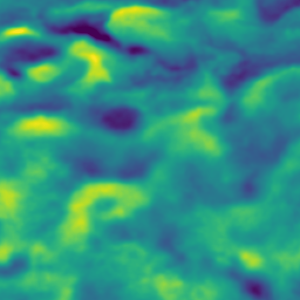}&
        \hspace{-10pt}\includegraphics[width=0.19\textwidth]{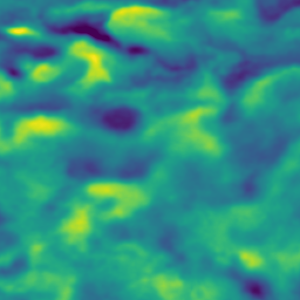}&
        \hspace{-10pt}\includegraphics[width=0.19\textwidth]{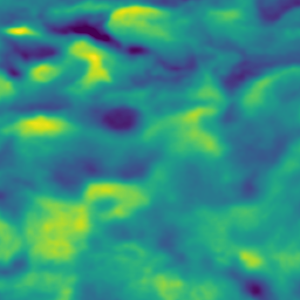}&
        \hspace{-10pt}\includegraphics[width=0.19\textwidth]{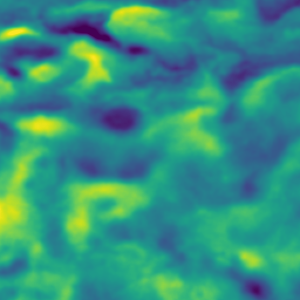}&
        \hspace{-10pt}\includegraphics[width=0.19\textwidth]{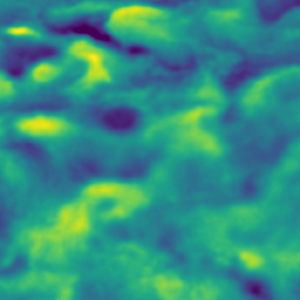}\\
        \includegraphics[width=0.19\textwidth]{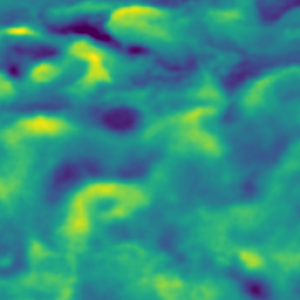}&
        \hspace{-10pt}\includegraphics[width=0.19\textwidth]{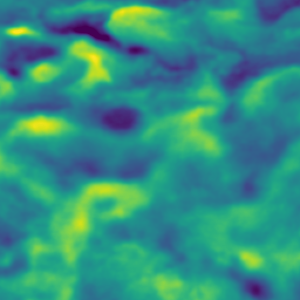}&
        \hspace{-10pt}\includegraphics[width=0.19\textwidth]{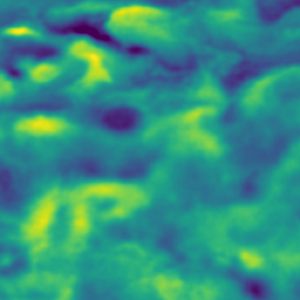}&
        \hspace{-10pt}\includegraphics[width=0.19\textwidth]{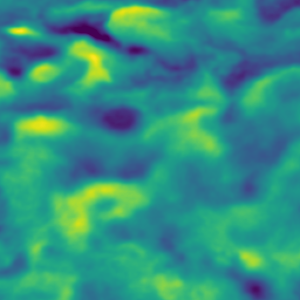}&
        \hspace{-10pt}\includegraphics[width=0.19\textwidth]{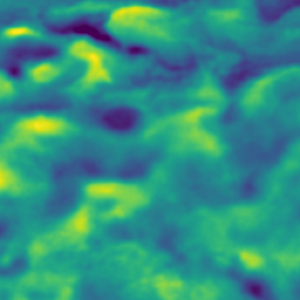}\\
    \end{tabular}
    \caption{A subset of samples of the {\mgdm} posterior for the Sea surface temperature experiment for case 2 from \Cref{table:rec_cloud}.}
    \label{fig:clouds_all_samples:2}
\end{figure}

We then present samples from {\mgdm} for the three cases in \Cref{fig:clouds_all_samples:0,fig:clouds_all_samples:1,fig:clouds_all_samples:2}.
\clearpage
\subsection{Confusion matrices and roc curves for {\c2st}}
\label{app:additional_exps:c2st}
\begin{figure}
    \begin{tabular}{
        M{0.1\linewidth}@{\hspace{0\tabcolsep}}
        M{0.4\linewidth}@{\hspace{0\tabcolsep}}
        M{0.4\linewidth}@{\hspace{0\tabcolsep}}
        }
         seed & Confusion matrix & ROC \\
         $42$  & \includegraphics[width=\hsize]{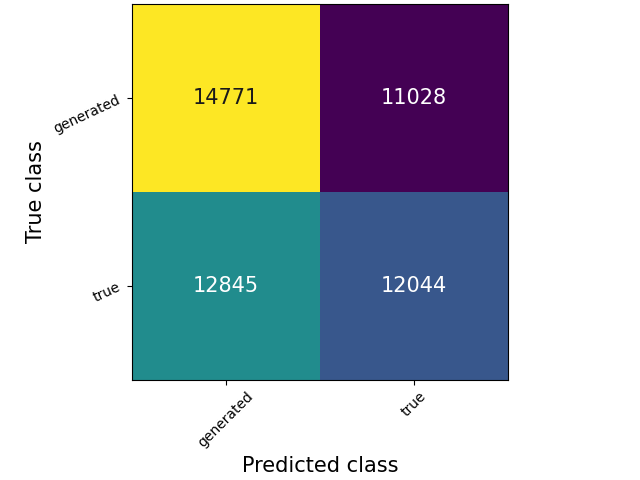} & \includegraphics[width=\hsize]{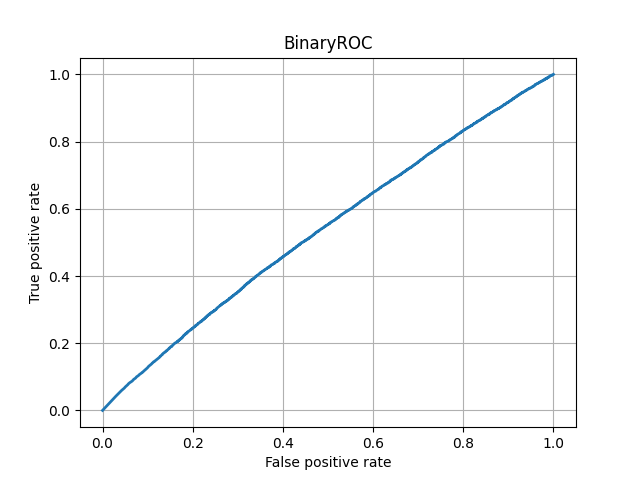}\\
         $43$  & \includegraphics[width=\hsize]{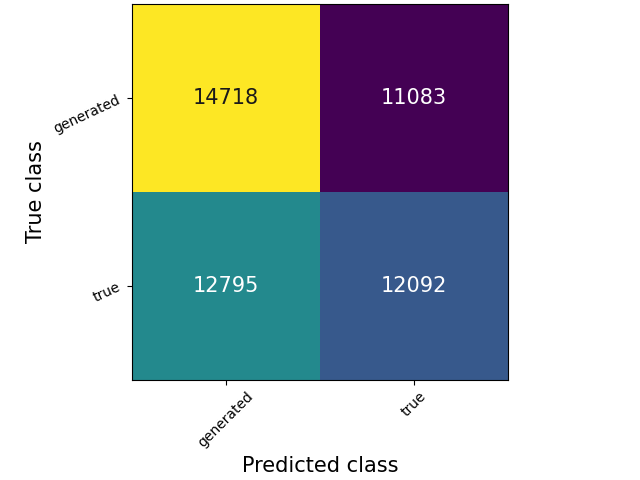} & \includegraphics[width=\hsize]{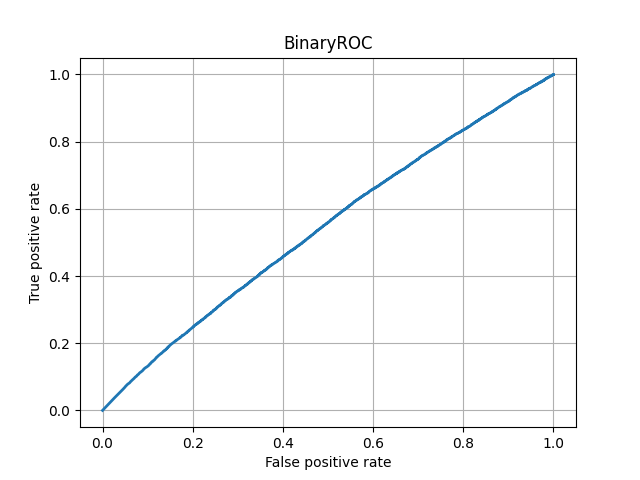}\\
         $44$  & \includegraphics[width=\hsize]{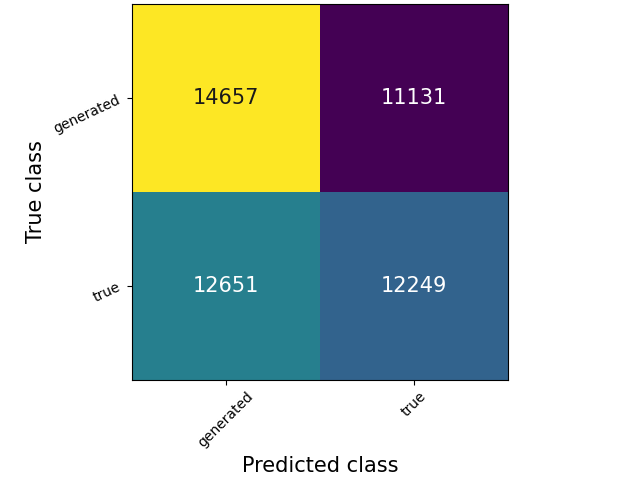} & \includegraphics[width=\hsize]{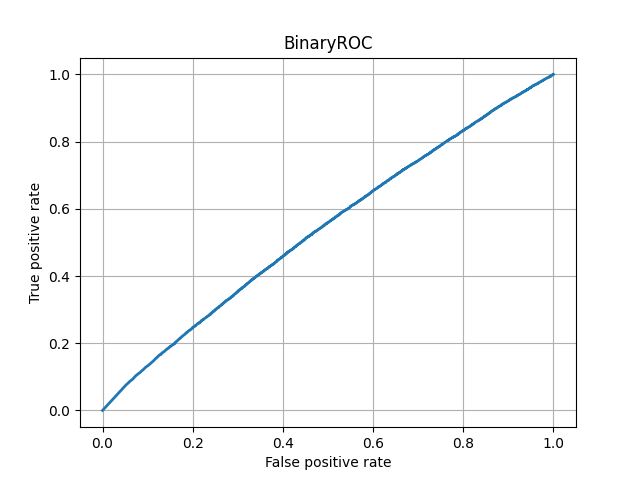}\\
         $45$  & \includegraphics[width=\hsize]{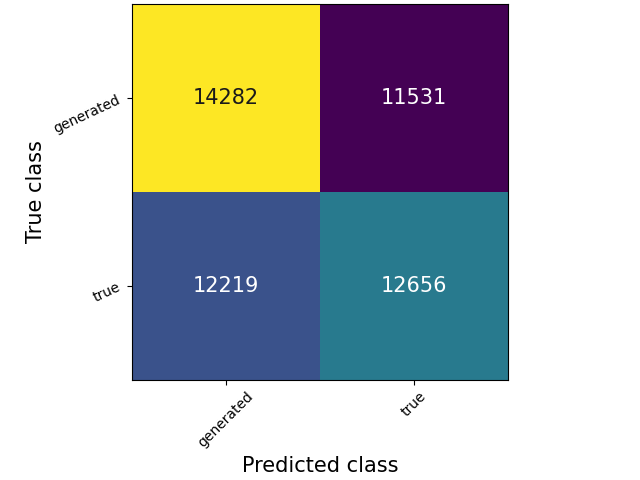} & \includegraphics[width=\hsize]{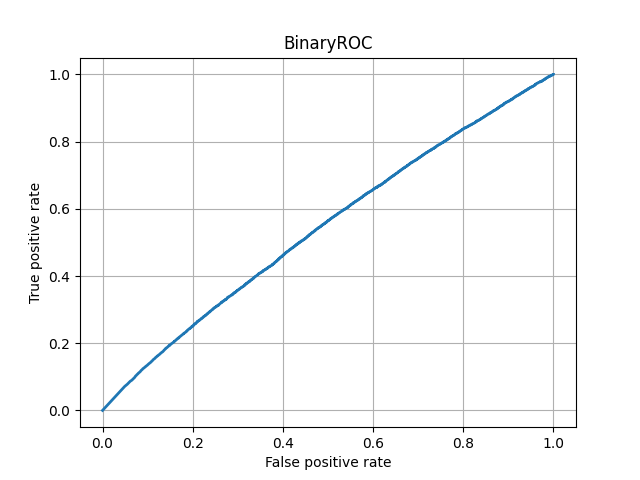}\\
         $46$  & \includegraphics[width=\hsize]{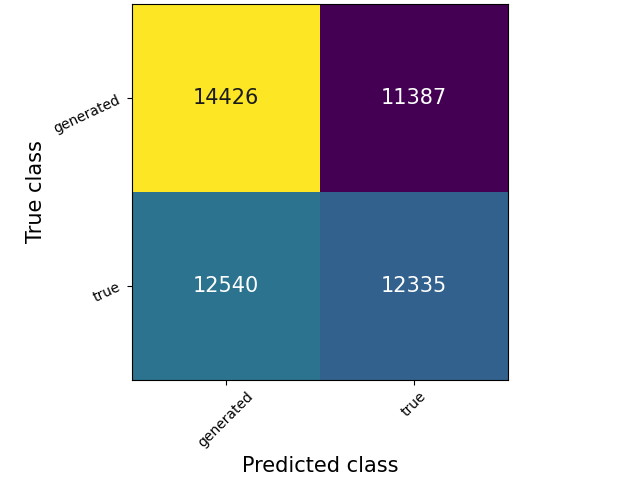} & \includegraphics[width=\hsize]{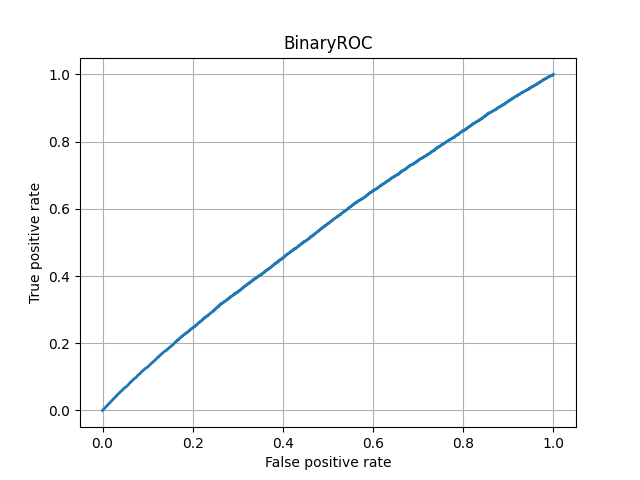}
    \end{tabular}
    \caption{Resnet 18}
    \label{fig:c2st_resnet18}
\end{figure}

\begin{figure}
    \begin{tabular}{
        M{0.1\linewidth}@{\hspace{0\tabcolsep}}
        M{0.4\linewidth}@{\hspace{0\tabcolsep}}
        M{0.4\linewidth}@{\hspace{0\tabcolsep}}
        }
         seed & Confusion matrix & ROC \\
         $42$  & \includegraphics[width=\hsize]{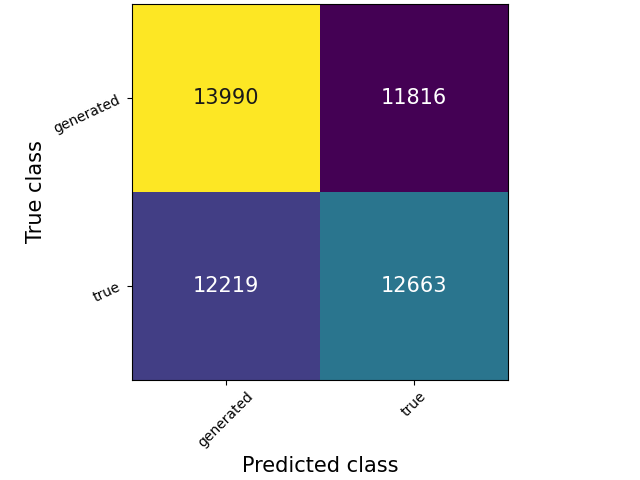} & \includegraphics[width=\hsize]{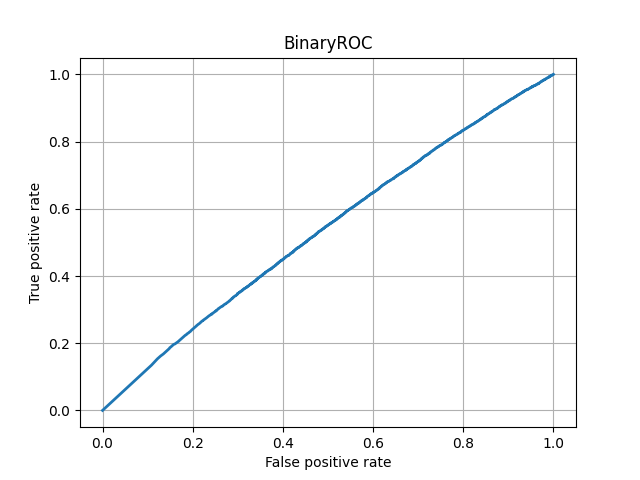}\\
         $43$  & \includegraphics[width=\hsize]{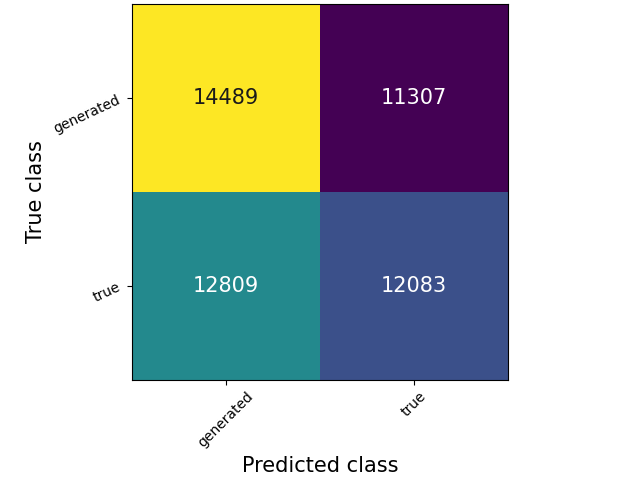} & \includegraphics[width=\hsize]{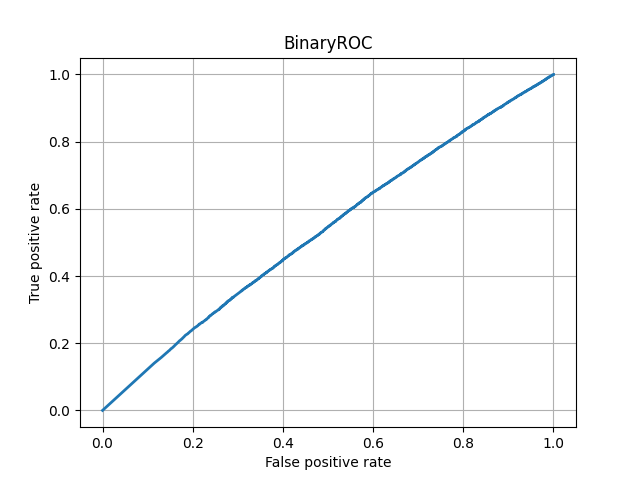}\\
         $44$  & \includegraphics[width=\hsize]{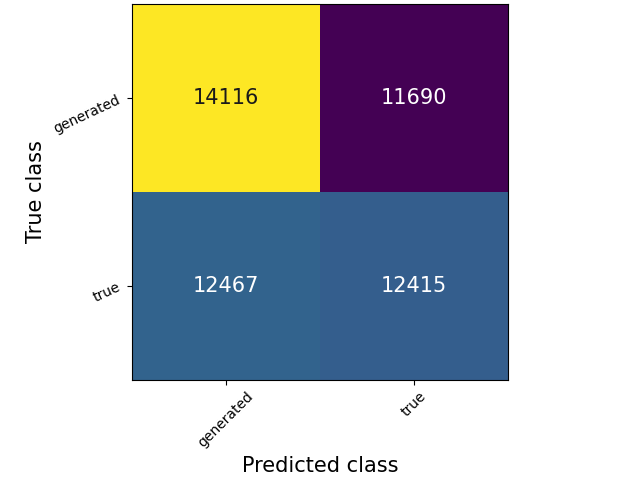} & \includegraphics[width=\hsize]{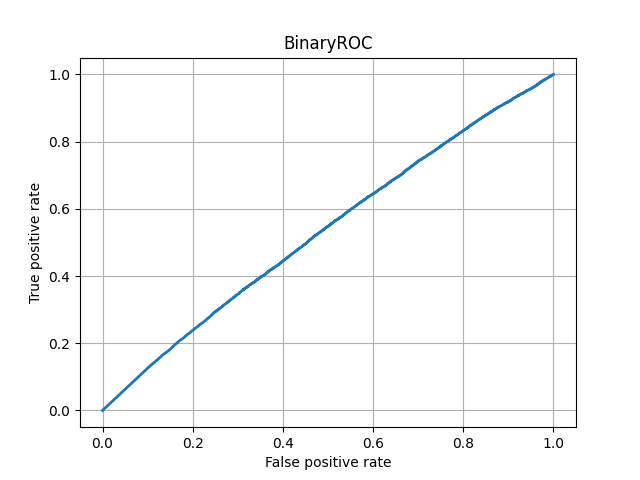}\\
         $45$  & \includegraphics[width=\hsize]{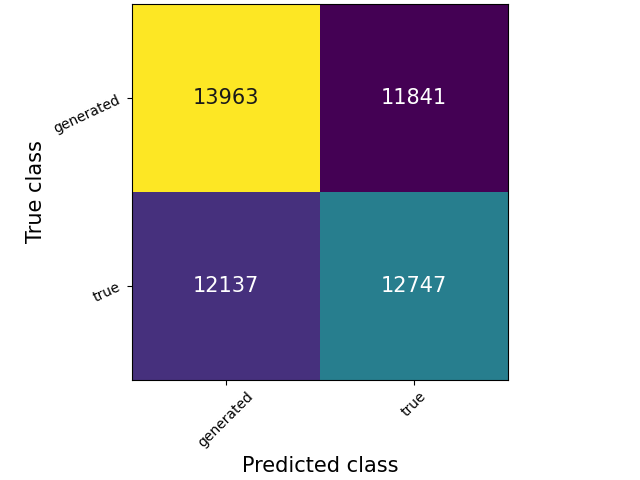} & \includegraphics[width=\hsize]{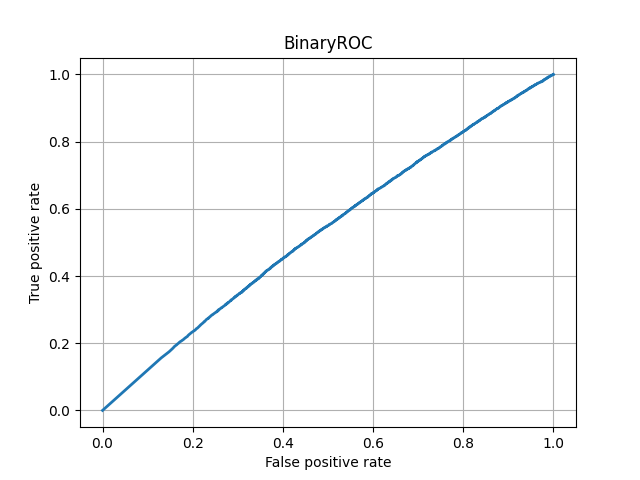}\\
         $46$  & \includegraphics[width=\hsize]{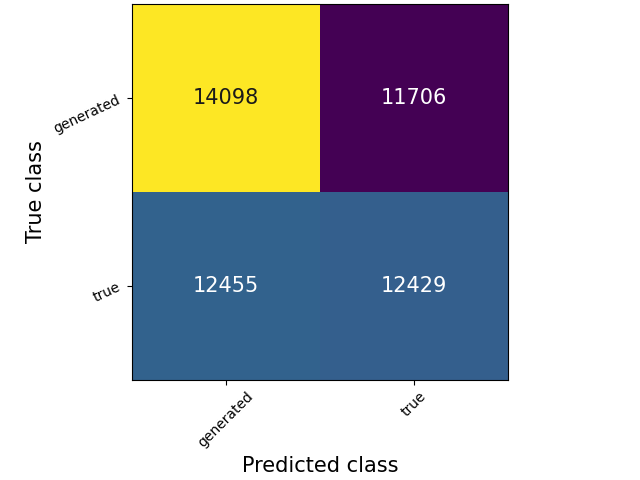} & \includegraphics[width=\hsize]{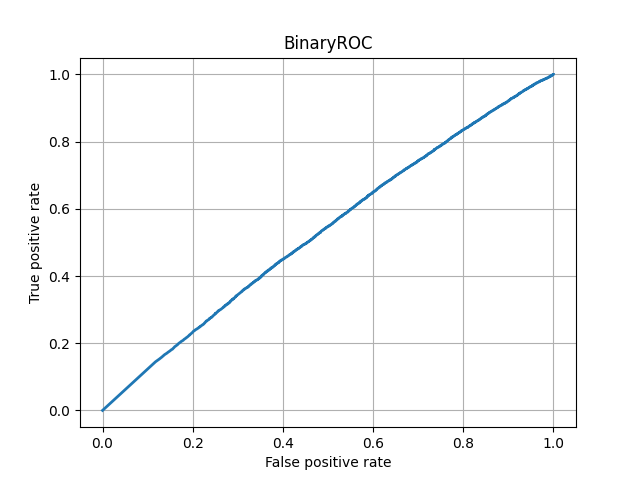}
    \end{tabular}
    \caption{Resnet 50}
    \label{fig:c2st_resnet50}
\end{figure}

\begin{figure}
    \begin{tabular}{
        M{0.1\linewidth}@{\hspace{0\tabcolsep}}
        M{0.4\linewidth}@{\hspace{0\tabcolsep}}
        M{0.4\linewidth}@{\hspace{0\tabcolsep}}
        }
         seed & Confusion matrix & ROC \\
         $42$  & \includegraphics[width=\hsize]{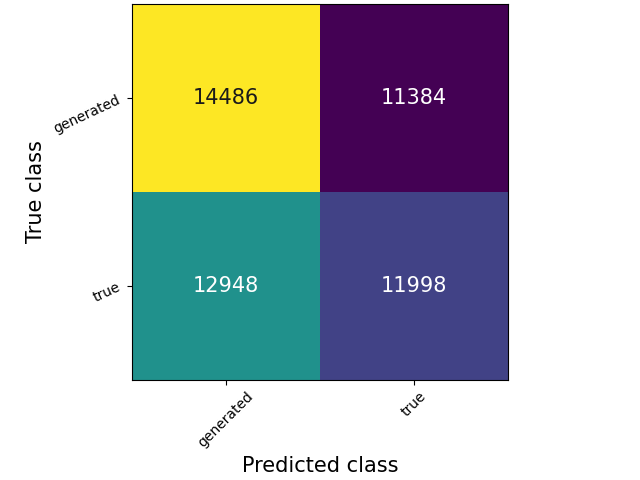} & \includegraphics[width=\hsize]{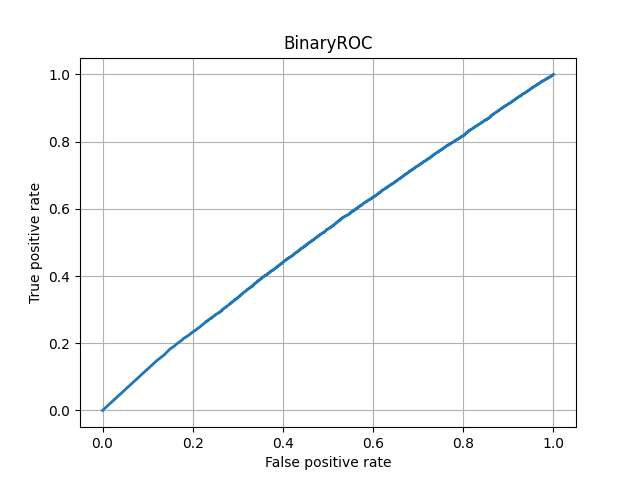}\\
         $43$  & \includegraphics[width=\hsize]{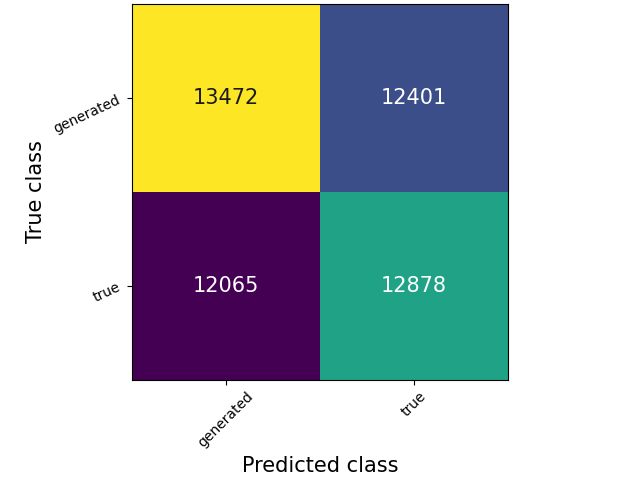} & \includegraphics[width=\hsize]{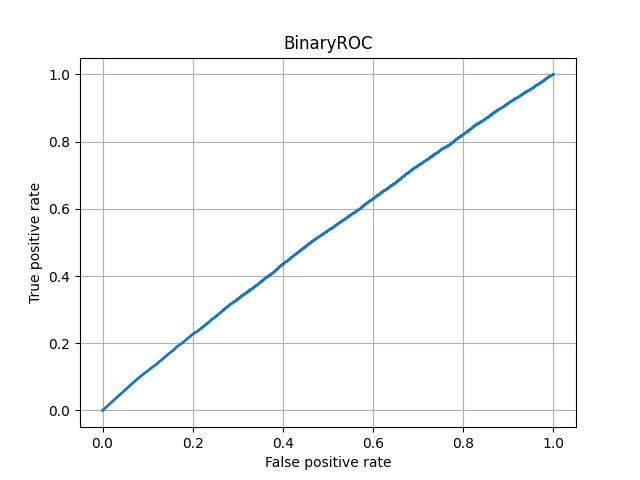}\\
         $44$  & \includegraphics[width=\hsize]{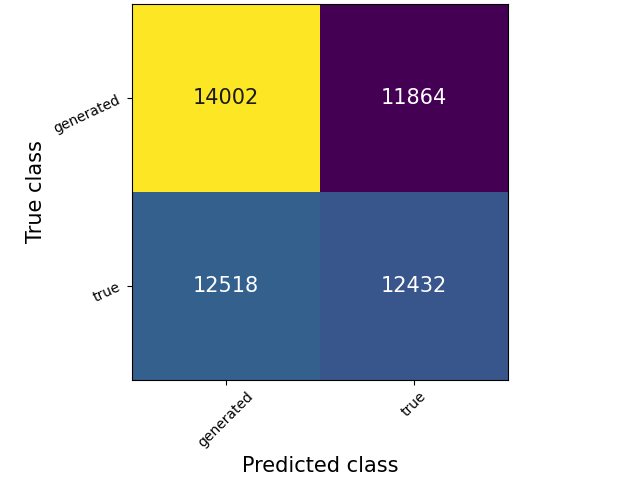} & \includegraphics[width=\hsize]{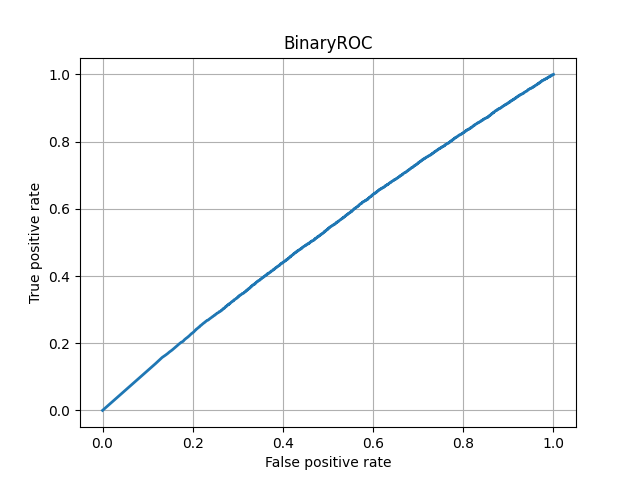}\\
         $45$  & \includegraphics[width=\hsize]{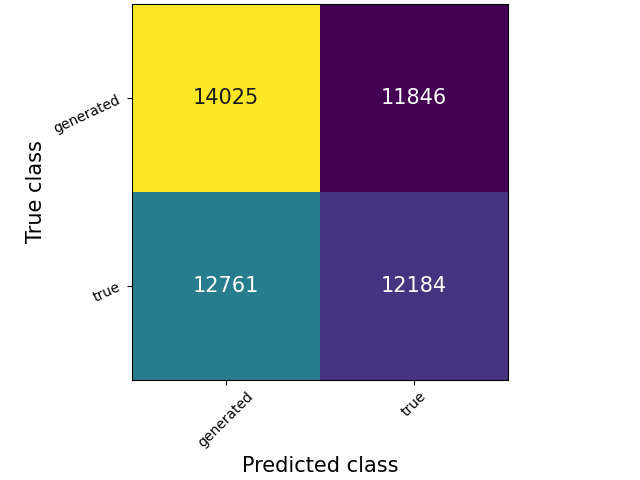} & \includegraphics[width=\hsize]{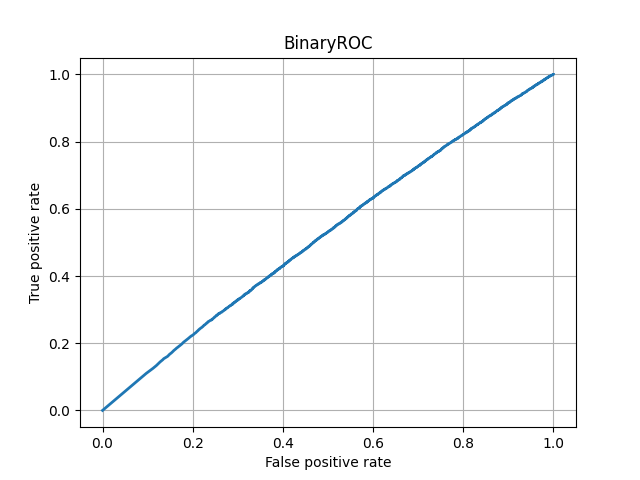}\\
         $46$  & \includegraphics[width=\hsize]{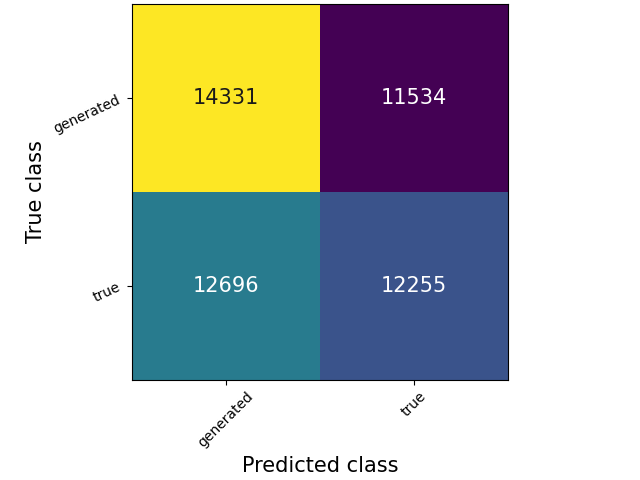} & \includegraphics[width=\hsize]{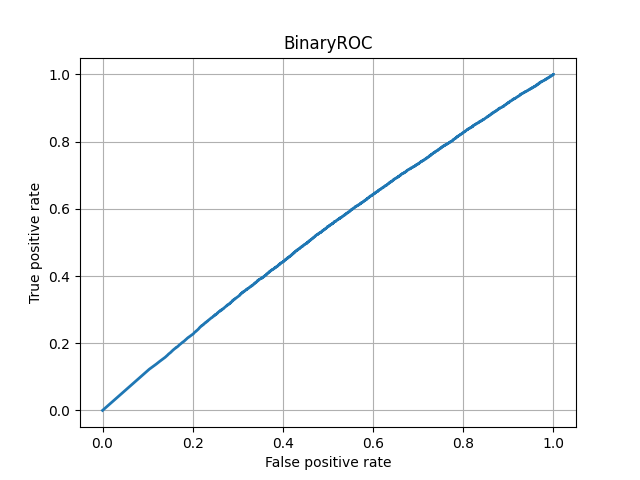}
    \end{tabular}
    \caption{Resnet 101}
    \label{fig:c2st_resnet101}
\end{figure}

In this section, we show the confusion matrix and the Roc curve for the last iteration over the validation set for all the different classifiers (and seeds) trained for the {\c2st} metric
with samplers generated using {\ddpm} with $\enddiff=1000$. The \Cref{fig:c2st_resnet18,fig:c2st_resnet50,fig:c2st_resnet101} show them for Resnet18, Resnet50 and Resnet101 respectively.
\subsection{Validation of posterior samplers:}
\label{app:additional_exps:posteriorsampling}

\paragraph{Definition of the inverse problems:} We start by generating 3 samples from the Global Anisotropy GRF prior defined  in \Cref{app:num:fine_tuning}, used to represent 3 \q{variables} defined across the spatial domain $\cD$. We refer to these variables through an index: $0$, $1$, or $4$. Based on these 3 variables, we define 8 inpainting inverse problems, by subsampling the variables. We use two approaches to subsample the variables. The first approach, which we refer to as \q{unif},  consists of drawing observation locations uniformly across the domain $\cD$. The second approach, which we refer to as \q{clust},  consists of drawing clustered observation locations across the domain $\cD$. This is done by simulating a Poisson Cluster Process across $\cD$, with a mean number of clusters of $10$, and points clustered uniformly on circles of radius $0.1$. 

The three variables generated at the beginning are subsampled as follows, to generate in total 8 inverse problems:
\begin{itemize}
	\item The variable $0$ is subsampled with a \q{unif} mask with 300 points, and a \q{clust} mask with 364 points,
	\item The variable $1$ is subsampled with a \q{unif} mask with 300 points, a \q{unif} mask with 600 points, a \q{clust} mask with 229 points, and  a \q{clust} mask with 355 points,
	\item The variable $4$ is subsampled with a \q{unif} mask with 600 points, and a \q{clust} mask with 612 points.
\end{itemize}
An independent (centered) Gaussian measurement noise with standard deviation 0.01 is added to each of these observations.

\begin{figure}
    \begin{tabular}{
        M{0.1\linewidth}@{\hspace{0\tabcolsep}}
        M{0.13\linewidth}@{\hspace{0.08\tabcolsep}}
        M{0.13\linewidth}@{\hspace{0.08\tabcolsep}}
        M{0.13\linewidth}@{\hspace{0.08\tabcolsep}}
        M{0.13\linewidth}@{\hspace{0.08\tabcolsep}}
        M{0.13\linewidth}@{\hspace{0.08\tabcolsep}}
        M{0.13\linewidth}@{\hspace{0.08\tabcolsep}}
        M{0.13\linewidth}@{\hspace{0.08\tabcolsep}}
        }
         sampler & observation & standard deviation & sample & sample & sample & sample & error \\
         MCMC  
         & \includegraphics[width=\hsize]{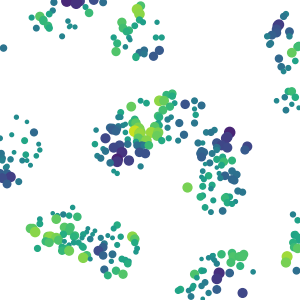} 
         & \includegraphics[width=\hsize]{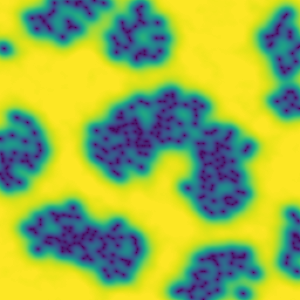}
         & \includegraphics[width=\hsize]{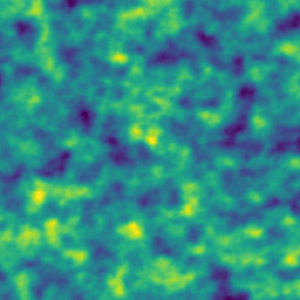}
         & \includegraphics[width=\hsize]{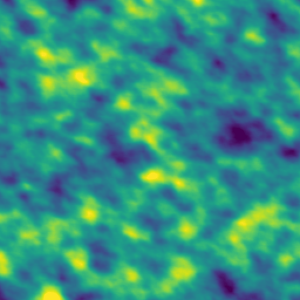}
         & \includegraphics[width=\hsize]{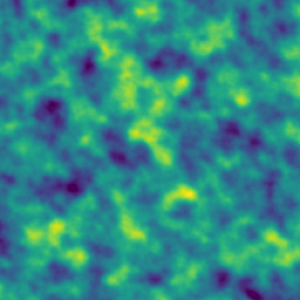}
         & \includegraphics[width=\hsize]{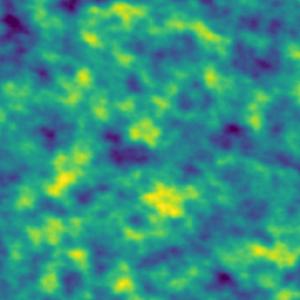}
         & \includegraphics[width=\hsize]{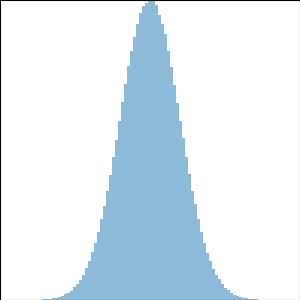}\\\addlinespace[-3pt]
         \dps
         & \includegraphics[width=\hsize]{images/posterior_samples/ganiso_img_0_size_2_clust/obs.png} 
         & \includegraphics[width=\hsize]{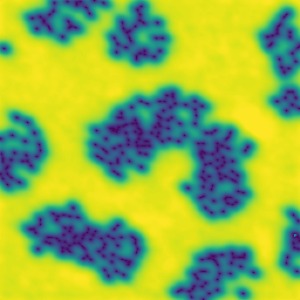}
         & \includegraphics[width=\hsize]{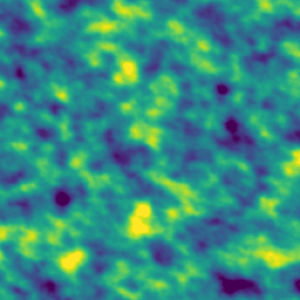}
         & \includegraphics[width=\hsize]{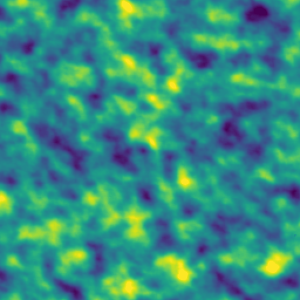}
         & \includegraphics[width=\hsize]{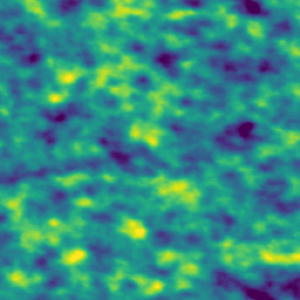}
         & \includegraphics[width=\hsize]{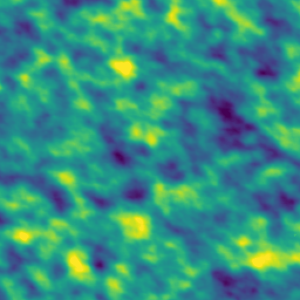}
         & \includegraphics[width=\hsize]{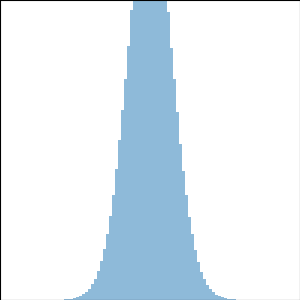}\\\addlinespace[-3pt]
         \mgdm
         & \includegraphics[width=\hsize]{images/posterior_samples/ganiso_img_0_size_2_clust/obs.png} 
         & \includegraphics[width=\hsize]{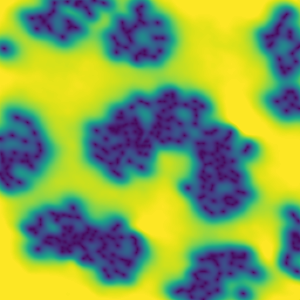}
         & \includegraphics[width=\hsize]{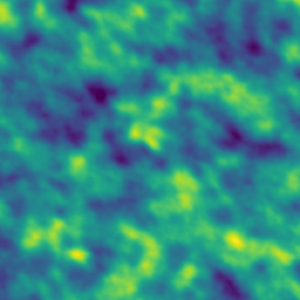}
         & \includegraphics[width=\hsize]{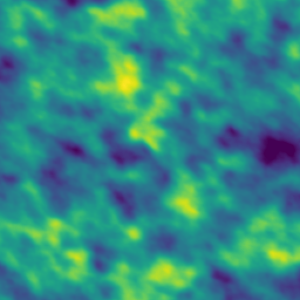}
         & \includegraphics[width=\hsize]{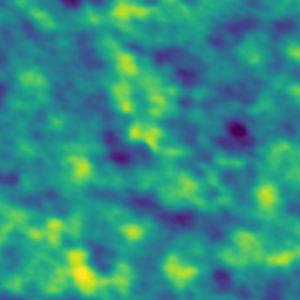}
         & \includegraphics[width=\hsize]{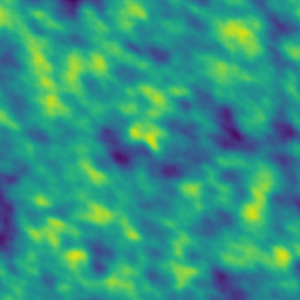}
         & \includegraphics[width=\hsize]{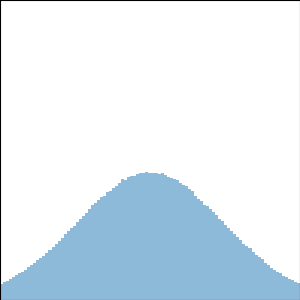}\\\addlinespace[-3pt]
         \mgps
         & \includegraphics[width=\hsize]{images/posterior_samples/ganiso_img_0_size_2_clust/obs.png} 
         & \includegraphics[width=\hsize]{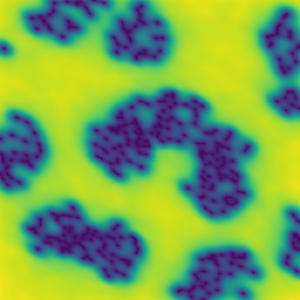}
         & \includegraphics[width=\hsize]{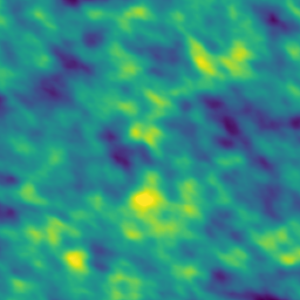}
         & \includegraphics[width=\hsize]{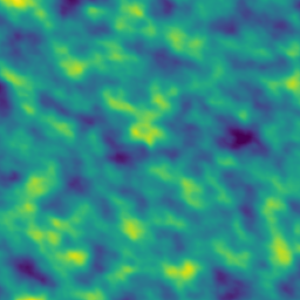}
         & \includegraphics[width=\hsize]{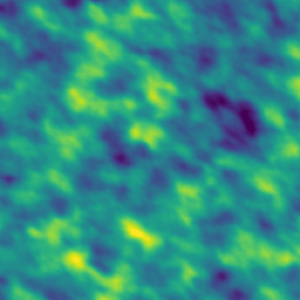}
         & \includegraphics[width=\hsize]{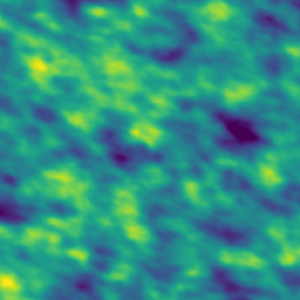}
         & \includegraphics[width=\hsize]{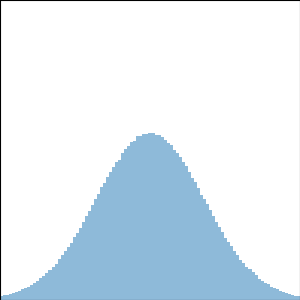}\\\addlinespace[-3pt]
    \end{tabular}
    \caption{Samples visualization for data index $0$ with $\measdim = 364$ and mask-type "clust", corresponding to the first line of \Cref{table:posterior_sampling}.}
    \label{fig:img_0_size_2_clust:show}
\end{figure}

\begin{figure}
    \begin{tabular}{
        M{0.1\linewidth}@{\hspace{0\tabcolsep}}
        M{0.13\linewidth}@{\hspace{0.08\tabcolsep}}
        M{0.13\linewidth}@{\hspace{0.08\tabcolsep}}
        M{0.13\linewidth}@{\hspace{0.08\tabcolsep}}
        M{0.13\linewidth}@{\hspace{0.08\tabcolsep}}
        M{0.13\linewidth}@{\hspace{0.08\tabcolsep}}
        M{0.13\linewidth}@{\hspace{0.08\tabcolsep}}
        M{0.13\linewidth}@{\hspace{0.08\tabcolsep}}
        }
         sampler & observation & standard deviation & sample & sample & sample & sample & error \\
         MCMC  
         & \includegraphics[width=\hsize]{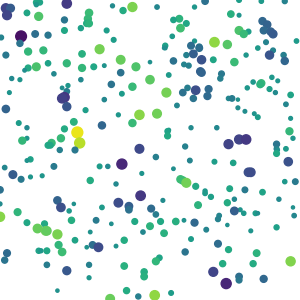} 
         & \includegraphics[width=\hsize]{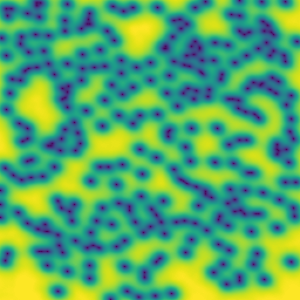}
         & \includegraphics[width=\hsize]{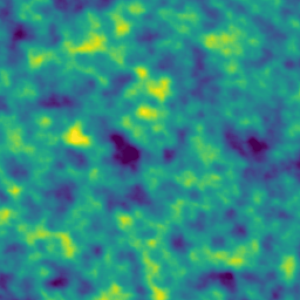}
         & \includegraphics[width=\hsize]{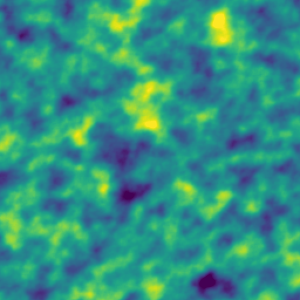}
         & \includegraphics[width=\hsize]{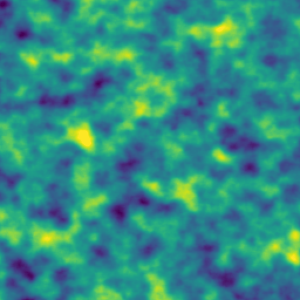}
         & \includegraphics[width=\hsize]{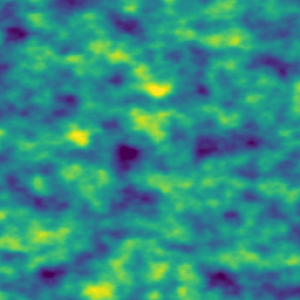}
         & \includegraphics[width=\hsize]{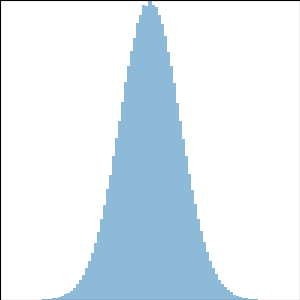}\\\addlinespace[-3pt]
         \dps
         & \includegraphics[width=\hsize]{images/posterior_samples/ganiso_img_0_size_2_unif/obs.png} 
         & \includegraphics[width=\hsize]{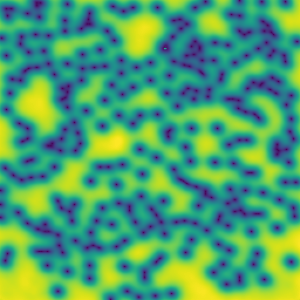}
         & \includegraphics[width=\hsize]{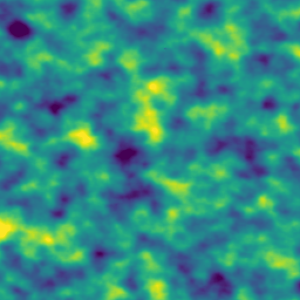}
         & \includegraphics[width=\hsize]{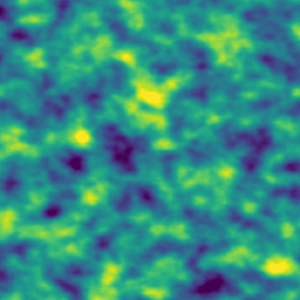}
         & \includegraphics[width=\hsize]{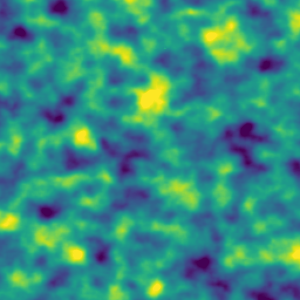}
         & \includegraphics[width=\hsize]{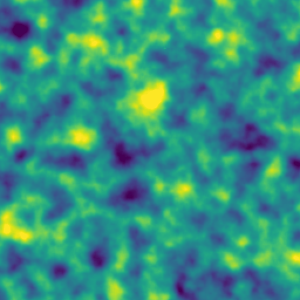}
         & \includegraphics[width=\hsize]{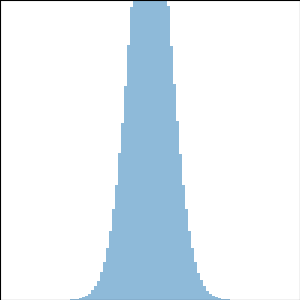}\\\addlinespace[-3pt]
         \mgdm
         & \includegraphics[width=\hsize]{images/posterior_samples/ganiso_img_0_size_2_unif/obs.png} 
         & \includegraphics[width=\hsize]{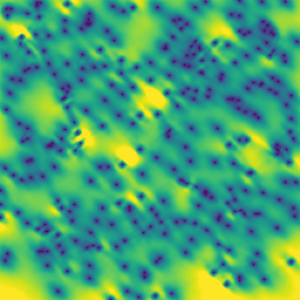}
         & \includegraphics[width=\hsize]{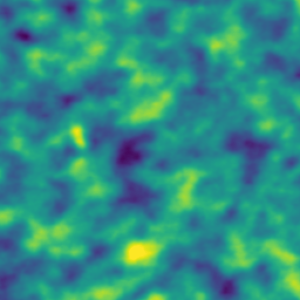}
         & \includegraphics[width=\hsize]{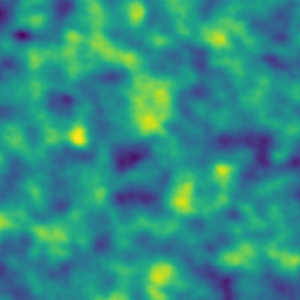}
         & \includegraphics[width=\hsize]{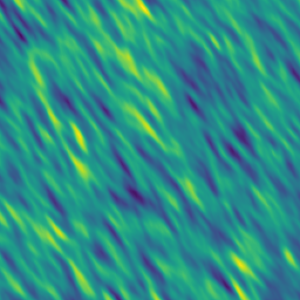}
         & \includegraphics[width=\hsize]{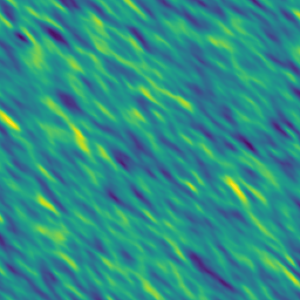}
         & \includegraphics[width=\hsize]{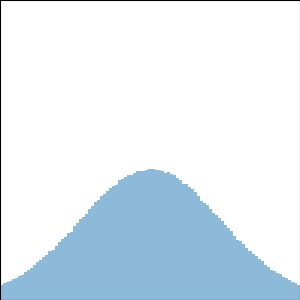}\\\addlinespace[-3pt]
         \mgps
         & \includegraphics[width=\hsize]{images/posterior_samples/ganiso_img_0_size_2_unif/obs.png} 
         & \includegraphics[width=\hsize]{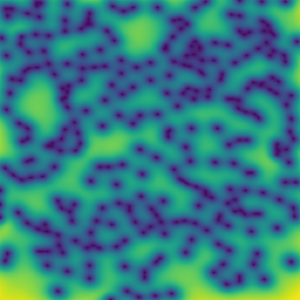}
         & \includegraphics[width=\hsize]{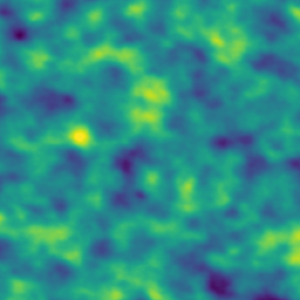}
         & \includegraphics[width=\hsize]{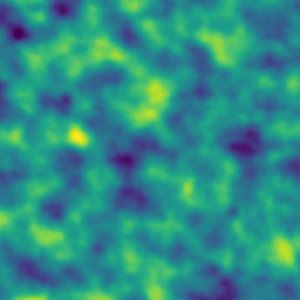}
         & \includegraphics[width=\hsize]{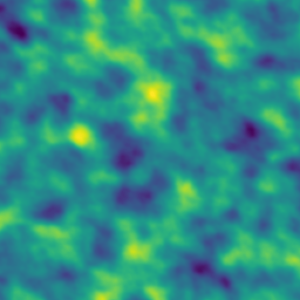}
         & \includegraphics[width=\hsize]{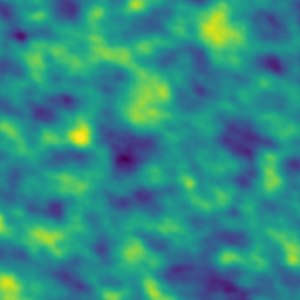}
         & \includegraphics[width=\hsize]{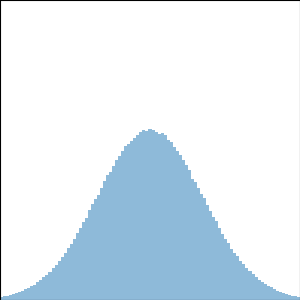}\\\addlinespace[-3pt]
    \end{tabular}
    \caption{Samples visualization for data index $0$ with $\measdim = 300$ and mask-type "unif", corresponding to the second line of \Cref{table:posterior_sampling}.}
    \label{fig:img_0_size_2_unif:show}
\end{figure}

\begin{figure}
    \begin{tabular}{
        M{0.1\linewidth}@{\hspace{0\tabcolsep}}
        M{0.13\linewidth}@{\hspace{0.08\tabcolsep}}
        M{0.13\linewidth}@{\hspace{0.08\tabcolsep}}
        M{0.13\linewidth}@{\hspace{0.08\tabcolsep}}
        M{0.13\linewidth}@{\hspace{0.08\tabcolsep}}
        M{0.13\linewidth}@{\hspace{0.08\tabcolsep}}
        M{0.13\linewidth}@{\hspace{0.08\tabcolsep}}
        M{0.13\linewidth}@{\hspace{0.08\tabcolsep}}
        }
         sampler & observation & standard deviation & sample & sample & sample & sample & error \\
         MCMC  
         & \includegraphics[width=\hsize]{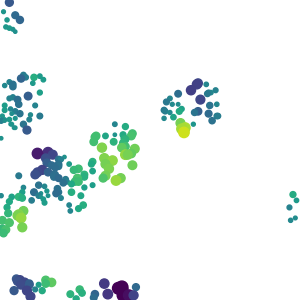} 
         & \includegraphics[width=\hsize]{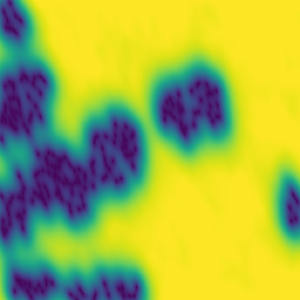}
         & \includegraphics[width=\hsize]{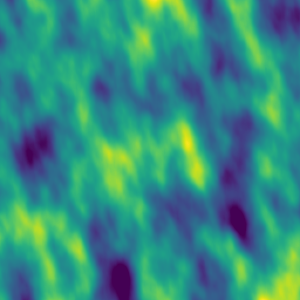}
         & \includegraphics[width=\hsize]{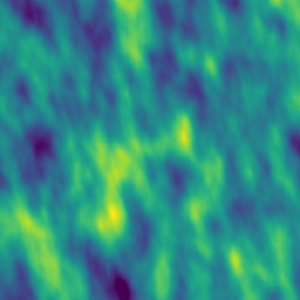}
         & \includegraphics[width=\hsize]{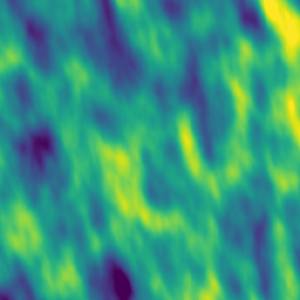}
         & \includegraphics[width=\hsize]{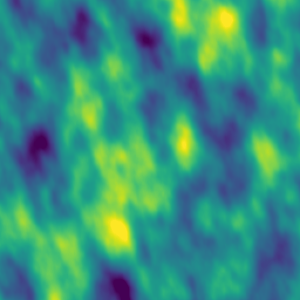}
         & \includegraphics[width=\hsize]{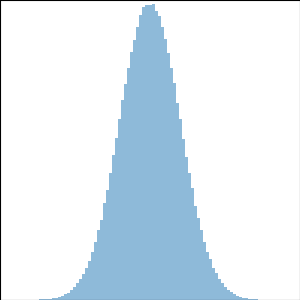}\\\addlinespace[-3pt]
         \dps
         & \includegraphics[width=\hsize]{images/posterior_samples/ganiso_img_1_size_2_clust/obs.png} 
         & \includegraphics[width=\hsize]{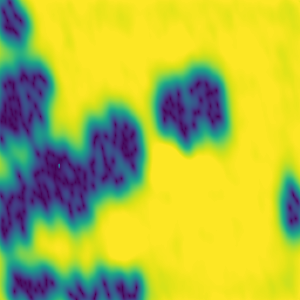}
         & \includegraphics[width=\hsize]{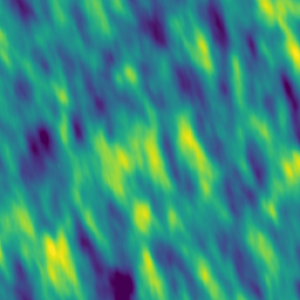}
         & \includegraphics[width=\hsize]{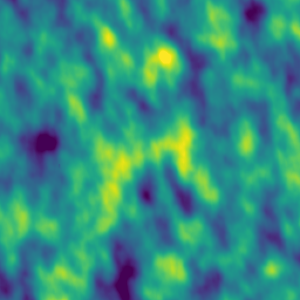}
         & \includegraphics[width=\hsize]{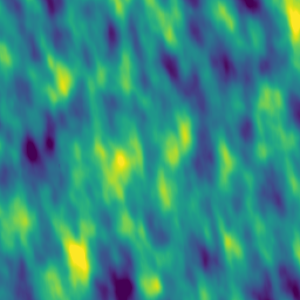}
         & \includegraphics[width=\hsize]{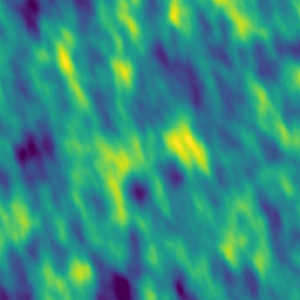}
         & \includegraphics[width=\hsize]{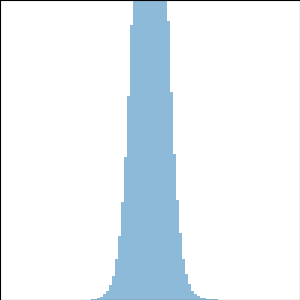}\\\addlinespace[-3pt]
         \mgdm
         & \includegraphics[width=\hsize]{images/posterior_samples/ganiso_img_1_size_2_clust/obs.png} 
         & \includegraphics[width=\hsize]{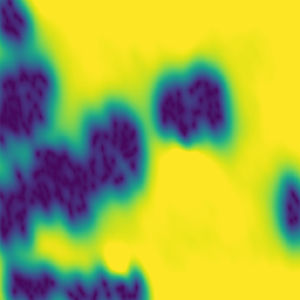}
         & \includegraphics[width=\hsize]{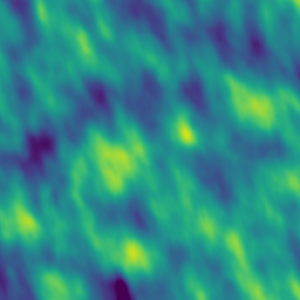}
         & \includegraphics[width=\hsize]{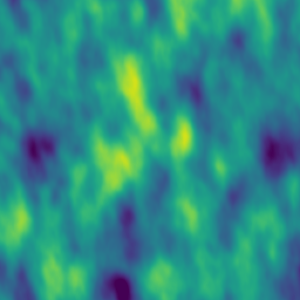}
         & \includegraphics[width=\hsize]{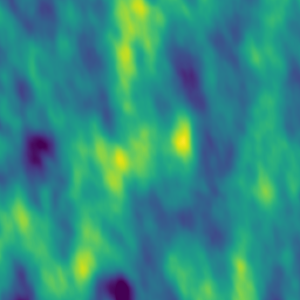}
         & \includegraphics[width=\hsize]{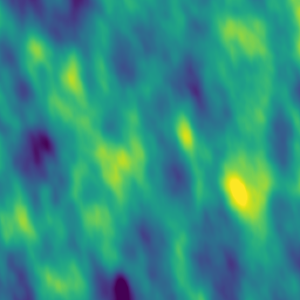}
         & \includegraphics[width=\hsize]{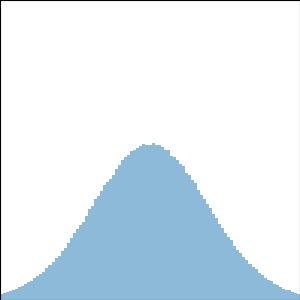}\\\addlinespace[-3pt]
         \mgps
         & \includegraphics[width=\hsize]{images/posterior_samples/ganiso_img_1_size_2_clust/obs.png} 
         & \includegraphics[width=\hsize]{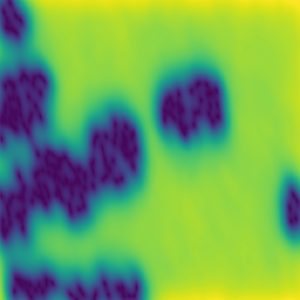}
         & \includegraphics[width=\hsize]{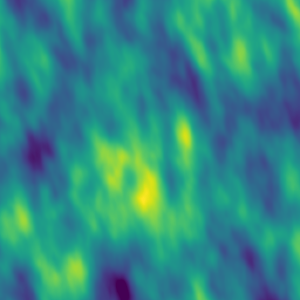}
         & \includegraphics[width=\hsize]{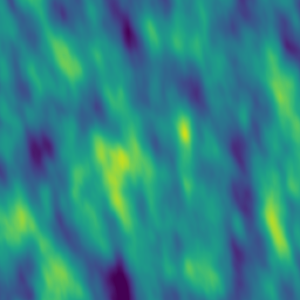}
         & \includegraphics[width=\hsize]{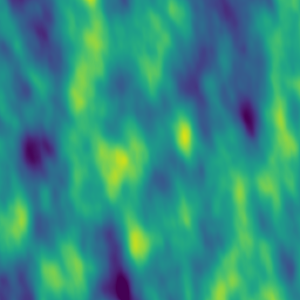}
         & \includegraphics[width=\hsize]{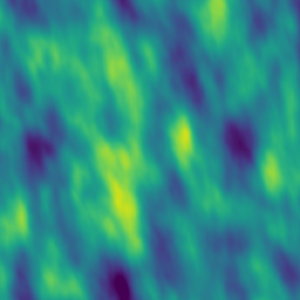}
         & \includegraphics[width=\hsize]{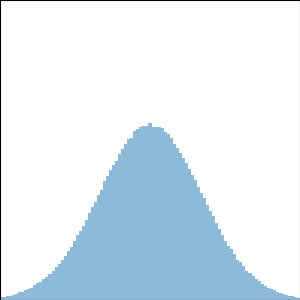}\\\addlinespace[-3pt]
    \end{tabular}
    \caption{Samples visualization for data index $1$ with $\measdim = 229$ and mask-type "clust", corresponding to the third line of \Cref{table:posterior_sampling}.}
    \label{fig:img_1_size_2_clust:show}
\end{figure}

\begin{figure}
    \begin{tabular}{
        M{0.1\linewidth}@{\hspace{0\tabcolsep}}
        M{0.13\linewidth}@{\hspace{0.08\tabcolsep}}
        M{0.13\linewidth}@{\hspace{0.08\tabcolsep}}
        M{0.13\linewidth}@{\hspace{0.08\tabcolsep}}
        M{0.13\linewidth}@{\hspace{0.08\tabcolsep}}
        M{0.13\linewidth}@{\hspace{0.08\tabcolsep}}
        M{0.13\linewidth}@{\hspace{0.08\tabcolsep}}
        M{0.13\linewidth}@{\hspace{0.08\tabcolsep}}
        }
         sampler & observation & standard deviation & sample & sample & sample & sample & error \\
         MCMC  
         & \includegraphics[width=\hsize]{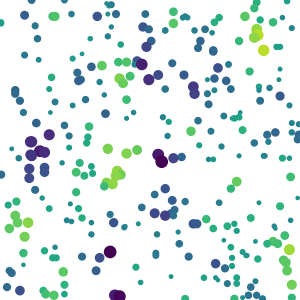} 
         & \includegraphics[width=\hsize]{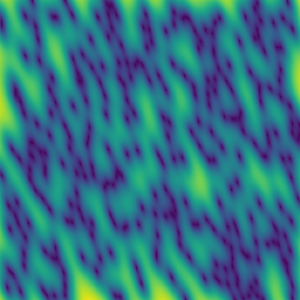}
         & \includegraphics[width=\hsize]{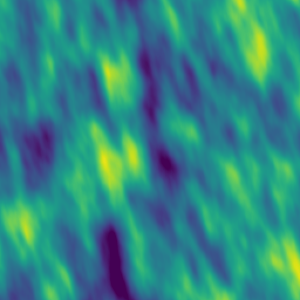}
         & \includegraphics[width=\hsize]{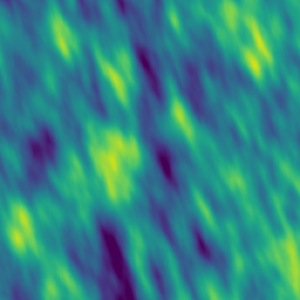}
         & \includegraphics[width=\hsize]{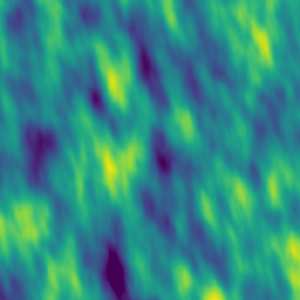}
         & \includegraphics[width=\hsize]{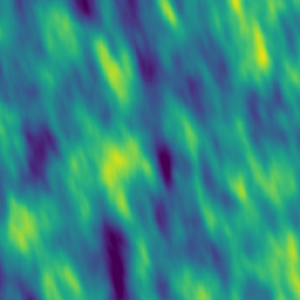}
         & \includegraphics[width=\hsize]{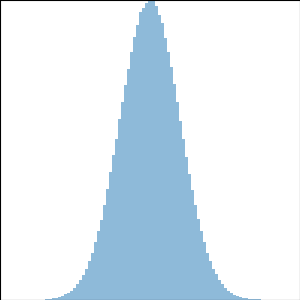}\\\addlinespace[-3pt]
         \dps
         & \includegraphics[width=\hsize]{images/posterior_samples/ganiso_img_1_size_2_unif/obs.png} 
         & \includegraphics[width=\hsize]{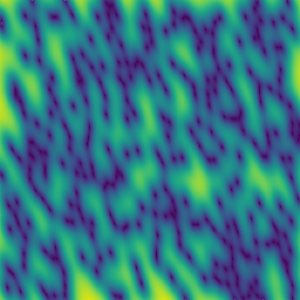}
         & \includegraphics[width=\hsize]{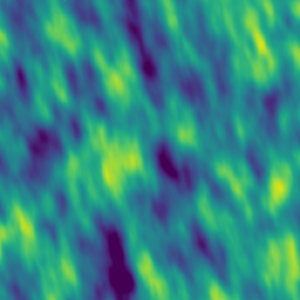}
         & \includegraphics[width=\hsize]{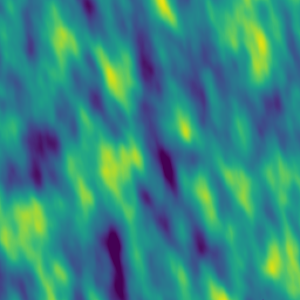}
         & \includegraphics[width=\hsize]{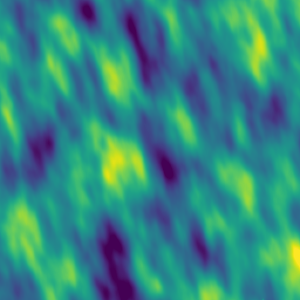}
         & \includegraphics[width=\hsize]{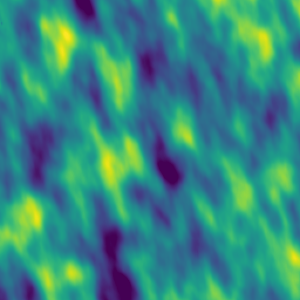}
         & \includegraphics[width=\hsize]{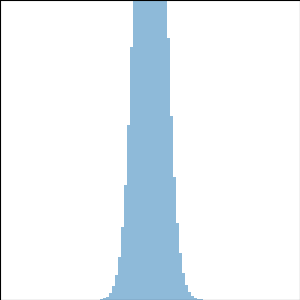}\\\addlinespace[-3pt]
         \mgdm
         & \includegraphics[width=\hsize]{images/posterior_samples/ganiso_img_1_size_2_unif/obs.png} 
         & \includegraphics[width=\hsize]{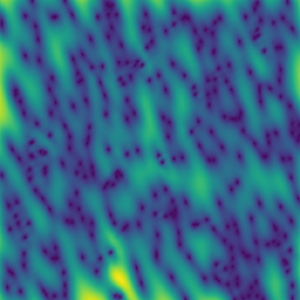}
         & \includegraphics[width=\hsize]{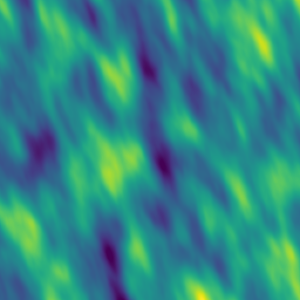}
         & \includegraphics[width=\hsize]{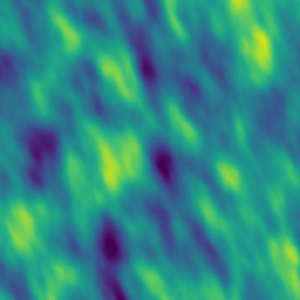}
         & \includegraphics[width=\hsize]{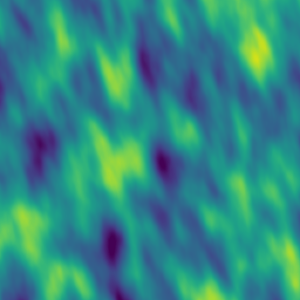}
         & \includegraphics[width=\hsize]{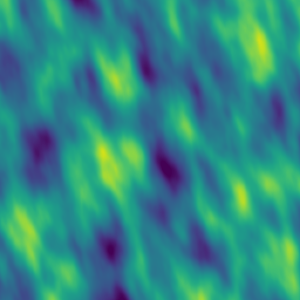}
         & \includegraphics[width=\hsize]{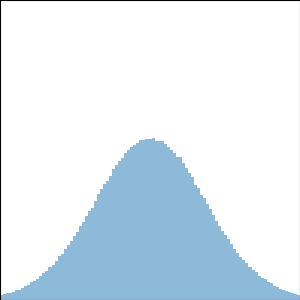}\\\addlinespace[-3pt]
         \mgps
         & \includegraphics[width=\hsize]{images/posterior_samples/ganiso_img_1_size_2_unif/obs.png} 
         & \includegraphics[width=\hsize]{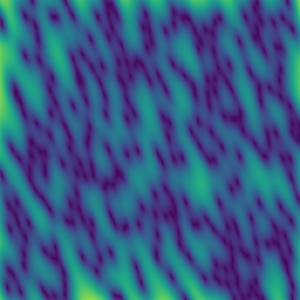}
         & \includegraphics[width=\hsize]{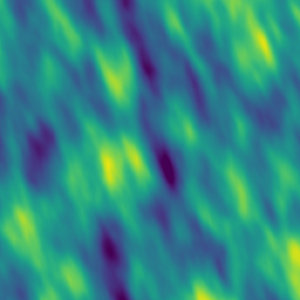}
         & \includegraphics[width=\hsize]{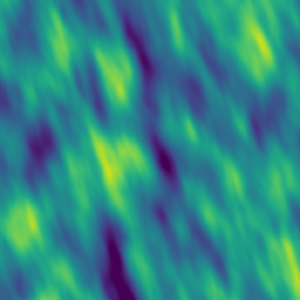}
         & \includegraphics[width=\hsize]{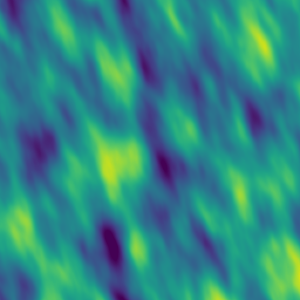}
         & \includegraphics[width=\hsize]{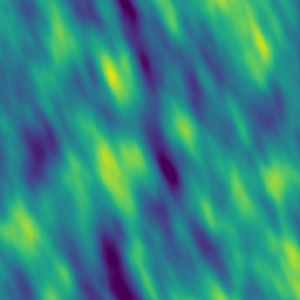}
         & \includegraphics[width=\hsize]{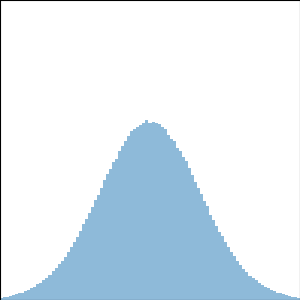}\\\addlinespace[-3pt]
    \end{tabular}
    \caption{Samples visualization for data index $1$ with $\measdim = 300$ and mask-type "unif", corresponding to the fourth line of \Cref{table:posterior_sampling}.}
    \label{fig:img_1_size_2_unif:show}
\end{figure}

\begin{figure}
    \begin{tabular}{
        M{0.1\linewidth}@{\hspace{0\tabcolsep}}
        M{0.13\linewidth}@{\hspace{0.08\tabcolsep}}
        M{0.13\linewidth}@{\hspace{0.08\tabcolsep}}
        M{0.13\linewidth}@{\hspace{0.08\tabcolsep}}
        M{0.13\linewidth}@{\hspace{0.08\tabcolsep}}
        M{0.13\linewidth}@{\hspace{0.08\tabcolsep}}
        M{0.13\linewidth}@{\hspace{0.08\tabcolsep}}
        M{0.13\linewidth}@{\hspace{0.08\tabcolsep}}
        }
         sampler & observation & standard deviation & sample & sample & sample & sample & error \\
         MCMC  
         & \includegraphics[width=\hsize]{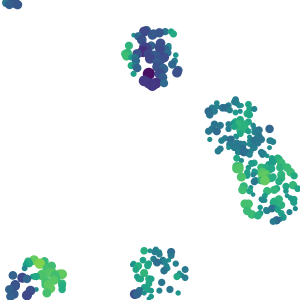} 
         & \includegraphics[width=\hsize]{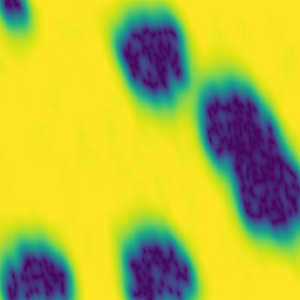}
         & \includegraphics[width=\hsize]{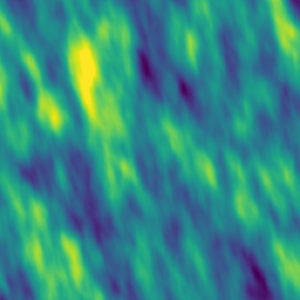}
         & \includegraphics[width=\hsize]{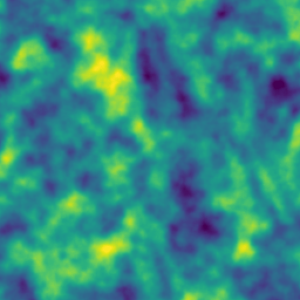}
         & \includegraphics[width=\hsize]{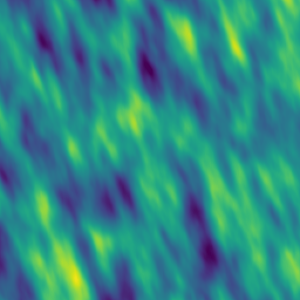}
         & \includegraphics[width=\hsize]{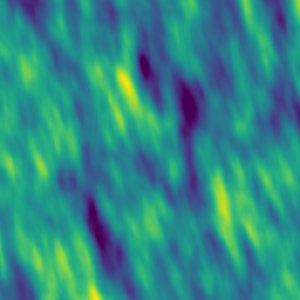}
         & \includegraphics[width=\hsize]{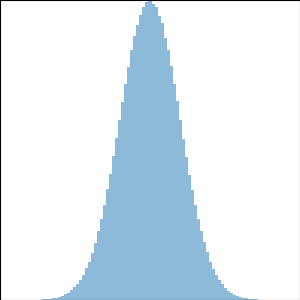}\\\addlinespace[-3pt]
         \dps
         & \includegraphics[width=\hsize]{images/posterior_samples/ganiso_img_1_size_3_clust/obs.png} 
         & \includegraphics[width=\hsize]{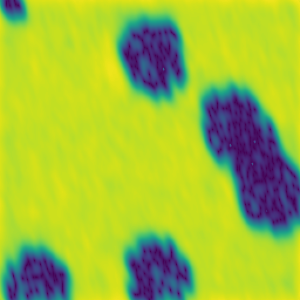}
         & \includegraphics[width=\hsize]{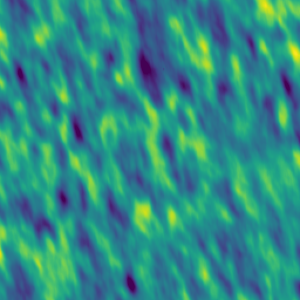}
         & \includegraphics[width=\hsize]{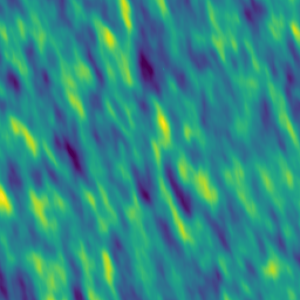}
         & \includegraphics[width=\hsize]{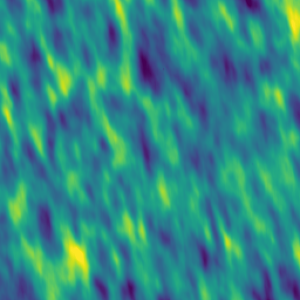}
         & \includegraphics[width=\hsize]{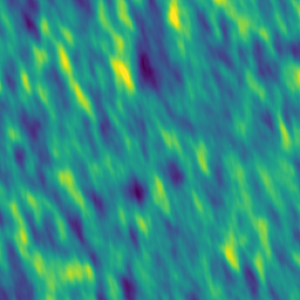}
         & \includegraphics[width=\hsize]{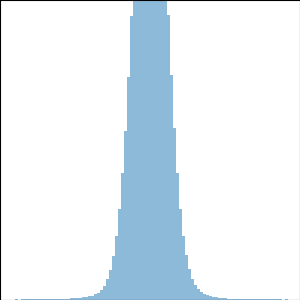}\\\addlinespace[-3pt]
         \mgdm
         & \includegraphics[width=\hsize]{images/posterior_samples/ganiso_img_1_size_3_clust/obs.png} 
         & \includegraphics[width=\hsize]{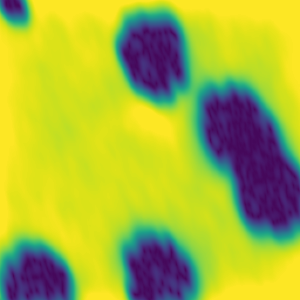}
         & \includegraphics[width=\hsize]{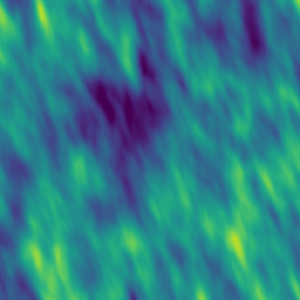}
         & \includegraphics[width=\hsize]{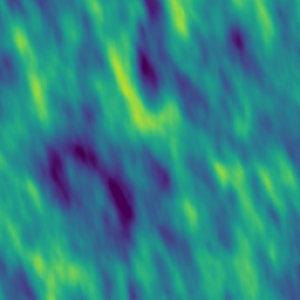}
         & \includegraphics[width=\hsize]{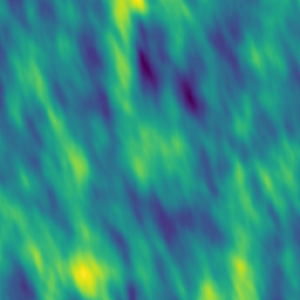}
         & \includegraphics[width=\hsize]{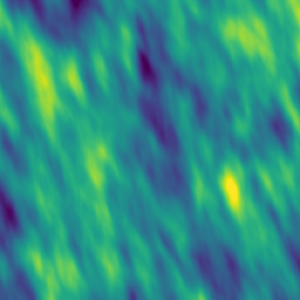}
         & \includegraphics[width=\hsize]{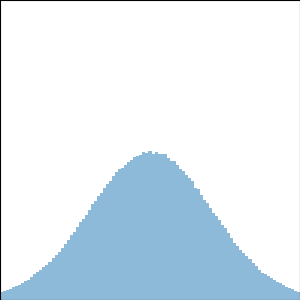}\\\addlinespace[-3pt]
         \mgps
         & \includegraphics[width=\hsize]{images/posterior_samples/ganiso_img_1_size_3_clust/obs.png} 
         & \includegraphics[width=\hsize]{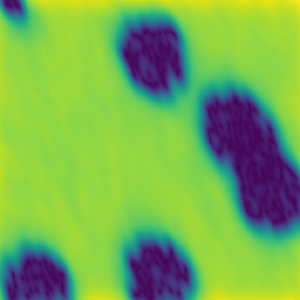}
         & \includegraphics[width=\hsize]{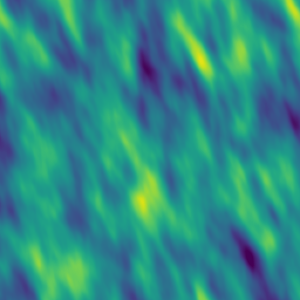}
         & \includegraphics[width=\hsize]{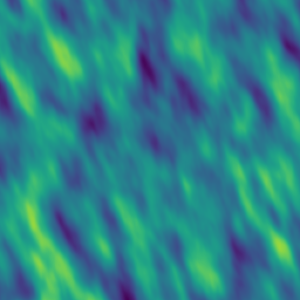}
         & \includegraphics[width=\hsize]{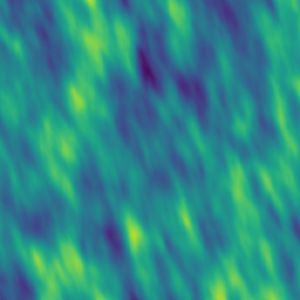}
         & \includegraphics[width=\hsize]{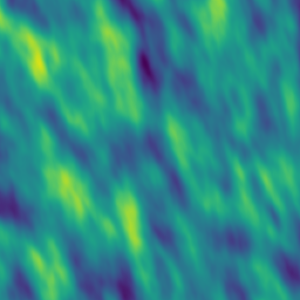}
         & \includegraphics[width=\hsize]{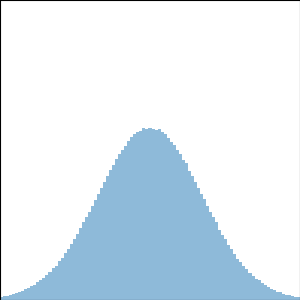}\\\addlinespace[-3pt]
    \end{tabular}
    \caption{Samples visualization for data index $1$ with $\measdim = 355$ and mask-type "clust", corresponding to the fifth line of \Cref{table:posterior_sampling}.}
    \label{fig:img_1_size_3_clust:show}
\end{figure}

\begin{figure}
    \begin{tabular}{
        M{0.1\linewidth}@{\hspace{0\tabcolsep}}
        M{0.13\linewidth}@{\hspace{0.08\tabcolsep}}
        M{0.13\linewidth}@{\hspace{0.08\tabcolsep}}
        M{0.13\linewidth}@{\hspace{0.08\tabcolsep}}
        M{0.13\linewidth}@{\hspace{0.08\tabcolsep}}
        M{0.13\linewidth}@{\hspace{0.08\tabcolsep}}
        M{0.13\linewidth}@{\hspace{0.08\tabcolsep}}
        M{0.13\linewidth}@{\hspace{0.08\tabcolsep}}
        }
         sampler & observation & standard deviation & sample & sample & sample & sample & error \\
         MCMC  
         & \includegraphics[width=\hsize]{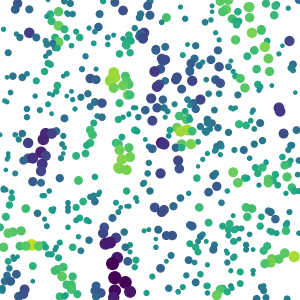} 
         & \includegraphics[width=\hsize]{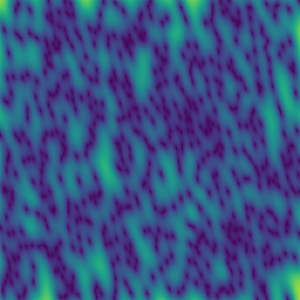}
         & \includegraphics[width=\hsize]{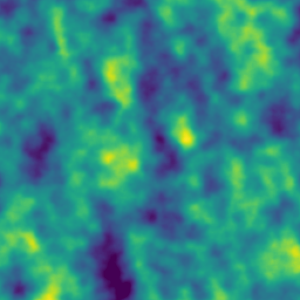}
         & \includegraphics[width=\hsize]{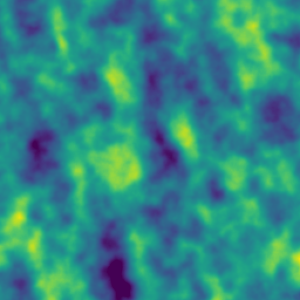}
         & \includegraphics[width=\hsize]{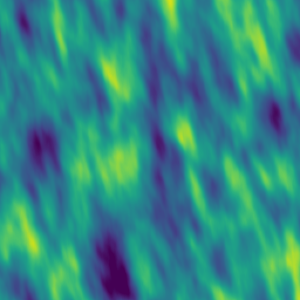}
         & \includegraphics[width=\hsize]{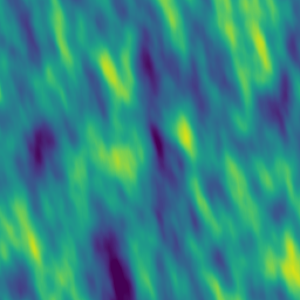}
         & \includegraphics[width=\hsize]{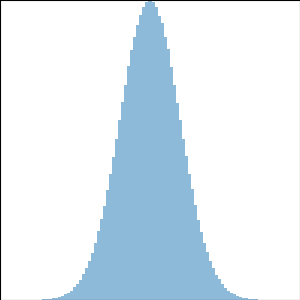}\\\addlinespace[-3pt]
         \dps
         & \includegraphics[width=\hsize]{images/posterior_samples/ganiso_img_1_size_3_unif/obs.png} 
         & \includegraphics[width=\hsize]{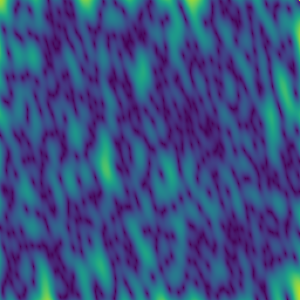}
         & \includegraphics[width=\hsize]{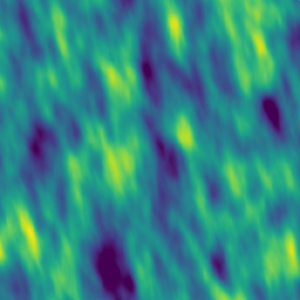}
         & \includegraphics[width=\hsize]{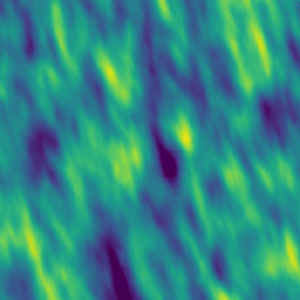}
         & \includegraphics[width=\hsize]{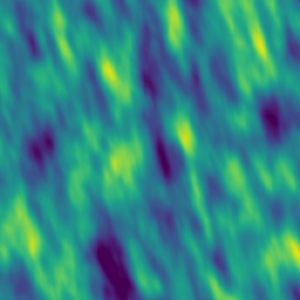}
         & \includegraphics[width=\hsize]{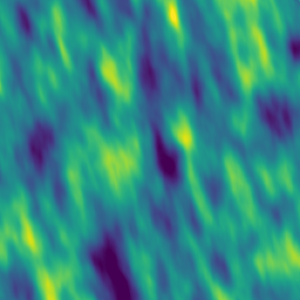}
         & \includegraphics[width=\hsize]{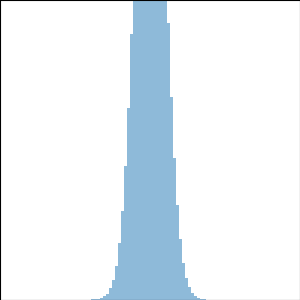}\\\addlinespace[-3pt]
         \mgdm
         & \includegraphics[width=\hsize]{images/posterior_samples/ganiso_img_1_size_3_unif/obs.png} 
         & \includegraphics[width=\hsize]{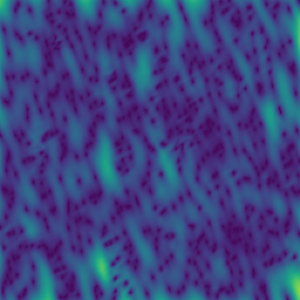}
         & \includegraphics[width=\hsize]{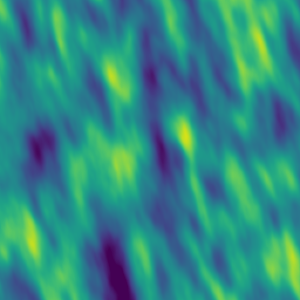}
         & \includegraphics[width=\hsize]{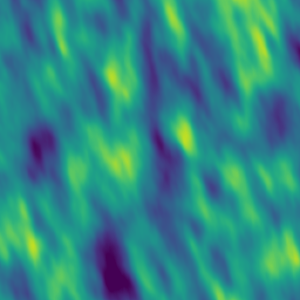}
         & \includegraphics[width=\hsize]{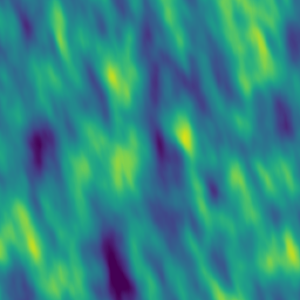}
         & \includegraphics[width=\hsize]{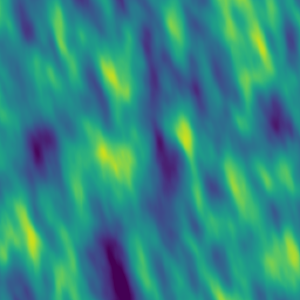}
         & \includegraphics[width=\hsize]{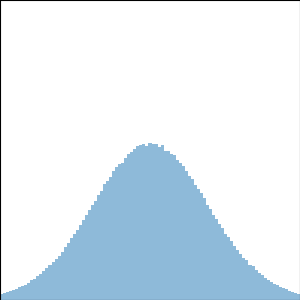}\\\addlinespace[-3pt]
         \mgps
         & \includegraphics[width=\hsize]{images/posterior_samples/ganiso_img_1_size_3_unif/obs.png} 
         & \includegraphics[width=\hsize]{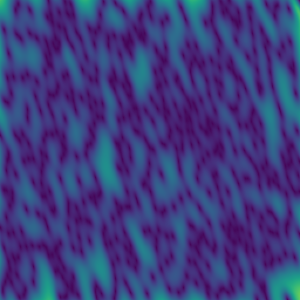}
         & \includegraphics[width=\hsize]{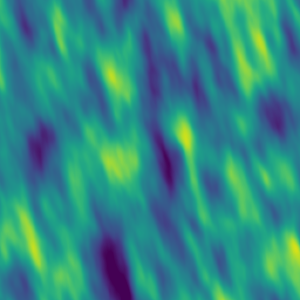}
         & \includegraphics[width=\hsize]{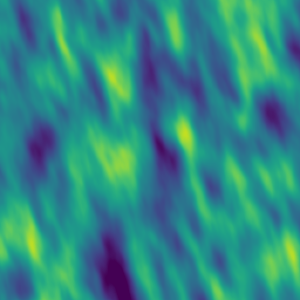}
         & \includegraphics[width=\hsize]{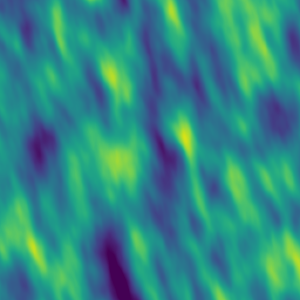}
         & \includegraphics[width=\hsize]{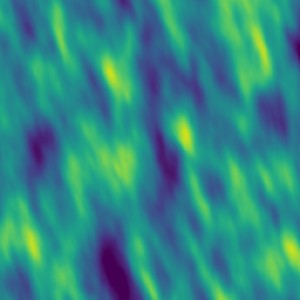}
         & \includegraphics[width=\hsize]{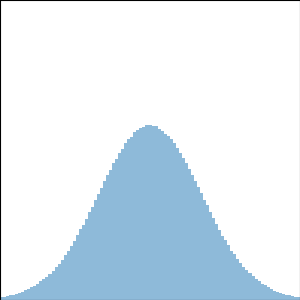}\\\addlinespace[-3pt]
    \end{tabular}
    \caption{Samples visualization for data index $1$ with $\measdim = 600$ and mask-type "unif", corresponding to the sixth line of \Cref{table:posterior_sampling}.}
    \label{fig:img_1_size_3_unif:show}
\end{figure}

\begin{figure}
    \begin{tabular}{
        M{0.1\linewidth}@{\hspace{0\tabcolsep}}
        M{0.13\linewidth}@{\hspace{0.08\tabcolsep}}
        M{0.13\linewidth}@{\hspace{0.08\tabcolsep}}
        M{0.13\linewidth}@{\hspace{0.08\tabcolsep}}
        M{0.13\linewidth}@{\hspace{0.08\tabcolsep}}
        M{0.13\linewidth}@{\hspace{0.08\tabcolsep}}
        M{0.13\linewidth}@{\hspace{0.08\tabcolsep}}
        M{0.13\linewidth}@{\hspace{0.08\tabcolsep}}
        }
         sampler & observation & standard deviation & sample & sample & sample & sample & error \\
         MCMC  
         & \includegraphics[width=\hsize]{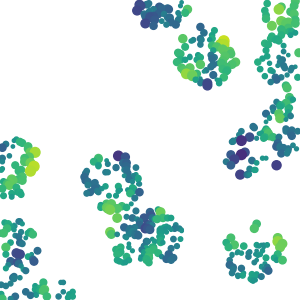} 
         & \includegraphics[width=\hsize]{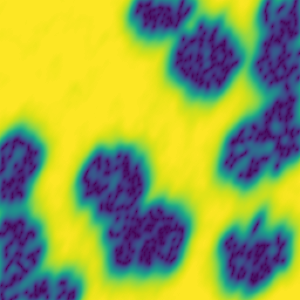}
         & \includegraphics[width=\hsize]{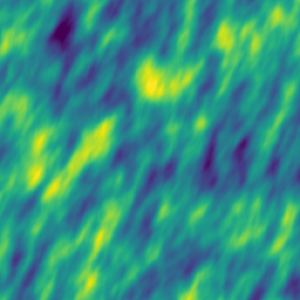}
         & \includegraphics[width=\hsize]{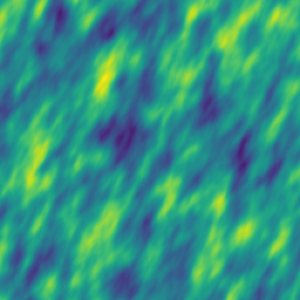}
         & \includegraphics[width=\hsize]{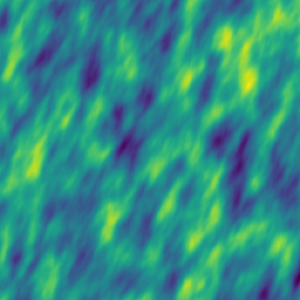}
         & \includegraphics[width=\hsize]{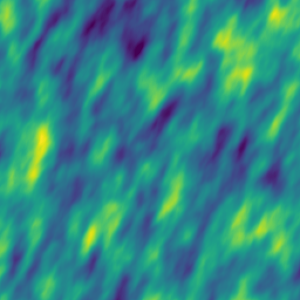}
         & \includegraphics[width=\hsize]{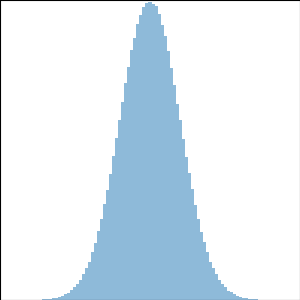}\\\addlinespace[-3pt]
         \dps
         & \includegraphics[width=\hsize]{images/posterior_samples/ganiso_img_4_size_3_clust/mcmc/obs.png} 
         & \includegraphics[width=\hsize]{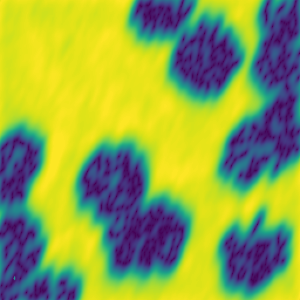}
         & \includegraphics[width=\hsize]{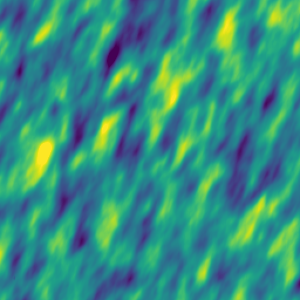}
         & \includegraphics[width=\hsize]{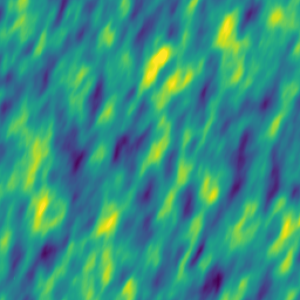}
         & \includegraphics[width=\hsize]{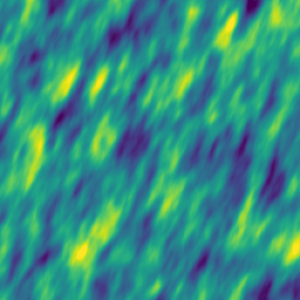}
         & \includegraphics[width=\hsize]{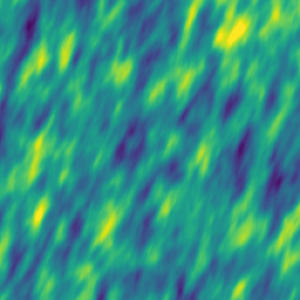}
         & \includegraphics[width=\hsize]{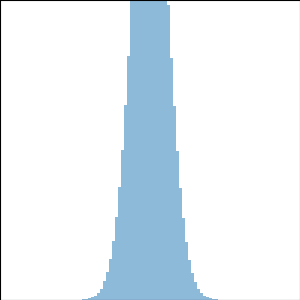}\\\addlinespace[-3pt]
         \mgdm
         & \includegraphics[width=\hsize]{images/posterior_samples/ganiso_img_4_size_3_clust/mcmc/obs.png} 
         & \includegraphics[width=\hsize]{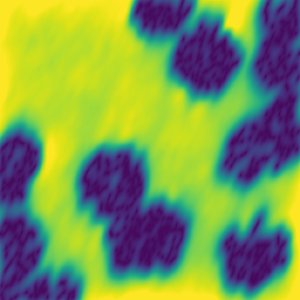}
         & \includegraphics[width=\hsize]{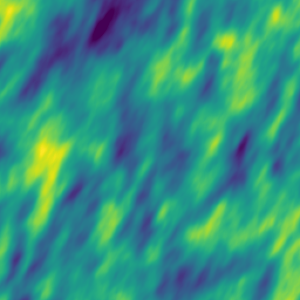}
         & \includegraphics[width=\hsize]{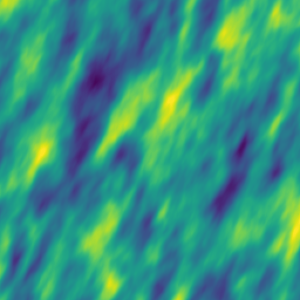}
         & \includegraphics[width=\hsize]{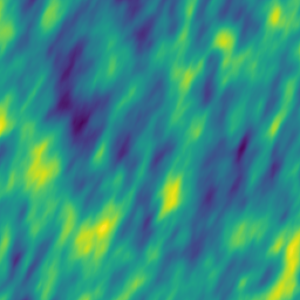}
         & \includegraphics[width=\hsize]{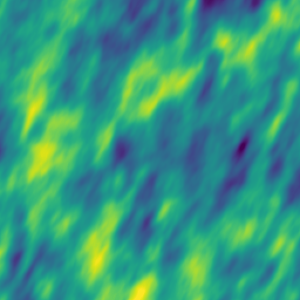}
         & \includegraphics[width=\hsize]{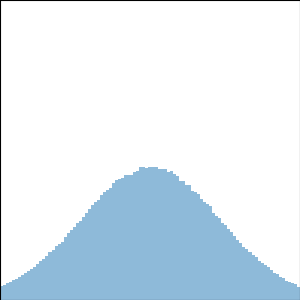}\\\addlinespace[-3pt]
         \mgps
         & \includegraphics[width=\hsize]{images/posterior_samples/ganiso_img_4_size_3_clust/mcmc/obs.png} 
         & \includegraphics[width=\hsize]{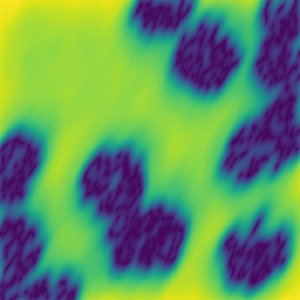}
         & \includegraphics[width=\hsize]{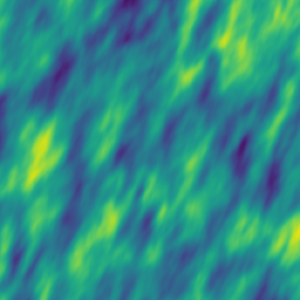}
         & \includegraphics[width=\hsize]{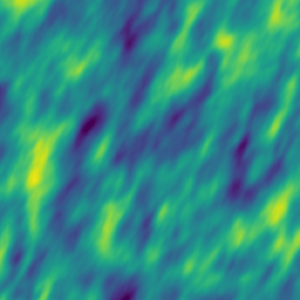}
         & \includegraphics[width=\hsize]{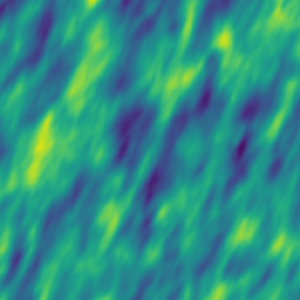}
         & \includegraphics[width=\hsize]{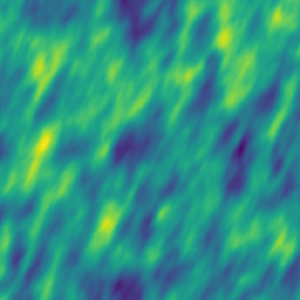}
         & \includegraphics[width=\hsize]{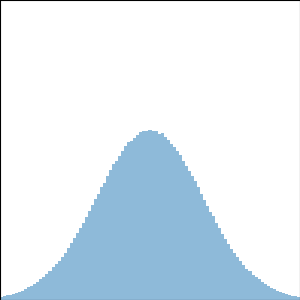}\\\addlinespace[-3pt]
    \end{tabular}
    \caption{Samples visualization for data index $4$ with $\measdim = 612$ and mask-type "clust", corresponding to the seventh line of \Cref{table:posterior_sampling}.}
    \label{fig:img_4_size_3_clust:show}
\end{figure}

\begin{figure}
    \begin{tabular}{
        M{0.1\linewidth}@{\hspace{0\tabcolsep}}
        M{0.13\linewidth}@{\hspace{0.08\tabcolsep}}
        M{0.13\linewidth}@{\hspace{0.08\tabcolsep}}
        M{0.13\linewidth}@{\hspace{0.08\tabcolsep}}
        M{0.13\linewidth}@{\hspace{0.08\tabcolsep}}
        M{0.13\linewidth}@{\hspace{0.08\tabcolsep}}
        M{0.13\linewidth}@{\hspace{0.08\tabcolsep}}
        M{0.13\linewidth}@{\hspace{0.08\tabcolsep}}
        }
         sampler & observation & standard deviation & sample & sample & sample & sample & error \\
         MCMC  
         & \includegraphics[width=\hsize]{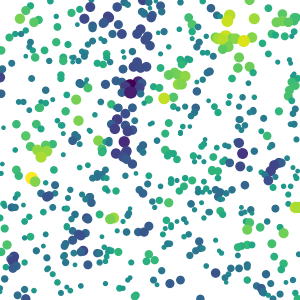} 
         & \includegraphics[width=\hsize]{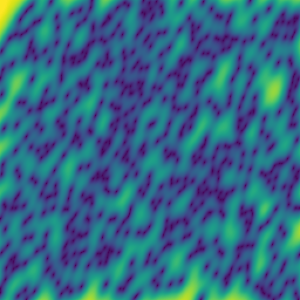}
         & \includegraphics[width=\hsize]{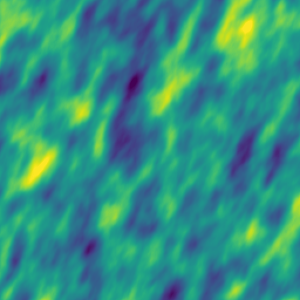}
         & \includegraphics[width=\hsize]{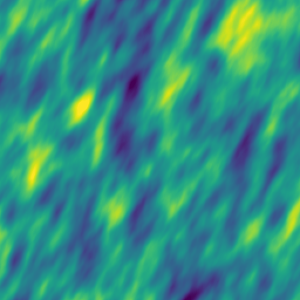}
         & \includegraphics[width=\hsize]{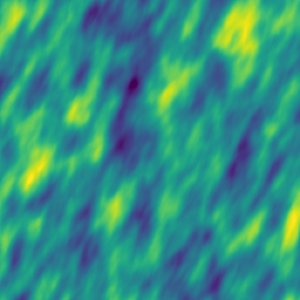}
         & \includegraphics[width=\hsize]{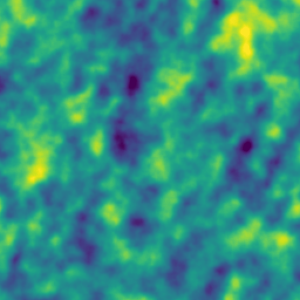}
         & \includegraphics[width=\hsize]{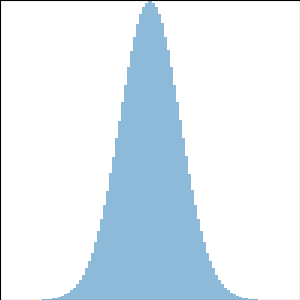}\\\addlinespace[-3pt]
         \dps
         & \includegraphics[width=\hsize]{images/posterior_samples/ganiso_img_4_size_3_unif/mcmc/obs.png} 
         & \includegraphics[width=\hsize]{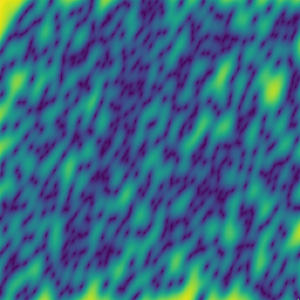}
         & \includegraphics[width=\hsize]{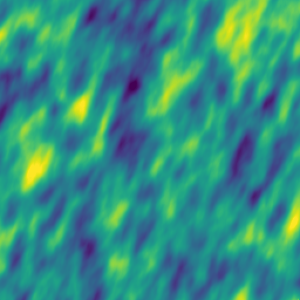}
         & \includegraphics[width=\hsize]{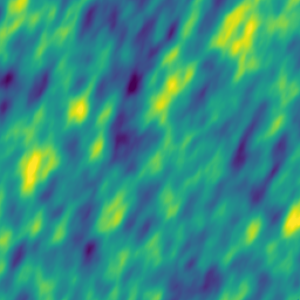}
         & \includegraphics[width=\hsize]{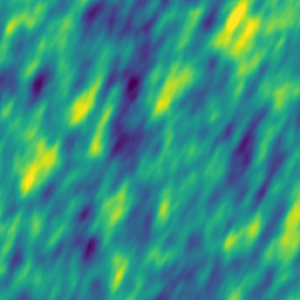}
         & \includegraphics[width=\hsize]{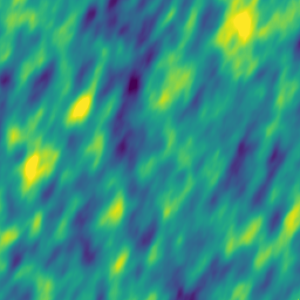}
         & \includegraphics[width=\hsize]{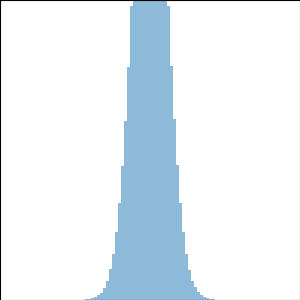}\\\addlinespace[-3pt]
         \mgdm
         & \includegraphics[width=\hsize]{images/posterior_samples/ganiso_img_4_size_3_unif/mcmc/obs.png} 
         & \includegraphics[width=\hsize]{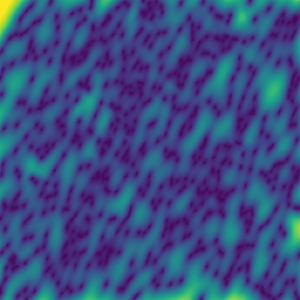}
         & \includegraphics[width=\hsize]{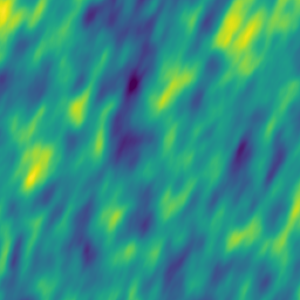}
         & \includegraphics[width=\hsize]{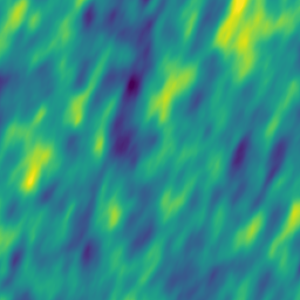}
         & \includegraphics[width=\hsize]{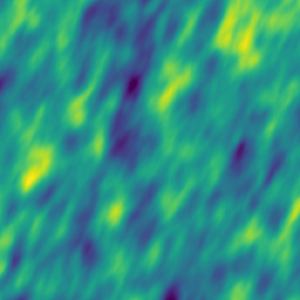}
         & \includegraphics[width=\hsize]{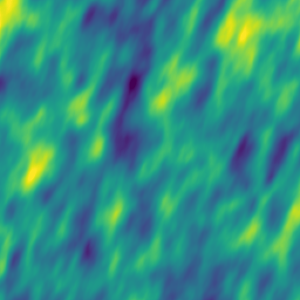}
         & \includegraphics[width=\hsize]{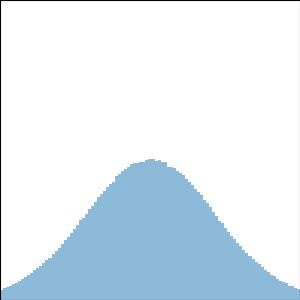}\\\addlinespace[-3pt]
         \mgps
         & \includegraphics[width=\hsize]{images/posterior_samples/ganiso_img_4_size_3_unif/mcmc/obs.png} 
         & \includegraphics[width=\hsize]{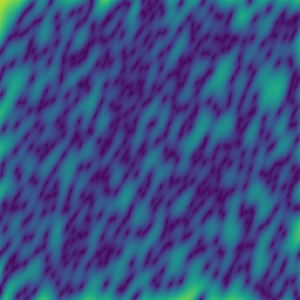}
         & \includegraphics[width=\hsize]{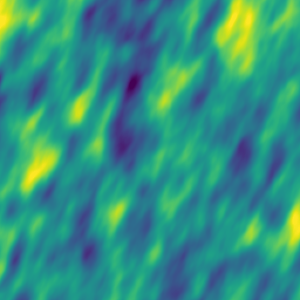}
         & \includegraphics[width=\hsize]{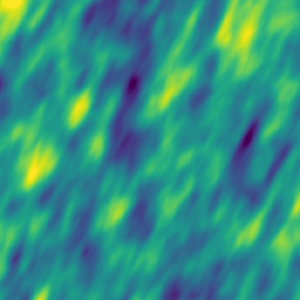}
         & \includegraphics[width=\hsize]{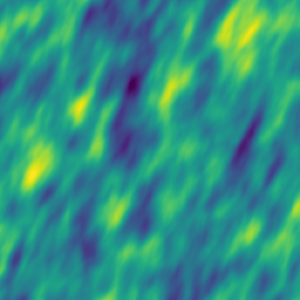}
         & \includegraphics[width=\hsize]{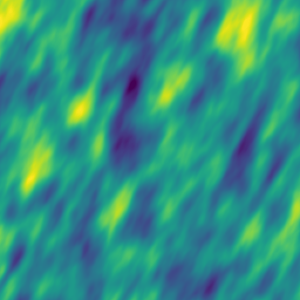}
         & \includegraphics[width=\hsize]{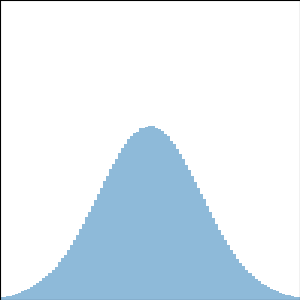}\\\addlinespace[-3pt]
    \end{tabular}
    \caption{Samples visualization for data index $4$ with $\measdim = 600$ and mask-type "unif", corresponding to the seventh line of \Cref{table:posterior_sampling}.}
    \label{fig:img_4_size_3_unif:show}
\end{figure}

\paragraph{Results:} We show in \Cref{fig:img_0_size_2_clust:show,fig:img_0_size_2_unif:show,fig:img_1_size_2_clust:show,fig:img_1_size_2_unif:show,fig:img_1_size_3_clust:show,fig:img_1_size_3_unif:show,fig:img_4_size_3_clust:show,fig:img_4_size_3_unif:show}
samples from all the samplers and configurations in \Cref{table:posterior_sampling}.

We note that, as described in \Cref{table:posterior_sampling}, the posterior samplers become better when the number of observations increases. What we can note in the figures is that
the standard deviation of the errors is not at all the prescribed standard deviation, indicating that all posterior samplers seem not to be calibrated. It would be interesting to see what are the possible fixes to achieve better calibration
and the impact that this has in the {\maxsw} metrics shown in \Cref{table:posterior_sampling}.
\section{Implementation details}
\label{app:num}
\subsection{Generative model Architecure}
\label{app:num:architecture}
\paragraph{Changes with respect to \cite{karras2024analyzing}.}

We followed the architecture defined in \cite{karras2024analyzing}\footnote{Github available at \url{https://github.com/NVlabs/edm2}}.
The main adaptation was to change the channel multiplier (see \cite[Table 6]{karras2024analyzing}), which dictates the first layer width of the Unet which is then multiplied by the "per layer multiplier" parameter to determine the width of all the other layers.
Namely, an UNet with channel multiplier $x$ and per layer multipliers $y_1, y_2, y_3$ will have depths $xy_1, xy_2, xy_3$ respectively.
This has a great impact on the memory footprint during training, mainly due to the skip connections that are kept.

Therefore, we went from the standard EDM parameterization where channel multiplier and per layer multipliers are $(192, [1, 2, 3, 4])$ to $(64, [1, 1, 2, 2, 4])$. The self-attention layer is only active in the last Unet layer corresponding to a resolution of $16$ pixels.
This allows using a batch size of $32$ instead of the batch size of $8$ allowed for the original parameterization. We do not claim that this choice is optimal, but it was imposed by computing budget constraints. We also fixed the $\sigma_{data}$ parameter to $1$, to better reflect the estimated data standard deviation. All the other parameters follow the configuration corresponding to the model size "S" in \cite{karras2024analyzing}. 

\paragraph{Changes on importance distribution}

\begin{figure}[h]
    \centering
    \includegraphics[width=0.5\textwidth]{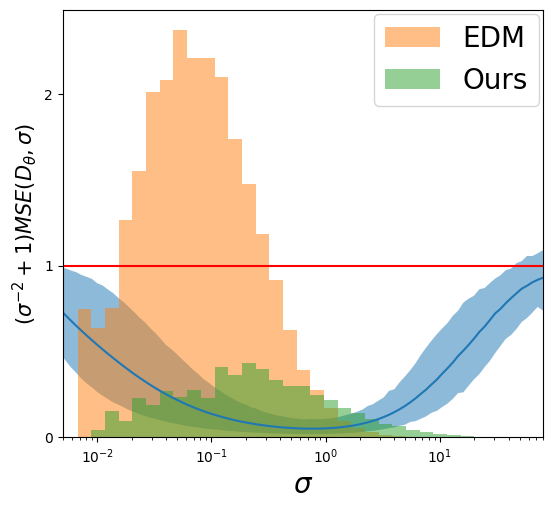}
    \caption{Reproduction of \cite[Figure 5a]{karras2022elucidating}. The blue curve represents the mean performance over a random chunk of $320$ elements of the training set for the final model (anisotropic) and the shaded region represents the percentile $90\%$ to $10\%$.
    The orange distribution represents an histogram of the log-Gaussian distribution proposed in \cite{karras2022elucidating} and the green one our adaptation.} 
    \label{fig:importance_distribution}
\end{figure}

A key aspect during training of the EDM architectures is the choice of importance distribution to be used during training, as explained in \cite[Section 5]{karras2022elucidating}. In \cite{karras2022elucidating} 
the authors propose using a log-Gaussian distribution with mean and standard deviation $(-1.2, 1.2)$. This choice is based on \cite[Figure 5a]{karras2022elucidating}, namely,
by focusing training in the parts of the $\vestd$ range where the network is able to learn the most. We reproduced the same figure in \Cref{fig:importance_distribution}
where we ploted as $(1 + \vestd^{-2}) \mseloss{\dnet{\cdot, \vestd}}{\vestd}$ as a function of $\vestd$. As per \cite{karras2022elucidating}, this quantity is chosen because it is around $1$ for all $\vestd$ in the beginning of training.
For our trained network (blue in \Cref{fig:importance_distribution}), we see that where we have the most improvement is around $\vestd \approx 1$. Thus, we changed the mean and variance of the log-Gaussian distribution to $(0.7, 1.5)$ respectively.
\subsection{Hardware}
\label{app:num:hardware}
\paragraph{GPU server:}
All the training and simulation for both the DGM and DGM related posterior samplers as well as the {\priorvae} examples were trained in a server 
with several Nvidia V100 SXM2 HBM2 with both the versions with $16$ Go and $32$ Go of RAM memory.
The nodes could scale up to 40 different CPU cores, consisting of both Intel Cascade Lake 6248 or Intel Cascade Lake 6226 and $8$ in-node GPUs.

\paragraph{CPU cluster:} All the MCMC chains computed in this work (cf. \Cref{sec:num:cond_gen}) were ran on a CPU cluster with 32 nodes, each composed of two  Intel(R) Xeon(R) CPU E5-2640 v4 (2.40GHz).
\subsection{C2ST: Training details}
We trained all the networks using the Adam optimizer \cite{kingma2015adam} with cosine annealing learning rate schedule between $3\times 10^{-4}$ and $10^{-8}$
for $1000$ epochs with batch size of $512$.
\label{app:num:c2st}
\subsection{Fine tuning}
\label{app:num:fine_tuning}
We considered two cases of fine tuning experiments, consisting of retraining for a few epochs the DGM prior learning from the GRF prior described in \Cref{sec:num:data}.

\paragraph{Global anisotropy:} This prior consists of a (centered) GRF priors, as defined in \Cref{sec:grf_spde}, but which anisotropies are constant across space (i.e. the matrix $\bm Q_\ptD$ does not depend on $\ptD\in\cD$). Hence, only three parameters are needed to characterize the prior: the range $a$, the anisotropy ratio $\min\lbrace\ascl_1, \ascl_2\rbrace/\max\lbrace\ascl_1, \ascl_2\rbrace$ and an angle $\theta$ parametrizing the unique (global) direction of correlation of the GRF. Compared to the prior parameters specified in \Cref{sec:num:data}, the following changes are made.  We  replace the spline parametrization of the function $f$, by a unique parameter $\theta \sim  \mathcal{U}([0,\pi])$ specifying the direction of anisotropy.  We fixe the parameter $\rho_1=1$ and take $\ascl_2 \sim \mathcal{U}([0.1,1])$ (to avoid a redundancy on the specification of the anisotropy direction). The other parameters ($a$, $\nu$) are specified in the same way as in \Cref{sec:num:data}.

\begin{figure}[h]
	\centering
	\begin{tabular}{
			M{0.16\linewidth}@{\hspace{0.15\tabcolsep}}
			M{0.16\linewidth}@{\hspace{0.15\tabcolsep}}
			M{0.16\linewidth}@{\hspace{0.15\tabcolsep}}
			M{0.16\linewidth}@{\hspace{0.15\tabcolsep}}
			M{0.16\linewidth}@{\hspace{0.15\tabcolsep}}
			M{0.16\linewidth}@{\hspace{0.15\tabcolsep}}
			M{0.16\linewidth}@{\hspace{0.15\tabcolsep}}
		} \includegraphics[width=\hsize]{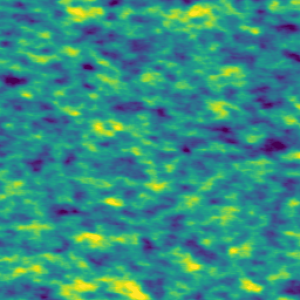} 
		& \includegraphics[width=\hsize]{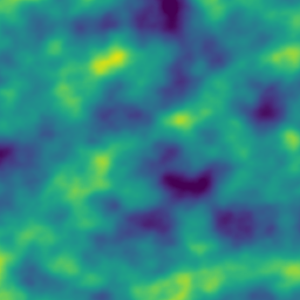} 
		& \includegraphics[width=\hsize]{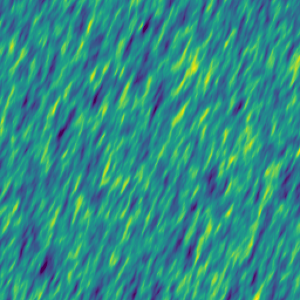} 
		& \includegraphics[width=\hsize]{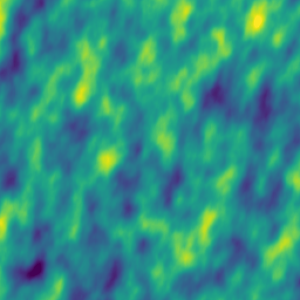} 
		& \includegraphics[width=\hsize]{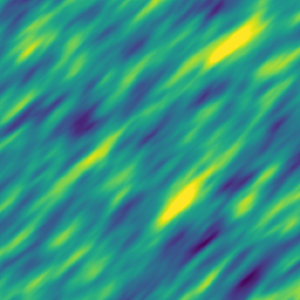}\\\addlinespace[-3pt]
	\end{tabular}
	\caption{Examples of samples from the Global Anisotropy GRF prior (generated by the fine-tuned DGM).}
	\label{fig:illustration_fine_tuning_ganiso}
\end{figure}

\begin{figure}[h]
	\centering
	\includegraphics[width=0.6\textwidth]{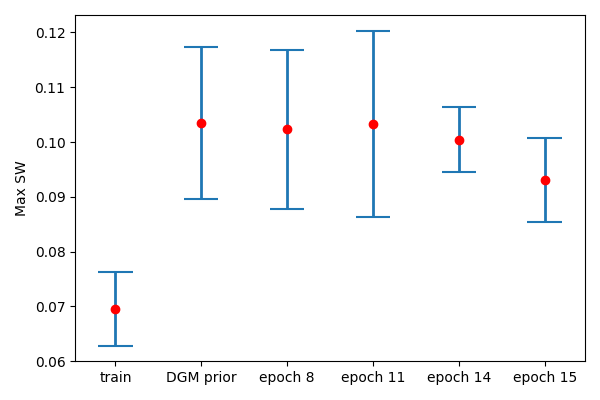}
	\caption{Figure representing the {\maxsw} computed for the DGM fine-tuned on the Global anisotropy GRF prior. We show the {\maxsw} obtained at different epochs for the DGM fine-tuning, and the {\maxsw} between the validation GRF samples and the GRF samples used during training (\q{train}), and the {\maxsw} computed using \q{untuned} DGM prior (\q{DGM prior}). The error bars are $2\sigma$ error bars, and the red do marks the mean.}\label{fig:max_sw_fine_tuning_ganiso}
\end{figure}

We generated 300,000 samples from this new GRF prior, and retrained the DGM prior based on 250,000 of these samples We show in \Cref{fig:illustration_fine_tuning_ganiso} examples of samples generated by the fine-tuned DGM. We ran a total of $16$ epochs and computed the {\maxsw} between 50,000 samples samples generated with the newly trained DGM (using the {\ddpm} sampler with 1000 steps and $\rho=3$), and the 50,000 GRF samples left. The results are shown in \Cref{fig:max_sw_fine_tuning_ganiso}. We calculated the {\maxsw} with $2^{17}$ slices and $50 000$ samples, with $10$ replicates (randomized over slices and subsamples).
The results are shown in \cref{fig:max_sw_vs_rho}, where the error bars correspond to $2$ times the standard deviation.
\paragraph{Swirly GRFs:} This prior consists of a (centered) GRF priors, as defined in \Cref{sec:grf_spde}, but which anisotropies are \q{swirl}-shaped. Mathematically, this means that the local anisotropy directions can be parametrized as being orthogonal to the gradient of a scalar function defined across $\cD$.
Compared to the prior parameters specified in \Cref{sec:num:data}, the following changes are made.  We fix the parameter $\rho_2=1$ and take $\ascl_1 \sim \mathcal{U}([0.1,1])$, so that the anisotropies are direct along the orthogonal of the gradient of the function $f$. The parameter $a$ is now drawn from a lognormal distribution with mean $0.01$ and $0.1$ and standard deviation $0.01$ (to force a relatively small anisotropy ratio). The other parameters ($f$, $\nu$) are specified in the same way as in \Cref{sec:num:data}.

\begin{figure}[b]
	\centering
	\begin{tabular}{
			M{0.16\linewidth}@{\hspace{0.15\tabcolsep}}
			M{0.16\linewidth}@{\hspace{0.15\tabcolsep}}
			M{0.16\linewidth}@{\hspace{0.15\tabcolsep}}
			M{0.16\linewidth}@{\hspace{0.15\tabcolsep}}
			M{0.16\linewidth}@{\hspace{0.15\tabcolsep}}
			M{0.16\linewidth}@{\hspace{0.15\tabcolsep}}
			M{0.16\linewidth}@{\hspace{0.15\tabcolsep}}
		} \includegraphics[width=\hsize]{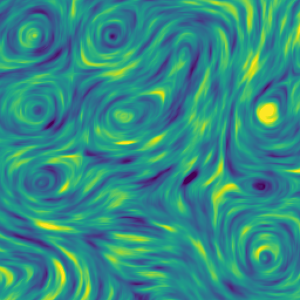} 
		& \includegraphics[width=\hsize]{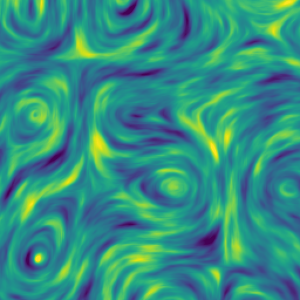}
		& \includegraphics[width=\hsize]{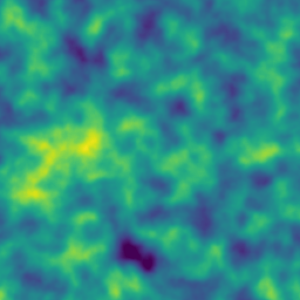}
		& \includegraphics[width=\hsize]{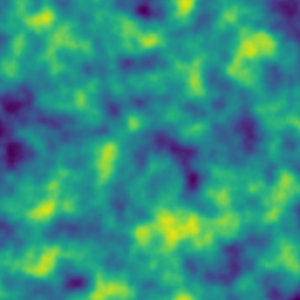}
		& \includegraphics[width=\hsize]{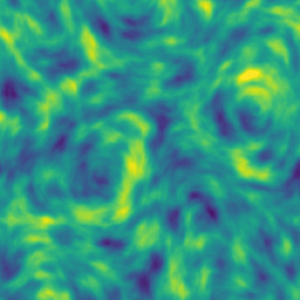}\\\addlinespace[-3pt]
	\end{tabular}
	\caption{Examples of samples from the Swirly GRF prior (generated by the fine-tuned DGM).}
	\label{fig:illustration_fine_tuning_turb}
\end{figure}

We generated 300,000 samples from this new GRF prior, and retrained the DGM prior based on 250,000 of these samples. We show in \Cref{fig:illustration_fine_tuning_turb} examples of samples generated by the fine-tuned DGM. We ran a total of $5$ epochs and computed the {\maxsw} between 50,000 samples generated with the newly trained DGM, and the 50,000 GRF samples left. We calculated the {\maxsw} with $2^{16}$ slices and $50 000$ samples, with $20$ replicates (randomized over slices and subsamples). The results are shown in \Cref{tab:max_sw_fine_tuning}.
\begin{filecontents*}{data/max_sws_fine_turb.csv}
,sampler,n_steps,rho,mean,std,len,clt_gap,value
1,ddpm,100,fine_tuning,0.11258386708796024,0.0024706472674926715,20,0.001082808906753393,0.113 + 0.001
3,ddpm,1000,fine_tuning,0.10191431790590286,0.0029034041223315915,20,0.0012724729608025644,0.102 + 0.001
0,train,0,,0.06757524379700067,0.0036605758284869224,20,0.0016043180923007163,0.068 + 0.002
\end{filecontents*}
\DTLloaddb{max_sws_fine_tun}{data/max_sws_fine_turb.csv}
\begin{table}[h]
    \begin{center}
        \begin{tabular}{|c|c|c|}%
            \hline
            Sampler & N steps & \maxsw  \\
            \hline
            \DTLforeach*{max_sws_fine_tun}{
                \samplerfin=sampler,
                \nstepswfin=n_steps,
                \mean=mean,
            	\std=std}
                {
                \samplerfin & \nstepswfin & \DTLround{\mean}{\mean}{3}$\mean$ (\DTLround{\std}{\std}{3}$\std$)
                \DTLiflastrow{}{\\}}
            \\\hline
        \end{tabular}% 
    \end{center}
    \caption{ {\maxsw} computed for the DGM fine-tuned on the Swirly GRF prior,  in the form "mean (standard deviation)". We show the {\maxsw} obtained using two {\ddpm} samplers for the DGM (same $\rho=3$, varying number of steps), and the {\maxsw} between the validation GRF samples, and the GRF samples used during training.}
    \label{tab:max_sw_fine_tuning}
\end{table}

These numerical experiments show that in few epochs one is able to adapt the denoiser to different priors, at least as long as they are in the same parametrization. 
\subsection{\vecchia: details}
\label{app:num:vecchia}
As the number of observations is of order $40,000$, the computational cost of evaluation of the Gaussian likelihood become prohibitive since the considered GRF priors do not yield sparse matrices, are and very non-stationary. To circumvent this problem, we follow use Vecchia approximations of the true likelihood, which allows to treat cases where the number of observations is of a few thousands. 

\begin{figure}
	\centering
	\includegraphics[height=0.3\textheight]{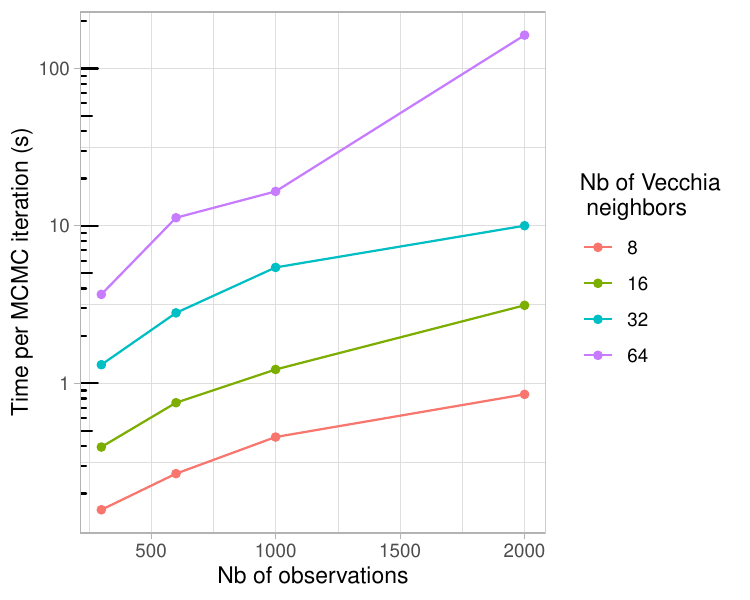}
	\caption{Computation time (in seconds) for a single {\vecchia} iteration. Each time was measured by running a chain for 30 iterations (10 for the cases where 64 neighbors are considered for the Vecchia approximation), and diving the running time by the number of iterations. The experiment was run on a CPU cluster with 32 nodes, each composed of two  Intel(R) Xeon(R) CPU E5-2640 v4 (2.40GHz) processors.}\label{fig:mcmc_time}
\end{figure}

We use the implementation of Sparse General Vecchia approximation of the R package BayesNSGP \cite{risser2020bayesian}, which we slightly tweaked to fit the GRF prior we consider. We favored the Sparse General Vecchia approximation to the Nearest Neighbors Gaussian process approach as it is known to provide a better approximation. However, despite the efficient implementation in the package, we could not consider the whole set of observations to get posterior samples in reasonable time. Indeed, as shown in \Cref{fig:mcmc_time}, the cost of a single MCMC iteration, and therefore the cost of sampling from the PPD scales dramatically as the number of neighbors used in the Vecchia approximation, and with the number of observations grows. For instance, using 1,000 observations with 64 neighbors requires around 16s per iterations. As tens of thousands of iterations are required (given the complexity of the prior and the number of parameters), running 50,000 iterations  (which was done in the numerical experiments of \cite{risser2020bayesian}, albeit with less observations) would result in a computation time of around 9 days. That is why we only consider subset of 1,000 observations uniformly drawn from the unclouded locations, and fixed the number of neighbors to 16.

We ran Random Walk Metropolis Hastings MCMCs to determine the posterior distribution of the 38 parameters from our GRF priors: the correlation range and the anisotropy ratio are sampled independently and the 36 spline nodes modeling the spatially varying anisotropy angles are sampled by block (4 blocks corresponding to a subdivision of $\cD$ into for quadrants). We then generated posterior GRF samples using these parameters (using the \textbf{nsgpPredict} function of the package). In order to mimic the prior we used to train the DGM, while fitting the was models are specified in the \textbf{BayesNSGP} package, we parameterized the angles with a spline method relying on the same nodes as the ones used for the GRF prior of the DGM. The difference is now that the value of the nodes represent a logit transform of the angle (instead of a function which gradient specifies the direction of the anisotropy).

We ran 100 independent chains, for 75,000 iterations, for each inverse problem, on the CPU cluster described in \Cref{app:num:hardware}. The computation time of each chain was of around 25 hours. As a reference, we show in \Cref{table:timing_algs} the computation time needed to generate a single posterior sample using {\vecchia} or the DGM-based posterior samplers.
\begin{filecontents*}{data/post_sampling_timing.csv}
,alg,n_obs,mean,std,n,vecchia_nbrs
0,DPS,256,1.7,0.0,10,
36,MGDM,256,6.4,0.0,10,
24,MGPS,256,3.7,0.0,10,
14,VMCMC,300,196.7,,1,8.0
2,VMCMC,300,492.5,,1,16.0
6,VMCMC,300,1641.2,,1,32.0
10,VMCMC,300,4589.6,,1,64.0
1,DPS,512,1.6,0.0,10,
37,MGDM,512,6.4,0.0,10,
25,MGPS,512,3.7,0.0,10,
15,VMCMC,600,333.7,,1,8.0
3,VMCMC,600,941.5,,1,16.0
7,VMCMC,600,3503.5,,1,32.0
11,VMCMC,600,14085.8,,1,64.0
12,VMCMC,1000,570.1,,1,8.0
0,VMCMC,1000,1530.9,,1,16.0
4,VMCMC,1000,6809.5,,1,32.0
8,VMCMC,1000,20692.1,,1,64.0
2,DPS,1024,1.7,0.0,10,
38,MGDM,1024,6.2,0.0,10,
26,MGPS,1024,3.6,0.0,10,
13,VMCMC,2000,1063.9,,1,8.0
1,VMCMC,2000,3912.4,,1,16.0
5,VMCMC,2000,12534.2,,1,32.0
9,VMCMC,2000,203395.8,,1,64.0
3,DPS,2048,1.6,0.0,10,
39,MGDM,2048,6.2,0.0,10,
27,MGPS,2048,3.6,0.0,10,
4,DPS,4096,1.6,0.0,10,
40,MGDM,4096,6.2,0.0,10,
28,MGPS,4096,3.6,0.0,10,
5,DPS,8192,1.6,0.0,10,
29,MGPS,8192,3.6,0.0,10,
\end{filecontents*}
\DTLloaddb{timing_algs}{data/post_sampling_timing.csv}
\begin{table}
    \begin{center}
        \begin{tabular}{|c|c|c|c|}%
            \hline
            Sampler & $\measdim$ & Nb Vecchia neighbors & Computation time (min) \\
            \hline
            \DTLforeach*{timing_algs}{
                \algnametiming=alg,
                \measdimtiming=n_obs,
                \nnbrstiming=vecchia_nbrs,
				\dttiming=mean,
				\dtstdtiming=std}
                {
                \algnametiming & \measdimtiming &\nnbrstiming & \dttiming \quad \DTLifnullorempty{\dtstdtiming}{}{(\dtstdtiming)}
                \DTLiflastrow{}{\\}}
            \\\hline
        \end{tabular}% 
    \end{center}
    \caption{Comparison of computation time to generate a single posterior sample using {\vecchia}  or DGM-based algorithm. The values are the mean and standard deviation. The cases where standard deviation does not appear correspond to cases where generating replications would be extremely expensive and we refrain to do so. For {\vecchia}, we compute the time needed to run a MCMC chain with 75,000 iterations.}
    \label{table:timing_algs}
\end{table}

\subsection{\priorvae: details}
\label{app:num:prior_vae}
As the original \priorvae (\cite{semenova2022priorvae}) code was only available for 1-d kernels, we trained a VAE using the same data as for the DGM.
For the architecture, we used a model inspired by the code used for \cite{rombach2022highresolution}\footnote{available at \url{https://github.com/CompVis/latent-diffusion}}.
The latent dimension is $8 \times 32 \times 32$ and we used a Gaussian Gaussian VAE, with diagonal covariance.
The training was done using Adam \cite{kingma2015adam} with learning rate $10^{-3}$.

\paragraph{Posterior sampling:}

The likelihood induced on the latents for the VAE is 
\begin{equation}
    \ell_{VAE}(\lmeas | z) = \int \ell(\lmeas | \lstate) \gauss(\lstate; \mu_{\param}(z), \Sigma_{\param}(z)) \rmd \lstate\eqsp.
\end{equation}
While it is possible to use the reparametrization trick estimator to compute gradients, it has shown to be unstable with the NUTS sampler \cite{hoffman2011nouturn} leading to vanishing learning rate.
Therefore, for the experiments running NUTS we used the simplified potential which consists of 
\begin{equation}
    \tilde{\ell}_{VAE}(\lmeas | z) = \ell(\lmeas | \mu_{\param}(z))\eqsp.
\end{equation}

We use the NUTS sampler for 120 iterations, where 20 iterations are considered warm-up iterations for setting the learning rate and the diagonal mass matrix to reach $0.8$ acceptance probability.
The initial learning rate is $10^{-4}$. We start NUTS using the outcome of an Unadjusted Langevin sampler \cite{durmus2017nonasymptotic} which was run for 1000 steps with learning rate $10^{-4}$ and started from a standard Gaussian distribution.
The full procedure lasted around $22$ minutes running on GPU.
\subsection{Sea surface temperature anomaly data: details}
\label{app:num:sea_temperature_details}

\begin{figure}
	\centering
	\includegraphics[width=0.95\textwidth]{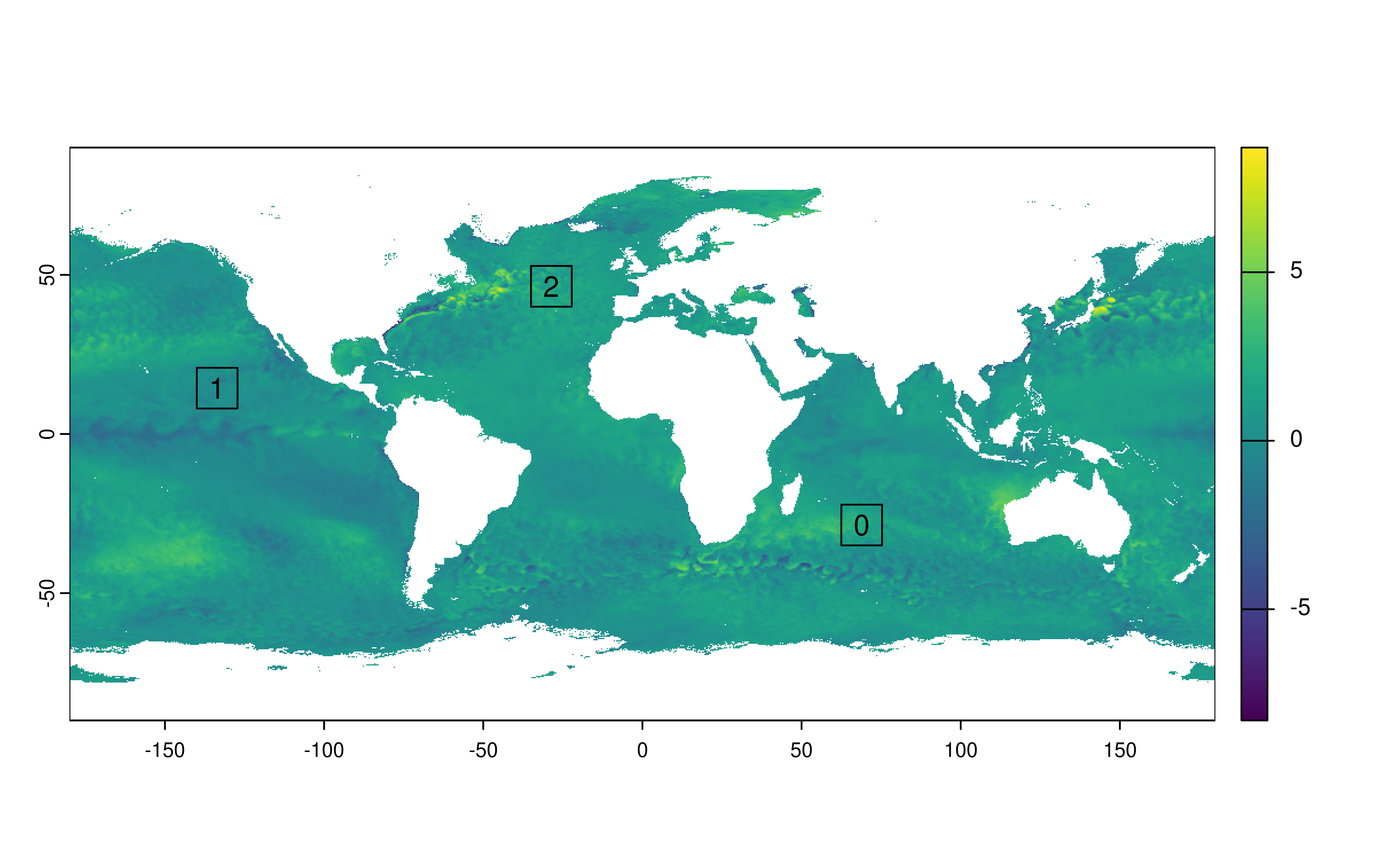}
	\caption{Map representing the SSTA data on January 1st, 2025. The three squares represent the zones selected to the numerical experiment presented in this work}\label{fig:map_ssta}
\end{figure}

The SSTA data are extracted from the NOAA Coral Reef Watch database \cite{noaa2019coral}, and corresponds to SSTA, on on January 1st, 2025 and on three parts of the globe represented in \Cref{fig:map_ssta}. The SSTA raw data were downloaded from the NOOA website: \url{https://www.ncei.noaa.gov/data/oceans/crw/5km/v3.1/nc/v1.0/daily/ssta/2025/} [last accessed: May 15th, 2025]. The data consist of gridded values, across the globe, of SSTA the . We picked 3 fairly separated zones on the globe, while targeting zones where the observed values could roughly be considered as having a constant mean. This because the GRF prior we consider is centered. This is a limitation of our approach, which we discuss in \Cref{sec:limits}. Hence, the extracted SSTA data on each zone are standardized prior to the PPD computations by removing the mean and scaling by the standard deviation of the observed values (i.e. the unclouded locations). The PPD results are presented in this standardized scale.  
%coast of South Africa.

The cloud mask is extracted from NASA’s MODIS/Aqua Cloud Mask product \cite{MODIS_Atmosphere_Science_Team2017}. We extracted three pairs of cloud masks from the website: \url{https://ladsweb.modaps.eosdis.nasa.gov/search/order/2/MYD35_L2--61} [last accessed : May 15th, 2025]. We entered the following query:
\begin{itemize}
	\item Product : MYD35 L2
	\item Time : 2025-01-01
	\item Location : World
	\item Times selected (for the cases 0, 1  and 2 respectively): 08:45, 10:05, 10:05
\end{itemize}

\subsection{Scoring rules for probabilistic forecasts}\label{app:scores}
The Continuous Ranked Probability Score (CRPS) is a metric used to evaluate (scalar) probabilistic forecasts \cite{matheson1976scoring}. Given a predictive distribution $p$ and a scalar value (denoting the observation of the variable we seek to predict) $y$, the CRPS is defined as
\begin{equation}
	\textrm{CRPS}(p,y)=\E[\vert Y-y\vert]-\frac{1}{2}\E[\vert Y-Y'\vert]
\end{equation}
where $Y, Y' \sim p$. As a proper scoring rule, it is minimal when $y$ is a sample from $p$. When only (independent) samples from the predictive distribution are available, the CRPS is approximated by via the following Monte-Carlo estimator
\begin{equation}
	\textrm{CRPS}(p,y) \approx \frac{1}{m}\sum_{k=1}^{m}\vert Y_k - y\vert - \frac{1}{2m^2}\sum_{k=1}^{m}\sum_{l=1}^{m}\vert Y_k - Y_l\vert
\end{equation}
where $Y_1, \dots, Y_m$ are independent samples from $p$. 

\begin{figure}
	\centering
	\includegraphics[width=\textwidth]{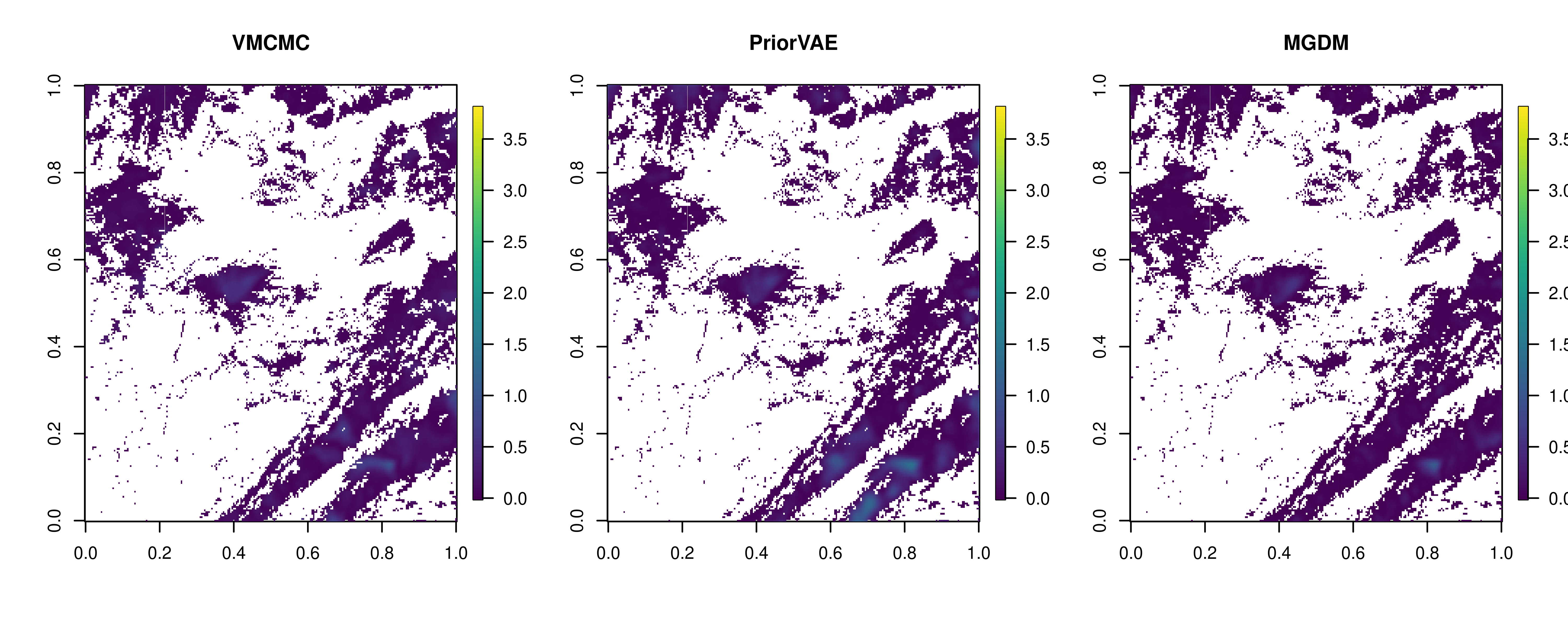}\\
	\includegraphics[width=\textwidth]{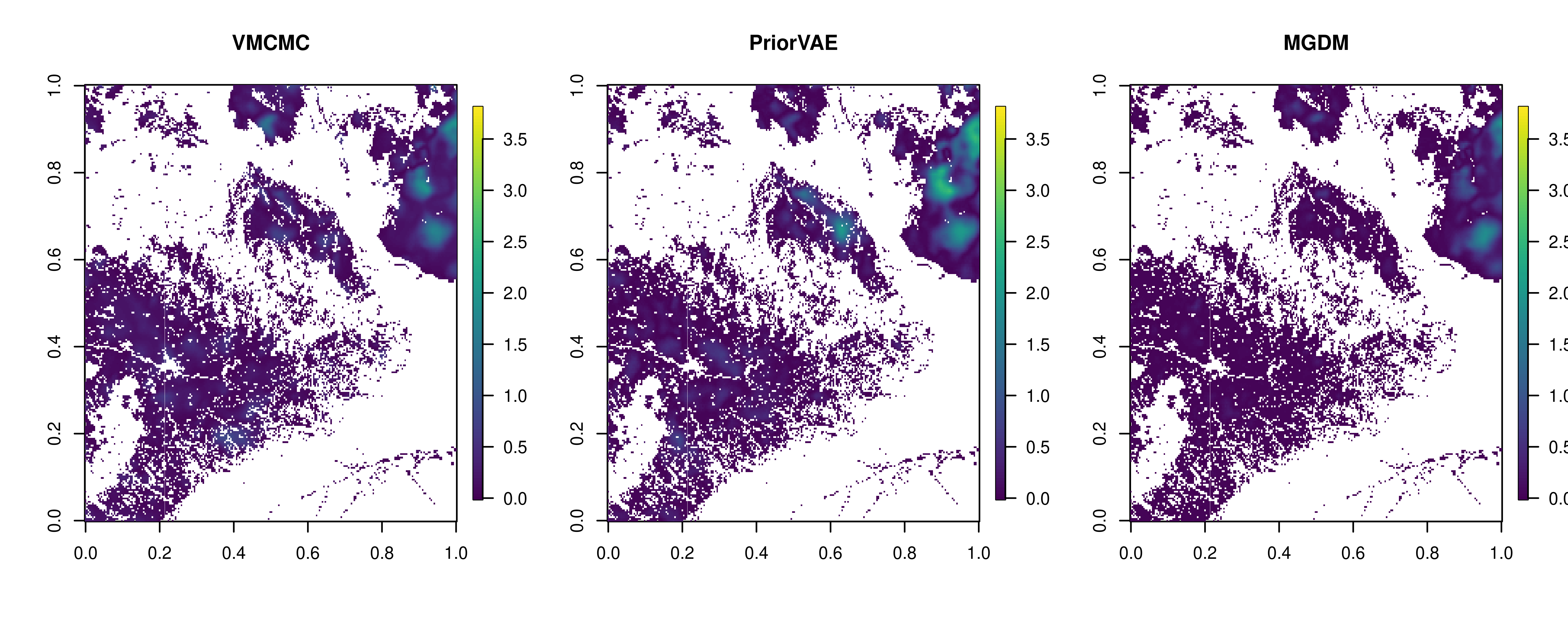}\\
	\includegraphics[width=\textwidth]{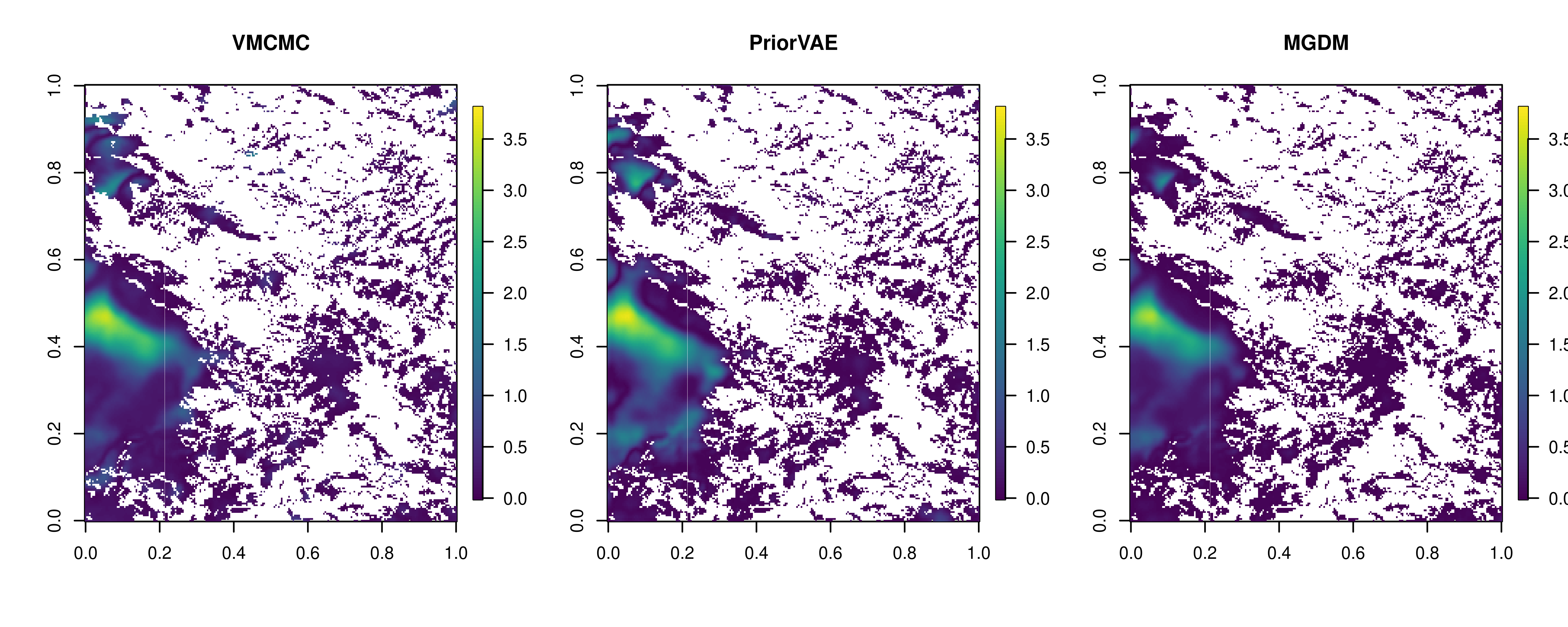}
	\caption{Maps of CRPS computed for the {\mgdm}, {\priorvae} and {\vecchia} PPDs, for each inverse problem (on each row) considered in the SSTA numerical experiment. The white parts correspond to the observations.}\label{fig:map_crps}
\end{figure}

In our numerical experiment on SSTA data (cf. \Cref{sec:num:cond_gen}), we use the CRPS to evaluate the PPDs (obtained by {\mgdm}, {\priorvae} or {\vecchia}) at locations covered by clouds. To do so, we started by generating 100 posterior samples for each sampling method.
As the CRPS is a metric suited for scalar predictions, we compute it on each unobserved location separately based on the samples obtained for the different posterior samplers. We show in \Cref{fig:map_crps} the CRPS maps obtained for each posterior sample and for each inverse problem. As we can notice, {\mgdm} seems to provide PPDs with low CRPS values on more prediction locations than the other two posterior samplers, while the areas where the PPDs have higher CRPS with  {\mgdm}  are also shared by the other samplers (and correspond roughly to locations the furthest away from the data).
We then create 32 (disjoint) sets of unobserved locations sampled uniformly, and compute, for each set, the mean CRPS over the set. The mean and standard-deviation of these averaged CRPS are presented in \Cref{table:rec_cloud}.

We also computed a multivariate scoring rule to evaluate the different posterior samplers: the Energy Score (ES).
The energy score is an extension of the CRPS, tailored to multivariate forecasts \cite{gneiting2008assessing}. It is defined, for a multivariate predictive distribution $\tilde p$ and an observed vector $y\in\rset^d$ ($d\ge 1$), as 
\begin{equation}
	\textrm{ES}(p,y)=\E[\Vert Y-y\Vert]-\frac{1}{2}\E[\Vert Y-Y'\Vert]
\end{equation}
where $Y, Y' \sim \tilde p$ and $\Vert\cdot\Vert$ denotes the Euclidean metric. The energy score is also a proper scoring rule, and can be approximated from independent samples $Y_1, \dots, Y_m \sim p$ as 
\begin{equation}
	\textrm{ES}(p,y) \approx \frac{1}{m}\sum_{k=1}^{m}\Vert Y_k - y\Vert - \frac{1}{2m^2}\sum_{k=1}^{m}\sum_{l=1}^{m}\Vert Y_k - Y_l\Vert
\end{equation}
We repeat the same approach consisting of separating the set of unobserved locations into 32 subsets to compute a mean and standard deviation for the ES. We present the results in in \Cref{table:rec_cloud_bis}. As we notice, once again, {\mgdm} systematically outperforms the other two posterior samplers.
\begin{filecontents*}{data/es_post_temp.csv}
,img_id,vecchia,priorvae,mgdm
0,0,3.577 (0.179),4.722 (0.406),1.843 (0.194)
0,1,8.452 (0.49),11.241 (0.713),5.887 (0.59)
0,2,17.807 (1.04),20.107 (1.228),14.955 (1.063)
\end{filecontents*}
\DTLloaddb{es_temp}{data/es_post_temp.csv}
\begin{table}[h]
	\begin{center}
		\begin{tabular}{|c|c|c|c|}%
			\hline
			Case & \vecchia & \priorvae & \mgdm \\
			\hline
			\DTLforeach*{es_temp}{
				\index=img_id,
				\vec=vecchia,
				\vae=priorvae,
				\mgd=mgdm}%
			{%
				\index & \vec &\vae & \mgd  \DTLiflastrow{}{\\}
			}
			\\\hline
		\end{tabular}%
	\end{center}
	\caption{ES on the SSTA problem for three cases (lower is better), in the form \q{mean (standard deviation)}. The unobserved locations are randomly separated into 32 disjoint subsets, on which the ES is computed. The mean and standard deviation of these values are shown above.}
	\label{table:rec_cloud_bis}
\end{table}%
We used the R package \textbf{scoringRules} to compute these two metrics in our numerical experiments \cite{scoringRules}.
\subsection{Hours used and CO2 equivalent budget}
During the full duration of the process, a total of $27764$ GPU hours were used, amounting to an equivalent $714$ kg$\operatorname{CO}_2$. This includes failed training and prototype experiments
which are estimated to have cost a total of $10 000$ hours.
In the \Cref{table:co2_per_procedure}, we recapitulate the order of magnitudes for the main tasks done in this work.
\begin{table}[h]
    \centering
    \begin{tabular}{|c|c|c|}
        \hline
        Task & GPU Hours & Eq kg$\operatorname{CO}_2$\\ \hline
        Training from scratch & 3200 & 82.3\\\hline
        Fine-tuning & 640 & 16.5\\\hline
        Generation 50k samples (worst-case) & 110.5 & 2.9 \\\hline
        {\maxsw} & 0.3 & 0.008\\\hline
        {\c2st} Resnet18 & 40 & 1.0 \\\hline
        {\c2st} Resnet50 & 88 & 2.25\\\hline
        {\c2st} Resnet101 & 130 & 3.32\\
        \hline
    \end{tabular}
    \label{table:co2_per_procedure}
    \caption{Approximate values for GPU hours and equivalent $\operatorname{CO}_2$ for all main tasks carried over in the paper.}
\end{table}
\label{app:num:impact}
\subsection{Posterior sampling implementation details}
\label{app:num:post_sampling_details}
All the details of the main parameters for the samplers in \Cref{table:posterior_sampling} are shown in \Cref{table:post_sampling:params}.
More information is available at \url{{\githublink}} in the folder \texttt{python/diff\_post\_gauss/configs/conditional\_sampler}.
\begin{table}
    \centering
    \begin{tabular}{|c|c|c|c|c|}
        \hline
        Sampler & learning rate & $\enddiff$ & Other parameters & $\Delta t$ \\
        \hline
        \dps & 1 & 1000 & NA & 1.7 \\\hline
        \mgps 
        & $3\times10^{-2}$ 
        & $300$ 
        & \begin{tabular}{c|c}
            $t \leq \frac{3\enddiff}{4}$ & $10$ \\
            $t > \frac{3\enddiff}{4}$ & $2$
         \end{tabular}
        & 3.7 \\\hline
        \mgdm &  
         \begin{tabular}{c|c}
            $t \geq \frac{3\enddiff}{4}$ & $1 \times 10^{-2}$ \\
            $t < \frac{3\enddiff}{4}$  & $3 \times 10^{-2}$ \\ 
         \end{tabular}& 100 &
         \begin{tabular}{c|c}
            Gibbs steps & $2$ 
         \end{tabular}& 6.4 \\ \hline
    \end{tabular}
    \caption{Table with parameterization used for all the experiments.}
    \label{table:post_sampling:params}
\end{table}
\subsection{Tier code and licenses}
\label{app:num:tier_code_licenses}
\begin{itemize}
    \item EDM2 \url{https://github.com/NVlabs/edm2} \cite{karras2024analyzing}: Creative Commons Attribution-NonCommercial-ShareAlike 4.0 International License \url{https://creativecommons.org/licenses/by-nc-sa/4.0/}.
    \item MGDM \url{https://github.com/Badr-MOUFAD/mgdm/tree/main} \cite{moufad2024variationala}: Creative Commons Attribution-NonCommercial-ShareAlike 4.0 International License \url{https://creativecommons.org/licenses/by-nc-sa/4.0/}.
    \item LDM \url{https://github.com/CompVis/latent-diffusion/tree/main} \cite{rombach2022highresolution}: MIT License 
    \item BayesNSGP \url{https://github.com/cran/BayesNSGP} \cite{risser2020bayesian}: GPL-3 License
\end{itemize}

Otherwise, the paper relies heavily on Pytorch \cite{ansel2024pytorch} and in particular the Pytorch Lightning library \cite{sawarkar2022deep}.
The NUTS implementation from Pyro \cite{bingham2019pyro} was used for the {\priorvae} examples.
\end{document}